\documentclass{article}
\usepackage{spconf,amsmath,epsfig}
\usepackage{hyperref}
\hypersetup{colorlinks,allcolors=black}
\usepackage{subfig}
\usepackage{url}
\usepackage{graphicx}
\graphicspath{ {./images/} }
\usepackage{algorithm} 
\usepackage{algpseudocode}
\usepackage{rotating}
\usepackage{tabularray}
\UseTblrLibrary{siunitx}
\usepackage{tabularx, booktabs, makecell, caption}
\usepackage{siunitx}
\usepackage{dsfont}
\usepackage{amssymb}
\usepackage{afterpage}
\usepackage{pifont}
\usepackage{mathtools}
\usepackage{amsmath}
\usepackage{pythonhighlight}
\lstnewenvironment{mypython}[1][]{\lstset{style=mypython,#1}}{}
\usepackage{cleveref}
\Crefname{table}{Tab.}{Tabs.}

\let\OLDthebibliography\thebibliography
\renewcommand\thebibliography[1]{
  \OLDthebibliography{#1}
  \setlength{\parskip}{0pt}
  \setlength{\itemsep}{0pt plus 0.3ex}
}

\begin{document}\sloppy

\def\x{{\mathbf x}}
\def\L{{\cal L}}

\title{DiffImpute: Tabular Data Imputation With Denoising Diffusion Probabilistic Model}
%
\name{Yizhu Wen\textsuperscript{1}, Kai Yi\textsuperscript{2}, Jing Ke\textsuperscript{3}, Yiqing Shen\textsuperscript{4,*}\thanks{* Corresponding Author. Email: yshen92@jhu.edu}}
\address{
\textsuperscript{1} Spin-Quantum Co. LTD, Xian, Shaanxi, China\\
\textsuperscript{2}School of Mathematics and Statistics, University of New South Wales, Sydney, NSW, Australia\\
\textsuperscript{3}School of Electronic Information and Electrical Engineering
Shanghai Jiao Tong University, China \\
\textsuperscript{4}Department of Computer Science, Johns Hopkins University, Baltimore, MD, USA
}

\maketitle

\begin{abstract}
Tabular data plays a crucial role in various domains but often suffers from missing values, thereby curtailing its potential utility. 
Traditional imputation techniques frequently yield suboptimal results and impose substantial computational burdens, leading to inaccuracies in subsequent modeling tasks.
To address these challenges, we propose \texttt{DiffImpute}, a novel Denoising Diffusion Probabilistic Model (DDPM).
Specifically, \texttt{DiffImpute} is trained on complete tabular datasets, ensuring that it can produce credible imputations for missing entries without undermining the authenticity of the existing data.
Innovatively, it can be applied to various settings of Missing Completely At Random (MCAR) and Missing At Random (MAR).
To effectively handle the tabular features in DDPM, we tailor four tabular denoising networks, spanning MLP, ResNet, Transformer, and U-Net.
We also propose \texttt{Harmonization} to enhance coherence between observed and imputed data by infusing the data back and denoising them multiple times during the sampling stage. 
To enable efficient inference while maintaining imputation performance, we propose a refined non-Markovian sampling process that works along with \texttt{Harmonization}.
Empirical evaluations on seven diverse datasets underscore the prowess of \texttt{DiffImpute}. 
Specifically, when paired with the Transformer as the denoising network, it consistently outperforms its competitors, boasting an average ranking of 1.7 and the most minimal standard deviation.
In contrast, the next best method lags with a ranking of 2.8 and a standard deviation of 0.9.
The code is available at \url{https://github.com/Dendiiiii/DiffImpute}.
\end{abstract}
\section{Introduction}
\label{sec:intro}
Tabular data is ubiquitous and crucial for data management and decision-making in various domains.
However, missing values often compromise the utility of tabular data since most deep-learning methods require complete datasets, and the causes of it are inherently complex~\cite{tan_tensor_based_2013}.
%
%
To counter this, researchers employ imputation methods, which are categorized into single and multiple imputation~\cite{Rubin_1987_multiple}, to replace missing entries.
%
%
Single imputation, using techniques like mean and median imputation, can introduce bias by homogenizing missing entries with singular values, leading to a misrepresentation of the genuine data distribution~\cite{Roderick_2002_Statistical}. 
On the opposite spectrum, multiple imputation suggests a gamut of plausible values for missing entries, leveraging iterative methods~\cite{Raghunathan_2000_Multivariate,Buuren_2011_MICE} and deep generative models~\cite{gondara2018mida,nazabal2020handling}.
Yet, these methods come with strings attached. Iterative methods might strain computational resources and demand robust data assumptions. 
Deep generative models, such as Generative Adversarial Networks (GANs) and Variation AutoEncoders (VAEs), grapple with challenges like mode collapse and posterior distribution alignment~\cite{Kingma_2019, goodfellow2014generative}. 
Moreover, previous work~\cite{ouyang2023missdiff} delineated an innovative score-centric approach, grounded on the gradient of the log-density score function. 
Therefore, the landscape still lacks a simple but efficient denoising diffusion stratagem crafted explicitly for tabular data imputation.
%
In light of these challenges, we propose \texttt{DiffImpute}, a Denoising Diffusion Probabilistic Model (DDPM) specifically tailored for tabular data imputation.
Unlike GANs and VAEs which are confined to Missing Completely At Random (MCAR) settings~\cite{Jarrett_2022_HyperImputeGI}, the diffusion models can be applied to more generous settings like Missing At Random (MAR).
Drawing inspiration from the principles of image inpainting~\cite{lugmayr_repaint_2022}, our method first involves training the DDPM~\cite{ho_denoising_2020} on complete datasets. 
During inference, our method effectively replaces the missing entries within an observed dataset while preserving the integrity of the observed values.
\texttt{DiffImpute} addresses mode collapse challenges observed in GAN-based approaches~\cite{salimans2016improved, goodfellow2015distinguishability} by the stability and simplicity of our training and inference process. 
Additionally, \texttt{DiffImpute} improves traceability by incorporating Gaussian noise throughout the diffusion process, as opposed to the prevalent practice of zero-padding in VAE-based approaches~\cite{mattei_2019_miwae}. Correspondingly, we propose a novel \texttt{Time Step Tokenizer} to embed temporal order information into the denoising network. 
Based on this, we explore four different denoising network architectures, including MLP, ResNet, U-Net, and Transformer, to demonstrate the improvement of incorporating time information in the imputation process.
Additionally, we propose \texttt{Harmonization} to meticulously aligns the imputed entries in data-deficient regions with the observed datasets through iterative processes of diffusion and denoising. 
Lastly, inspired by Song \emph{et al}'s work~\cite{song_denoising_2022}, we introduces the \texttt{Impute-DDIM} to boost the imputation speed while keeping the imputation quality for the tabular data. Our major contributions are four-fold:
%
\begin{itemize}
    \item The \texttt{DiffImpute}, a method that trains a diffusion model on complete data under MCAR and MAR missing mechanisms. \texttt{DiffImpute} offers a more stable and simplified training and inference process compared to other generative approaches.
    %
    %
    %
    \item DDPM, originally developed for image data, is adapted for tabular data by introducing the \texttt{Time Step Tokenizer} to encode temporal order information. 
    This modification enables the customization of four tabular denoising network architectures: MLP, ResNet, Transformer, and U-Net in our experiment. 
    %
    %
    \item We also introduce \texttt{Harmonization} to enhance coherence between imputed and observed data during the sampling stage. 
    %
    %
    %
    \item The \texttt{Impute-DDIM} allows repetitive and condensed time step sequences inference speed-up while implementing \texttt{Harmonization}.
    
    %
\end{itemize}
Correspondingly, we conduct extensive experiments on seven tabular datasets which suggest Transformer as the denoising network demonstrates faster training and inference, along with state-of-the-art performance.

\section{Methods}
In this section, we elaborate on \texttt{DiffImpute} and unpack the four denoising network architectures correspondingly.
Specifically, \texttt{DiffImpute} encompasses two stages: 
(1) the training of a diffusion model using complete tabular data; 
(2) the imputation of missing data from observed values.

\begin{figure}[!t]
\centering
\makebox[0pt]{\includegraphics[width=1.0\linewidth]{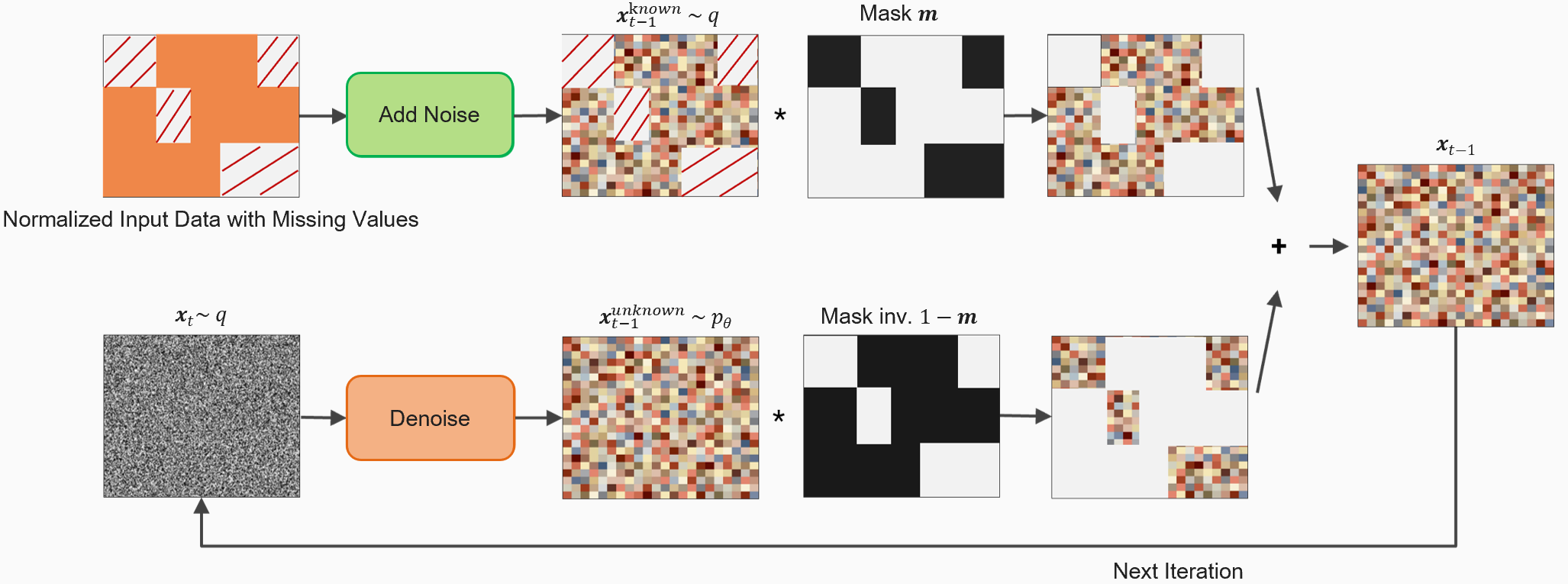}}
\caption{
    Schematic representation of \texttt{DiffImpute}. During inference, noisy data is extracted from known regions and supplemented with data imputed from the unknown region.
}
\label{fig1}
\end{figure}

\subsection{Training Stage of \texttt{DiffImpute}.}
The training phase of \texttt{DiffImpute} leverages DDPM on complete tabular data, denoted as $\mathbf{x}_0=(x_0^1,x_0^2,\cdots,x_0^k)\in \mathbb{R}^k$, where $k$ signifies the tabular data's dimensionality \textit{i.e.,} the number of columns. 
Within DDPM, Gaussian noise $\boldsymbol{\epsilon}$ is introduced to drive the transition from input $\mathbf{x}_0$ to distorted latent feature $\mathbf{x}_t$ across a span of $t$ time steps~\cite{ho_denoising_2020}. 
Then, the objective during the training of \texttt{DiffImpute} is to adeptly approximate the authentic data distribution of the complete tabular set.
To accomplish this, a denoising network is trained to acutely predict the noise profile $\boldsymbol{\epsilon}$ that has been infused into $\mathbf{x}_t$. 
Specifically, we employ the smooth L1 loss function, motivated by the function's proficiency in discerning the discrepancies between the anticipated and the genuine noise~\cite{gokcesu_generalized_2021}. Training pseudo-code is in the appendix.

\textbf{Missing Data Imputation.}
In the sampling stage, the observed tabular data $\mathbf{x}$ is categorized into two distinct regions~\cite{lugmayr_repaint_2022}. 
The ``known region'' defined by truly observed values is represented as $\mathbf{m} \odot \mathbf{x}$, where $\mathbf{m}\in\{0,1\}^k$ is a Boolean mask pinpointing the known data with $\odot$ denoting element-wise multiplication. 
Conversely, the ``unknown region'' harbors the missing values, denoted by $(1-\mathbf{m}) \odot \mathbf{x}$.
Imputation is executed by leveraging our trained denoising network within \texttt{DiffImpute}, symbolized as $f_{\theta}(\mathbf{x}_t, t)$.
This network focuses on the unknown region while retaining the values in the known sector, as illustrated in Fig.~\ref{fig1}. 
Diving deeper, this denoising network embarks on a stepwise refinement of the ``unknown region'', commencing with unadulterated Gaussian noise.
By tapping into the Markov Chain property of DDPM, Gaussian noise is injected at each time step $t$ to aid in sampling from the known region, $\mathbf{m} \odot \mathbf{x}$, depicted as follows:
\begin{eqnarray} \label{add_noise}
\mathbf{x}_{t-1}^{\text {known }}=\sqrt{\bar{\alpha}_{t-1}} \cdot\mathbf{x}_0+\sqrt{1-\bar{\alpha}_{t-1}}\cdot \boldsymbol{\epsilon},
\end{eqnarray}
where $\bar{\alpha}_{t-1}$ signifies the aggregate diffusion level or noise imposed on the initial input data $\mathbf{x}_0$ until time step $t-1$, and $\boldsymbol{\epsilon}\in\mathbb{R}^k$ is drawn from a Gaussian distribution.
However, for the unknown territories, the denoising network facilitates the sampling of progressively refined data with every backward step as follows:
\begin{eqnarray}
\mathbf{x}_{t-1}^{\text{unknown}}=\frac{1}{\sqrt{\alpha_t}}\cdot\left(\mathbf{x_t}-\frac{1-\alpha_{t}}{\sqrt{1-\bar{\alpha}_t}}\cdot f_{\theta}(\mathbf{x}_t, t)\right)+\sigma_t\cdot \boldsymbol{\epsilon},
\end{eqnarray}
where $\alpha_{t}$ represents the diffusion coefficient at time step $t$, ${\sigma}_t$ denotes the posterior standard deviation at time step $t$.
To synthesize the imputed data, the segments $\mathbf{x}_{t-1}^{\text{known}}$ and $\mathbf{x}_{t-1}^{\text{unknown}}$ are amalgamated based on their respective masks, yielding $\mathbf{x}_{t-1}$ at the $t-1$ time step: 
\begin{eqnarray}
    \mathbf{x}_{t-1} =\mathbf{m} \odot \mathbf{x}_{t-1}^{\text {known }}+(1-\mathbf{m}) \odot \mathbf{x}_{t-1}^{\text {unknown}}.
    \label{eq:output}
\end{eqnarray}
This procedure is reiterated in every reverse step until the final imputed data, $x_0$, emerges.

To further bolster the quality of our imputation, we propose \texttt{Harmonization} as a means to enhance the coherence between $\mathbf{x}_{t-1}^{\text{known}}$ and $\mathbf{x}_{t-1}^{\text{unknown}}$, thereby improving the quality of imputation. 
While \texttt{Harmonization} promises improved performance, extended time steps might inadvertently prolong inference runtime. 
To counterbalance this, we design \texttt{Impute-DDIM} to expedite the sampling process.

\begin{figure}[!t]
\centering
\makebox[0pt]{\includegraphics[width=1.0\linewidth]{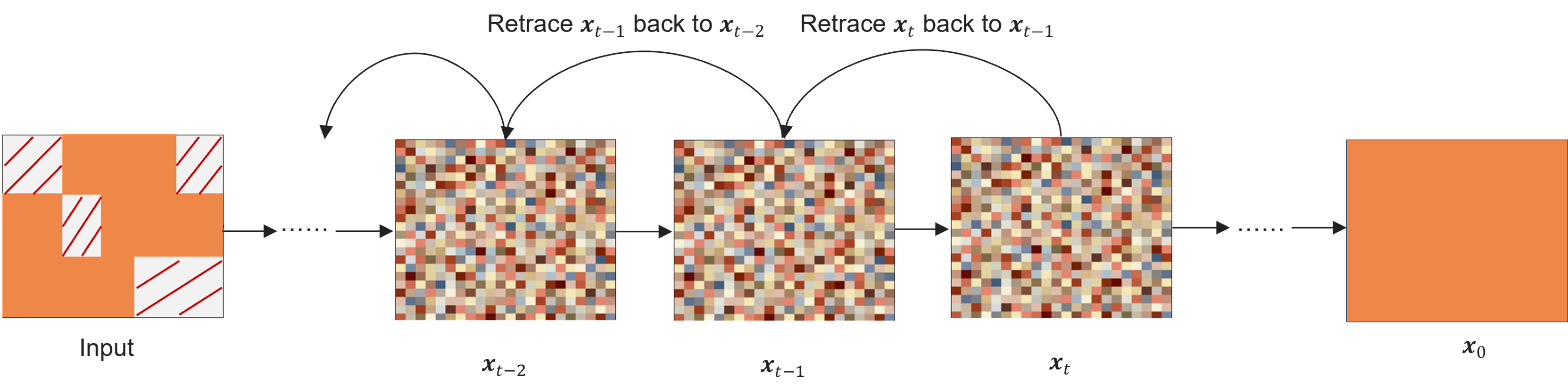}}
\caption{
    \texttt{Harmonization} process with \(t\) steps retrace back.
}
\label{fig_har}
\end{figure}

\textbf{\texttt{Harmonization}.} During the sampling of $\mathbf{x}_{t-1}$, we observed notable inconsistencies despite the model's active efforts to harmonize data at each interval~\cite{lugmayr_repaint_2022}, because the current methodologies are suboptimal in leveraging the generated components from the entire dataset. 
To overcome this challenge and enhance the consistency during the sampling stage, we introduce \texttt{Harmonization} (Fig. \ref{fig_har}) to retrace the output $\mathbf{x}_{t-1}$ in Eq.~(\ref{eq:output}) back by one or multiple steps to $\mathbf{x}_{t-1+j}$ by calculating $\sqrt{\alpha_{t}} \cdot\mathbf{x}_{t-1}+\sqrt{1-\alpha_{t}} \cdot\boldsymbol{\epsilon}$ incrementally $j$ times, where $j\ge1$ represents the number of steps retraced. 
For instance, $j = 1$ indicates a single-step retrace. 
It should be noted that as $j$ increases, the semantic richness of the data is amplified. 
However, a trade-off emerges as the run-time during the inference phase grows since the denoising network having to initiate its operation from the time step $t-1+j$. 

\textbf{\texttt{Impute-DDIM}.} To accelerate the sampling stage without compromising the benefits of \texttt{Harmonization}, we introduced \texttt{Impute-DDIM}, inspired by DDIM~\cite{song_denoising_2022}.
Central to its merit is the capacity to sample data at a substantially condensed time step $\tau$ for $\mathbf{x}_{t-1}^{\text{unknown}}$ during inference. 
By honing in on the forward procedure, specifically within the subset ${\mathbf{x}_{\tau1},\ldots, \mathbf{x}_{\tau S}}$ where $S\in\{1,\ldots,T\}$, the computational weight tied to inference is appreciably reduced.
Here, $\tau$ represents a sequentially increasing subset extracted from the range $\{1,\ldots,T\}$. 
It's worth noting that the derivation of $\mathbf{x}_{t-1}^{\text{unknown}}$ from its preceding time step $\mathbf{x}_{t}^{\text{unknown}}$ underwent a slight alteration:
\begin{eqnarray} 
\nonumber
\mathbf{x}_{t-1}^{\text{unknown}}=\sqrt{\alpha_{t-1}} \cdot\left(\frac{\mathbf{x}_t-\sqrt{1-\alpha_t} f_{\theta}(\mathbf{x}_t^{\text{unknown}}, t)}{\sqrt{\alpha_t}}\right)+ \\
\sqrt{1-\alpha_{t-1}-\sigma_t^2} \cdot f_{\theta}(\mathbf{x}_t^{\text{unknown}}, t)+\sigma_t \boldsymbol{\epsilon},
\end{eqnarray}

where $f_{\theta}(\mathbf{x}_t^{\text{unknown}}, t)$ refers to the predicted noise at the time step for the unknown region of $\mathbf{x}$ using a trained denoising model. The sampling pseudo-code is in the appendix.

\textbf{Overview.} In brief, the overall sampling process of \texttt{DiffImpute} starts at time step $T$ and backtracks to time step $0$, the initial step involves drawing the noise-laden observation $\mathbf{x}^{\text{known}}_{t-1}$ at time step $t-1$. 
This is followed by its multiplication with the mask $\mathbf{m}$ to derive the known section. 
For the unknown region $(1-\mathbf{m}) \odot \mathbf{x}$, $\mathbf{x}^{\text{unknown}}_{t-1}$ is sourced using the reverse procedure. 
The denoising network $f_{\theta}(\mathbf{x}_t, t)$ underpins this reverse modeling. 
Subsequently, the algorithm amalgamates the known and uncertain data facets to compute the imputed value at $t-1$. 
When the \texttt{Harmonization} setting with $j=1$ is active, a diffusion of the output $\mathbf{x}_{t-1}$ back to $\mathbf{x}_t$ is executed. 

\subsection{Denoising Network Architecture.}
\begin{figure}[!ht]
\centering
\includegraphics[width=\linewidth]{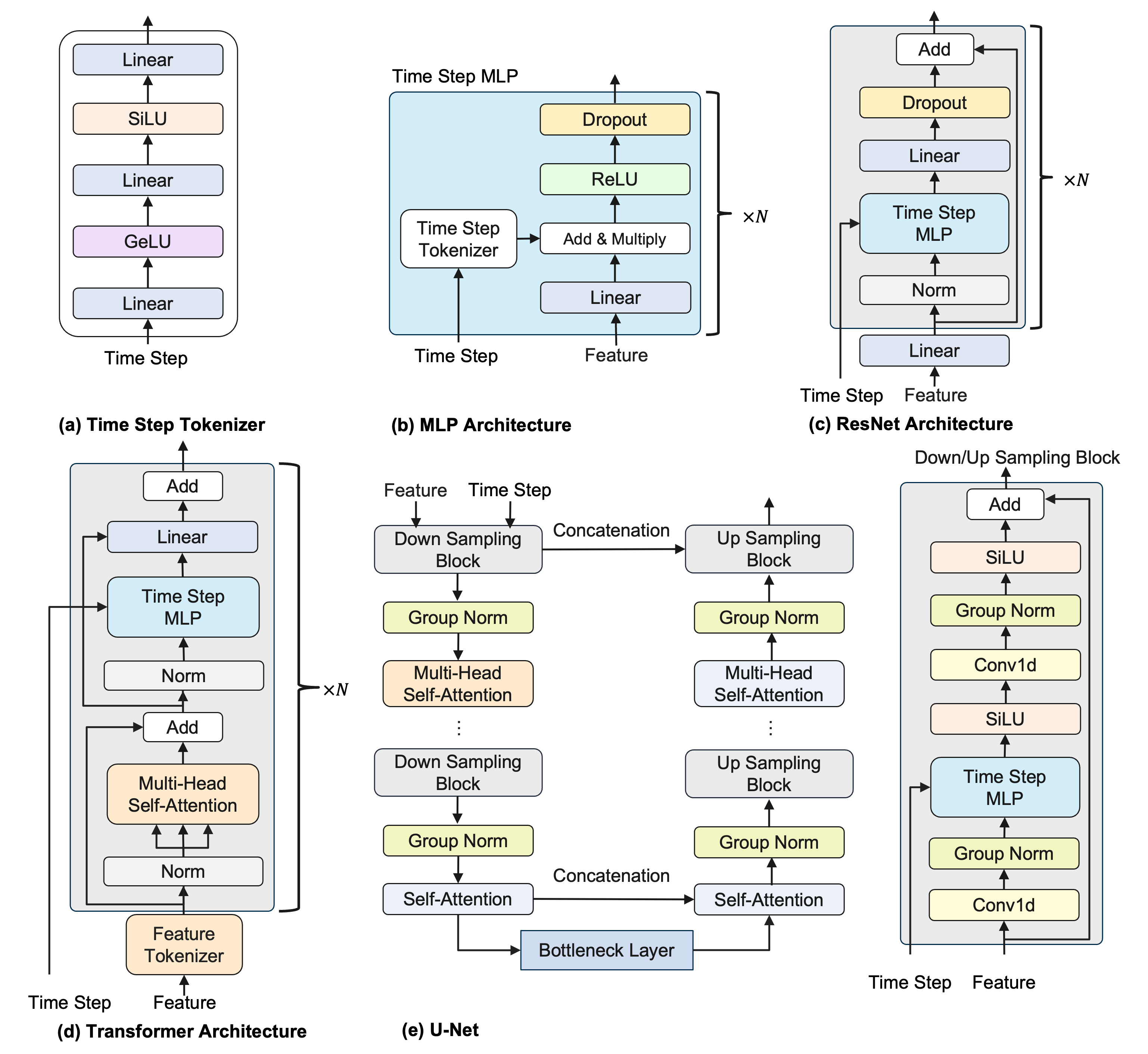}
\caption{Four types of denoising network architecture for tabular data. (a) Time Step Tokenzier, (b) MLP; (c) ResNet; (d) Transformer; (e) U-Net.}
\label{fig2}
\end{figure}

To obtain a denoising network tailored specifically for tabular data, we introduce the \texttt{Time Step Tokenizer} to encode temporal information into the denoising procedure. 
Building upon this foundational component, we have adapted four prominent denoising network architectures: MLP, ResNet, Transformer, and U-Net, as illustrated in Fig.~\ref{fig2}.

\textbf{\texttt{Time Step Tokenizer}.} 
Time step tokenizer is designed to encapsulate the information of time step  $t\in\mathbb{R}$, written as $\mathbf{t}_{\text{emb}} =\texttt{TimeStepTokenier}(t)\in\mathbb{R}^{2k}$. 
The tokenizer achieves this by formulating two distinct embeddings for scale and shift respectively, denoted as $ \mathbf{t}_{\text{emb}}=\texttt{Concat}(\mathbf{t}_{\text{emb}\_\text{scale}}, \mathbf{t}_{\text{emb}\_\text{shift}})\in \mathbb{R}^{2k}$, where $\texttt{Concat}$ signifies the concatenation of the two tensors $\mathbf{t}_{\text{emb}\_\text{scale}}$ and $\mathbf{t}_{\text{emb}\_\text{shift}}$ along the same dimension.
These learnable embeddings, $\mathbf{t}_{\text{emb}\_\text{scale}}$ and $\mathbf{t}_{\text{emb}\_\text{shift}}$, are inspired by the fixed sine and cosine transformations of $t$~\cite{vaswani_attention_2017}, defined as:
\begin{equation}
\begin{aligned}
\mathbf{t}_{\text{emb}} &= \texttt{Concat}(\mathbf{t}_{\text{emb}\_\text{scale}}, \mathbf{t}_{\text{emb}\_\text{shift}}) \nonumber \\ 
&= \texttt{Linear}(\texttt{SiLU}(\texttt{Linear}(\texttt{GELU}(\texttt{Linear}\\
&\quad([\mathbf{t}_{\text{scale}}, \mathbf{t}_{\text{shift}}]))))), \\
\mathbf{t}_{\text{scale}} &=\sin(t\cdot\exp{(\frac{-\log(10^4)}{k}\cdot [0,1,\ldots,k-1])}
),\\
\mathbf{t}_{\text{shift}} &=\cos(t\cdot\exp{(\frac{-\log(10^4)}{k}\cdot [0,1,\ldots,k-1])}
),
\end{aligned}
\end{equation}
where $\mathbf{t}_{\text{scale}}\wedge\mathbf{t}_{\text{shift}}\in \mathbb{R}^{k}$, \texttt{Linear} is a learnable linear layer, \texttt{SiLU} refers to the Sigmoid Linear Unit activation~\cite{elfwing2017sigmoidweighted}, and \texttt{GeLU} applies the Gaussian Error Linear Units function~\cite{hendrycks2023gaussian}.
Thus, each of the $ \mathbf{t}_{\text{emb}\_\text{scale}},\mathbf{t}_{\text{emb}\_\text{shift}}$ maintain the same dimension with $\mathbf{x}_t\in\mathbb{R}^k$.
To integrate these time step embeddings with the feature  $\mathbf{x}$, we compute the update as $\mathbf{x} \cdot (\mathbf{t}_{\text{emb}\_\text{scale}}+1) + \mathbf{t}_{\text{emb}\_\text{shift}}$, as depicted by ``Add \& Multiply'' in Fig.~\ref{fig2}(b).

\textbf{{\texttt{MLP}.}}
By leveraging the time step tokenizer, we can adapt the MLP~\cite{gorishniy_revisiting_2021} to serve as a denoising network by incorporating $t$ as an auxiliary input. 
Specifically, we introduce the time embedding, $\mathbf{t}_{\text{emb}}$, derived from the time step tokenizer, into a modified block named \texttt{TimeStepMLP}. 
This new block is an evolution of the traditional MLP Block. 
The architecture of this adaptation is depicted in Fig.~\ref{fig2}(b) and can be mathematically represented as
\begin{equation}
\begin{aligned} 
&\texttt{MLP}(\mathbf{x},\mathbf{t}_{\text{emb}}) = \texttt{Linear}(\texttt{TimeStepMLP}(\ldots(\\
&\quad\texttt{TimeStepMLP}(\mathbf{x},\mathbf{t}_{\text{emb}})))),
\end{aligned}
\end{equation}
where detailed \texttt{TimeStepMLP} expression can be found in the appendix.

\textbf{{\texttt{ResNet.}}} Building on the foundation of the \texttt{TimeStepMLP}, we then introduce a variant of ResNet~\cite{gorishniy_revisiting_2021} tailored for tabular DDPM. 
In this design, the \texttt{TimeStepMLP} block is seamlessly integrated into each ResNet block, as illustrated in Fig.~\ref{fig2}(c). 
We hypothesize that due to the depth of its representations, this ResNet variant will outperform the MLP-based models. 
Formally, the representation of our ResNet architecture is:
\begin{equation}
\begin{aligned} 
&\texttt{ResNet}(\mathbf{x}, \mathbf{t}_{\text{emb}}) = \texttt{Prediction}(\texttt{ResBlock}(\\
&\quad\ldots(\texttt{ResBlock}(\texttt{Linear}(\mathbf{x}),\mathbf{t}_{\text{emb}})))), \\
\end{aligned}
\end{equation}
where detailed \texttt{ResBlock} expression can be found in the appendix.

\textbf{{\texttt{Transformer.}}}
To further enhance our imputation capabilities, we adapt the Transformer architecture to tailor it explicitly for the tabular domain, as shown in Fig.~\ref{fig2}(d).
The transformer processes the feature and time step embeddings through a series of sequential layers, with each layer focusing on the feature level associated with a specific time stamp, $t$.
To elevate the representation of input tabular data, $\mathbf{x}$, we employ a learnable linear layer, aptly named \texttt{Feature Tokenizer}~\cite{gorishniy_revisiting_2021}.  
Then, for a given feature $\mathbf{x}=(x^1,\cdots,x^k)\in\mathbb{R}^k$, its embeddings are constructed as $\mathbf{x}_{\text{emb}}^k = \mathbf{b}^k + x^k \cdot \mathbf{W}^k \in\mathbb{R}^d$, where $\mathbf{b}^k\in\mathbb{R}^d$ is the learnable bias and $\mathbf{W}^k\in\mathbb{R}^d$ represents the learnable weight.
The aggregated embeddings are then represented as $\mathbf{x}_{\text{emb}}=[\mathbf{x}_{\text{emb}}^1,\ldots,\mathbf{x}_{\text{emb}}^k]\in\mathbb{R}^{k \times d}$, with $d$ being the feature embedding dimension. 
To capture global contexts and further enhance the model's performance on downstream tasks, we introduce the $[\textbf{\texttt{CLS}}]\in\mathbb{R}^d$ token~\cite{devlin2019bert}. 
This token is concatenated with the embedding matrix $\mathbf{x}_{\text{emb}}$, resulting in $\texttt{Concat}([\textbf{\texttt{CLS}}],\mathbf{x}_{\text{emb}})\in\mathbb{R}^{(k+1)\times d}$.
The architecture can be mathematically described as:
\begin{equation}
\begin{aligned} 
&\texttt{Transformer}(\mathbf{x}, \mathbf{t}_{\text{emb}}) = \texttt{Prediction}(\\
&\quad\texttt{TransBlock}(\ldots(\texttt{TransBlock}(\texttt{Concat}(\\
&\quad[\textbf{\texttt{CLS}}],\texttt{FeatureTokenizer}(\mathbf{x})),\mathbf{t}_{\text{emb}}))))\\
\end{aligned}
\end{equation}
where detailed \texttt{TransBlock} expression can be found in the appendix.

\textbf{{\texttt{U-Net.}}}
U-Net~\cite{ronneberger_u-net_2015} has garnered significant acclaim in the domain of diffusion models. 
Historically, its prowess has been predominantly demonstrated in image and text sequence processing. 
This has inadvertently led to a dearth of U-Net architectures specifically fine-tuned for tabular data. 
Addressing this gap, we introduce a novel U-Net tailored for tabular data, integrating both an encoder and decoder, as illustrated in Fig.~\ref{fig2}(e). 
This design uniquely amalgamates a variant of \texttt{TimeStepMLP} and self-attention mechanisms, ensuring optimal performance for tabular data.
Mathematically, our U-Net is represented as:
\begin{equation}
\begin{aligned} 
&\texttt{UNet}(\mathbf{x}, \mathbf{t}_{\text{emb}}) = 
\texttt{Linear}(\texttt{DecoderBlock}( \cdots (\\
&\quad\texttt{DecoderBlock}((\texttt{BottleneckBlock}( \cdots(\\
&\quad\texttt{EncoderBlock}( \cdots\texttt{EncoderBlock}((\mathbf{x}, \mathbf{t}_{\text{emb}}))\\
&\quad))))))),\\
%
\end{aligned}
\end{equation}
where detailed \texttt{DecoderBlock} and \texttt{EncoderBlock} expressions can be found in the appendix.
%

\textbf{Denoising Network Formulation.} Consequently, the denoising network is formulated as $f_\theta(\mathbf{x},t) = \texttt{Network}(\mathbf{x},\texttt{TimeTokenizer}(t))$.
Here, \texttt{Network} can be any of the following architectures: \texttt{MLP}, \texttt{ResNet}, \texttt{Transformer}, or \texttt{U-Net}.

\section{Experiments}
\label{experiment}

\subsection{Dataset and Implementations.}

\textbf{{Dataset.}}
We leverage seven publicly accessible datasets, offering a diverse representation of domains. These datasets are: 
(1) California Housing (CA), real estate data~\cite{Kelley_Statistics_1997}; 
(2) Helena (HE) and (3) Jannis (JA) are both anonymized datasets~\cite{guyon_analysis_2019}; 
(4) Higgs (HI), simulated data of physical particles~\cite{baldi_nature_2014}, where we adopted the version housing 98K samples from the OpenML repository~\cite{Vanschoren_ACM_2013}; 
(5) ALOI (AL), an image-centric dataset~\cite{Geusebroek_IJCV_2005};
(6) Year (YE), dataset capturing audio features~\cite{Bertin_ISMIR_2011};
(7) Covertype (CO), it describes forest characteristics~\cite{Blackard_CEA_1999}. Preprocessing information is in the appendix.
%

\textbf{Evaluation Metrics.} To gauge the precision of imputed values, we manually induce random masks on the test set data. 
The randomness of the mask is characterized by a percentage $p_{\text{random}} \in \{10\%,\ldots,90\% \}$ for each row (MCAR) and column mask (MAR) number $p_{\text{col}}\in\{1,\ldots,4\}$. 
Three evaluative criteria have been established: (1) Mean Squared Error (MSE); (2) Pearson Correlation Coefficient; (3) Downstream Tasks Performance. Hyper-parameter information is in the appendix.

\subsection{Results.}
\begin{figure}[!t]
\centering
\makebox[0pt]{\includegraphics[width=1.0\linewidth]{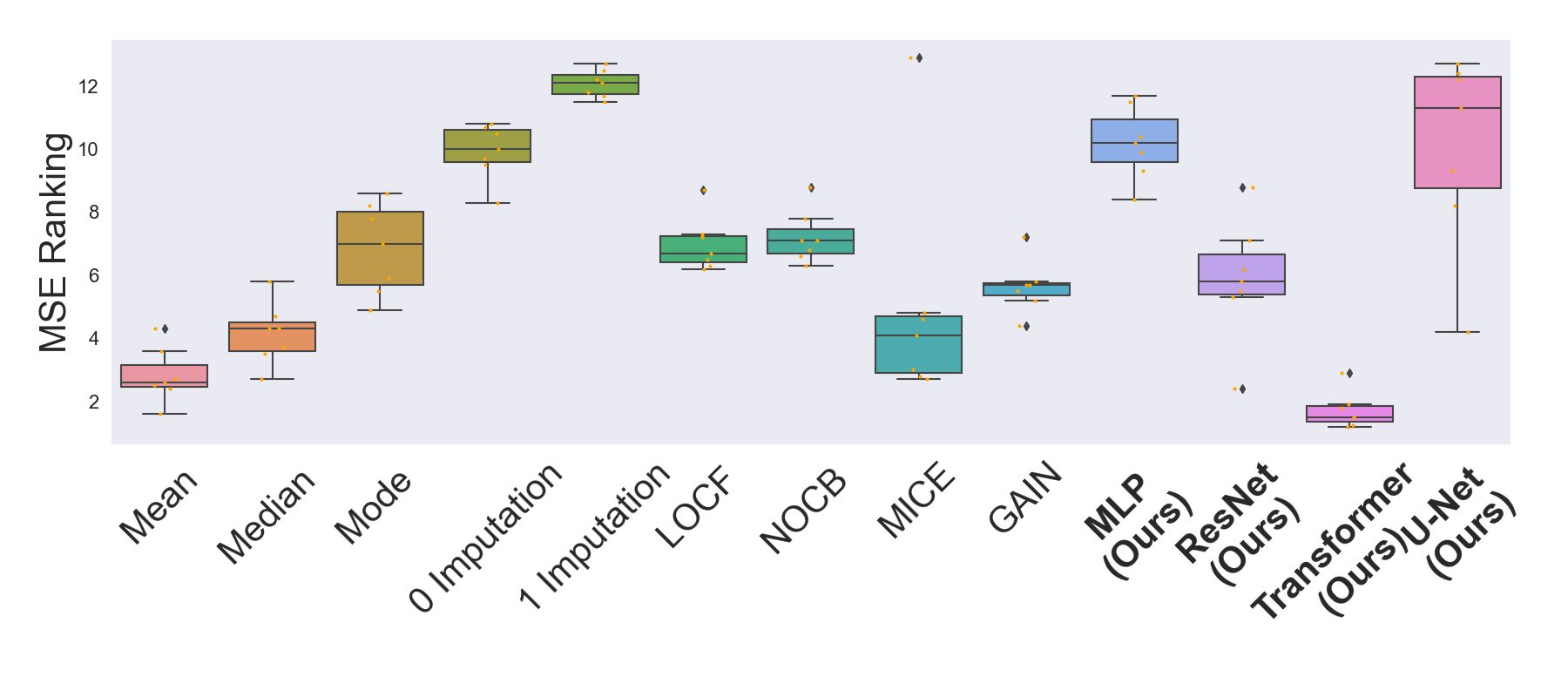}}
\caption{Imputation performance rankings of imputation methods in terms of MSE. The lower the better.}
\label{fig3}
\end{figure}

\begin{table}[!b]
\caption{Downstream task rank performance comparison using the imputed dataset. For each dataset, ranks are calculated by averaging rank across all missing percentage of the metrics (RMSE for regression problem and accuracy score for classification problem) for each dataset. The lower the better.}
\begin{center}
\resizebox{\linewidth}{!}{\begin{tabular}{l|lllllll|ll}
\toprule
\multicolumn{1}{c|}{\bf Imputation Methods}  &\multicolumn{1}{c}{\bf CA}  &\multicolumn{1}{c}{\bf HE}  &\multicolumn{1}{c}{\bf JA}  &\multicolumn{1}{c}{\bf HI}  &\multicolumn{1}{c}{\bf AL}  &\multicolumn{1}{c}{\bf YE}  &\multicolumn{1}{c|}{\bf CO} &\multicolumn{1}{c}{\bf Mean} &\multicolumn{1}{c}{\bf Std}\\
\hline
Mean Imputation         &3.9 &4.5 &6.5 &\textbf{1.8} &6.9 &3.9 &4.3 &4.5 &1.7 \\
Median Imputation       &5.2 &5.6 &6.9 &2.9 &3.7 &3.7 &2.9 &4.4 &1.5 \\
Mode Imputation         &6.6 &7.3 &5.8 &4.1 &5.5 &6.9 &6.2 &6.0 &1.1 \\
0 Imputation            &10.1 &9.2 &8.1 &7.6 &7.9 &8.0 &9.5 &8.7 &1.0 \\
1 Imputation            &10.7 &11.0 &10.2 &11.5 &11.3 &9.7 &10.6 &10.7 &0.6 \\
Last Observation Carrying Forfward (LOCF)         &8.2 &10.5 &10.1 &9.7 &11.5 &10.5 &8.5 &9.9 &1.2 \\
Next Observation Carrying Backward (NOCB)       &9.2 &12.1 &12.1 &12.0 &12.0 &12.2 &10.0 &11.4 &1.2 \\
MICE                    &2.8 &2.1 &3.0 &6.0 &2.8 &3.9 &9.6 &4.3 &2.6 \\
GAIN                    &4.9 &3.5 &4.0 &7.3 &4.9 &5.2 &7.7 &5.4 &1.6 \\
\hline
\textbf{\texttt{DiffImpute} w/ MLP}      &8.5 &8.5 &7.7 &8.5 &10.2 &8.9 &8.2 &8.7 &0.8 \\
\textbf{\texttt{DiffImpute} w/ ResNet }   &6.2 &5.1 &5.4 &6.6 &6.6 &6.1 &3.3 &5.6 &1.2 \\
\textbf{\texttt{DiffImpute} w/ Transformer}     &\textbf{1.5} &\textbf{2.2} &\textbf{2.4} &2.4 &\textbf{1.4} &\textbf{3.4} &\textbf{1.4} &2.1  &0.7 \\
\textbf{\texttt{DiffImpute} w/ U-Net}    &12.1 &9.0 &8.2 &10.1 &5.2 &6.1 &6.2 &8.1 &2.5 \\
\bottomrule
\end{tabular}}
\label{Tab:tb1}
\end{center}
\end{table}

\textbf{Comparison on Imputation Performance and Downstream Tasks.}
We start our evaluation by contrasting the performance of \texttt{DiffImpute} with a range of established single and iterative tabular imputation methods. 
As illustrated in Fig.~\ref{fig3} and Tab.~\ref{Tab:tb1}, when equipped with a Transformer as the denoising network, \texttt{DiffImpute} consistently surpasses its peers, both in terms of MSE that measures the imputation performance and downstream tasks on the imputed data. 
However, an anomaly is observed with the HI dataset. 
Its second-place performance can be traced back to the dataset's distinct characteristics, notably its dominant normal distributions and scant tail densities. 
This particular outcome accentuates the effectiveness of the mean imputation technique. 
Interestingly, mean imputation not only holds its own but even outperforms well-regarded methods such as MICE, GAIN, and \texttt{DiffImpute} with ResNet. 
While MICE does outshine mean imputation in specific datasets like HE, AL, and YE, its overall rank suffers due to variable performance on other datasets. 
Within the sphere of deep generative models, GAIN's performance parallels that of \texttt{DiffImpute} with ResNet, albeit at a slower inference speed. Detailed experiment results are in the appendix.

\textbf{Effect of Denoising Network Architectures.} Among the four denoising networks, the Transformer consistently stands out, marking its dominance in the tabular data domain. 
ResNets, on the other hand, serve as a robust baseline, delivering both impressive performance and swift inference speeds, thereby outperforming other models. 
The MLP and U-Net architectures face challenges in grasping sequential data, such as time step inputs. 
However, U-Net exhibits exceptional performance on the AL dataset, aligning with its foundational design for image data processing. 
Yet, its extended training and inference times make it a less optimal choice for tabular imputation. 
In summary, the Transformer within \texttt{DiffImpute} emerges as a leading solution.

\textbf{Ablation Study.}
%
We conducted an ablation study on the time embedding layers, \texttt{Harmonization}, and \texttt{Impute-DDIM} on the CA dataset to gain insights into their individual contributions.
We initiated our investigation by excluding the \texttt{time step tokenizer} from the denoising network. 
Interestingly, the impact on MSE performance was not uniform across models. 
This omission led to a noticeable decline in performance for the Transformer achitecture, with a 7.96\% drop in MSE performance and 6.28\% drop in the downstream task efficacy respectively. 
The U-Net and MLP architectures experienced significant improvements, recording a 63.81\% and 94.76\% enhancement in MSE, respectively.
Subsequently, we evaluated the impact of incorporating the \texttt{Harmonization} with $j=5$. The results, as detailed in Tab.~\ref{ablation_result}, highlight the performance boosts achieved by \texttt{Harmonization} across various architectures. 
To illustrate, when integrated into the \texttt{DiffImpute} with the MLP model, there was a remarkable 53.81\% increase in MSE and a 22.84\% improvement in downstream task performance for the CA dataset.
Lastly, we assessed the efficacy of \texttt{Impute-DDIM} in enhancing the inference speed, experimenting with different $\tau$ sampling steps, specifically $\tau \in \{10, 25, 50, 100, 250\}$ with $j=5$. 
As shown in Tab.~\ref{ddim_result}, when $\tau$ increases, the quality of imputation improves. 
Remarkably, with \texttt{Impute-DDIM} and a $\tau$ setting of 250, we managed to double the inference speed without compromising the MSE performance for both our MLP and Transformer models.
%
%
%
%
%
%
%


\begin{table}
\parbox{.48\linewidth}{
\centering
\caption{Ablation on \texttt{Time Step Tokenizer} (`TST`) and \texttt{Harmonization} (`H') with four denoising networks. 
We use the CA dataset and report the imputation performance in terms of MSE.}
\resizebox{\linewidth}{!}{\begin{tabular}{cc|cccc}
\toprule
\multicolumn{1}{c}{TST} &\multicolumn{1}{c|}{H} &\multicolumn{1}{c}{MLP} &\multicolumn{1}{c}{ResNet} &\multicolumn{1}{c}{ Transformer}&\multicolumn{1}{c}{U-Net}\\
\hline
    \ding{53} & \ding{53} &0.0212 & 0.0457 & 0.0210 & 0.0497 \\
    \checkmark & \ding{53} &0.0585 & 0.0498 & 0.0194 & 0.6831 \\
     \ding{53}&\checkmark & 0.0164 & 0.0199 & 0.0174 & 0.0184 \\
    \checkmark &\checkmark & 0.0268 & 0.0181 & 0.0191 & 4.2497 \\
\bottomrule
\end{tabular}}
\label{ablation_result}
}
\hfill
\parbox{.45\linewidth}{
\centering
\caption{
Ablation on \texttt{Impute-DDIM} with four denoising networks.
%
Note that when $\tau=500$, no \texttt{Impute-DDIM} is applied.
}
\resizebox{\linewidth}{!}{\begin{tabular}{c|cccc}
\toprule
\multicolumn{1}{c|}{$\tau$} &\multicolumn{1}{c}{MLP} &\multicolumn{1}{c}{ResNet} &\multicolumn{1}{c}{ Transformer}&\multicolumn{1}{c}{U-Net}\\
\hline
    10 & 0.2791 & 0.2574 & 0.2576 & 0.2741  \\
    25 & 0.2396 & 0.1892 & 0.1808 & 0.2274  \\
    50 & 0.1895 & 0.1164 & 0.0986 & 0.1727  \\
   100 & 0.1252 & 0.0525 & 0.0353 & 0.1145   \\
   250 & 0.0556 & 0.0240 & 0.0193 & 0.0795   \\
   500 & 0.0585 & 0.0498 & 0.0194 & 0.6831   \\
\bottomrule
\end{tabular}}
\label{ddim_result}
}
\end{table}

\section{Conclusion}
In this work, we introduce \texttt{DiffImpute}, a novel denoising diffusion model for imputing missing tabular data.
By seamlessly incorporating the \texttt{Time Step Tokenizer}, we have adapted four distinct denoising network architectures to enhance the capabilities of \texttt{DiffImpute}. 
Moreover, the amalgamation of the \texttt{Harmonization} technique and \texttt{Impute-DDIM} ensures that \texttt{DiffImpute} delivers superior performance without incurring extended sampling time. 
Our empirical evaluations, spanning seven diverse datasets, underscore the potential of \texttt{DiffImpute} as a foundational tool, poised to catalyze future innovations in the realm of tabular data imputation. 
One future direction is to further accelerate the sampling stage by distillation~\cite{salimans2022progressive}. 
Additionally, we envision broadening the scope of \texttt{DiffImpute} to cater to missing multimodal scenarios, given that latent space features can be intuitively treated as tabular data.

\bibliographystyle{IEEEbib}
\bibliography{icme2023template}
\section{Appendix}
\subsection{Related Works}
\label{sec:related_work}
\textbf{Missing Tabular Data Imputation.}
Most deep learning solutions often encounter challenges when dealing with missing data, while ensemble learning approaches tend to experience a decrease in predictive power due to the presence of missing data. Missing data originates from a myriad of sources including human error, equipment malfunction, and data loss~\cite{tan_tensor_based_2013} and basic single imputation methods such as mean and median imputation, while convenient, are notorious for introducing bias~\cite{Roderick_2002_Statistical}. To tackle this, the field has advanced toward more complex imputation strategies, broadly categorized into iterative and generative methods. 
Iterative techniques like Multiple Imputation by Chained Equations (MICE)~\cite{Buuren_2011_MICE} and MissForest~\cite{Stekhoven_2011} harness the conditional distributions between features to iteratively estimate missing values. 
On the other hand, generative models like GAIN~\cite{yoon_2018_gain} and MIWAE~\cite{mattei_2019_miwae} use deep function approximators to capture the joint probability distribution of features and impute missing values accordingly. 
Despite their sophistication, these approaches have limitations, including complicated optimization landscapes~\cite{Jarrett_2022_HyperImputeGI} and strong assumptions about data missingness patterns~\cite{li2019misgan, Seongwook_2020_GAMIN, nazabal2020handling}.

\textbf{Diffusion Models for Tabular Data.}
Generative models like GANs and VAEs have carved a niche in realms such as computer vision and natural language processing~\cite{rombach2022highresolution, chen2022analog}, but their foray into tabular data is still in its nascency. 
The reasons for this limited penetration are multifaceted, including the constrained sample sizes and the intricate task of integrating domain knowledge~\cite{liu2023goggle}.
Stepping into this milieu are diffusion models, which uniquely harness Markov chains to emulate the target distribution~\cite{sohldickstein_2015_deep, ho_denoising_2020}. 
Their distinctive edge is twofold: the capacity to spawn high-caliber samples~\cite{ho_denoising_2020} and the simplicity and robustness of their training paradigm~\cite{goodfellow2014generative, sohldickstein_2015_deep}. 
In fact, burgeoning literature indicates that DDPMs can potentially overshadow their generative counterparts~\cite{dhariwal2021diffusion, nichol2021improved}.
Yet, the potential of diffusion models in the tabular data context remains under-leveraged. 
A handful of pioneering studies have blazed the trail,~\cite{tashiro_csdi_2021} charted a course with a score-based diffusion model targeted at imputing lacunae in time series data, while~\cite{zheng_diffusion_2022} broadened this scope to envelop general tabular data imputation. 
Moreover, previous work~\cite{ouyang2023missdiff} delineated an innovative score-centric approach, grounded on the gradient of the log-density score function. 
However, the landscape still lacks a simple but efficient denoising diffusion stratagem crafted explicitly for tabular data imputation.

\section{Dataset Details}

\subsection{Dataset Descriptions and Statistics.}
We employed seven benchmark datasets in our experiments, the specifics of which are elaborated in Tab.~\ref{data_properties}. 
These datasets span two primary tasks, namely classification and regression. 
For evaluation, we adopt the mean square error (MSE) for regression tasks and the accuracy score for classification tasks. 
The data distribution for each dataset is structured such that 80\% is allocated for training and the remaining 20\% for testing.

\begin{table*}[ht!]
\caption{Statistics of the seven datasets used in our experiments. Regression tasks utilize mean square error (MSE) for evaluation, while classification tasks employ accuracy score.}
\label{data_properties}
\begin{center}

\begin{tabular}{l|cccccc}
\toprule
\multicolumn{1}{c|}{\bf Name}  &\multicolumn{1}{c}{\bf Abbr.}  &\multicolumn{1}{c}{\bf \# Train}  &\multicolumn{1}{c}{\bf \# Test}  &\multicolumn{1}{c}{\bf \# Num}  &\multicolumn{1}{c}{\bf Task Type}  &\multicolumn{1}{c}{\bf Batch Size} \\
\hline
California Housing     &CA &16512 &4128 &8 &Regression &256     \\
Helena             &HE &52156  &13040 &27 &Multiclass &256  \\
Jannis            &JA &66986 &16747 &54 &Multiclass &256  \\
Higgs Small        &HI &78439  &19610 &28 &Binclass &256  \\
ALOI         &AL &86400  &21600 &128 &Multiclass &256  \\
Year         &YE &463715  &51630 &90 &Regression &256  \\
Covtype      &CO &464809  &116203 &54 &Multiclass &256  \\
\bottomrule
\end{tabular}
\end{center}
\end{table*}

\subsection{Download Link.}
All datasets, formatted as \texttt{Numpy.darray}, are accessible for download from \url{https://www.dropbox.com/s/o53umyg6mn3zhxy/data.tar.gz?dl=1}. The source of these datasets is \url{https://github.com/Yura52/tabular-dl-revisiting-models}.

\subsection{Preprocessing.} 
For preprocessing, we standardized the numerical features and target values of each dataset using the \texttt{scikit-learn} library~\cite{pedregosa_2011_scikitlearn}. 
The standardization is based on the following equations:
\begin{equation} \label{min_max} 
\begin{aligned}
\mathbf{X}_{\text{std}} &= \frac{(\mathbf{X} - X_{\text{min}})}{X_{\text{max}}-X_{\text {min}}}, \\ 
\mathbf{X}_{\text{scaled}} &= \mathbf{X}_{\text{std}}\cdot(\text{max} - \text{min}) + \text{min}.
\end{aligned}
\end{equation}
This preprocessing is applied to all variables, excluding the classification labels $\mathbf{y}$ for datasets CA, HE, JA, HI, AL, and CO. 
The feature values are scaled to lie between 0 and 1, with min=0 and max=1. 
Then we maintain a consistent 80\% and 20\% train-test split across all datasets, enabling uniform evaluation. 

\section{Methodological Details}
This section elaborates on the details of the methodology.

\subsection{Detailed Formulation of of MLP.}
\begin{equation}
\begin{aligned} 
&\texttt{TimeStepMLP}(\mathbf{x}, \mathbf{t}_{\text{emb}}) = \texttt{Dropout}(\texttt{ReLU}(\\
&\quad\texttt{Linear}(\mathbf{x})\cdot (\mathbf{t}_{\text{emb}\_\text{scale}}+1) + \mathbf{t}_{\text{emb}\_\text{shift}})),\\
\end{aligned}
\end{equation}
where \texttt{Dropout} randomly zeroes some of the elements of the input tensor using samples from a Bernoulli distribution, and \texttt{ReLU} stands for the rectified linear unit function~\cite{agarap2019deep}.

\subsection{Detailed Formulation of of ResNet.}
\begin{equation}
\begin{aligned} 
&\texttt{ResBlock}(\mathbf{x}, \mathbf{t}_{\text{emb}}) = \mathbf{x} + \texttt{Dropout}(\texttt{Linear}(\\
&\quad\texttt{TimeStepMLP}(\texttt{BatchNorm}(\mathbf{x}),\mathbf{t}_{\text{emb}}))),\\
&\texttt{Prediction}(\mathbf{x}) = \texttt{Linear}(\texttt{ReLU}(\texttt{BatchNorm}\\
&\quad(\mathbf{x}))),
\end{aligned}
\end{equation}
where \texttt{BatchNorm} refers to the 1D batch normalization~\cite{Sergey_2015_batch}.

\subsection{Detailed Formulation of of Transformer.}
\begin{equation}
\begin{aligned} 
&\texttt{TransBlock}(\mathbf{x}, \mathbf{t}_{\text{emb}}) = \texttt{ResPreNorm}(\texttt{FFN}_{\mathbf{t}_{\text{emb}}},\\
&\quad\texttt{ResPreNorm}(\texttt{MHSA},\mathbf{x})),\\
&\texttt{ResPreNorm}(\texttt{Operator}, \mathbf{x}) = \mathbf{x} + \texttt{Dropout}(\\
&\quad\texttt{Operator}(\texttt{LayerNorm}(\mathbf{x}))),\\
&\texttt{FFN}_{\mathbf{t}_{\text{emb}}}(\mathbf{x}) = \texttt{Linear}(\texttt{TimeStepMLP}(\mathbf{x}, \mathbf{t}_\text{{emb}})),\\
&\texttt{Prediction}(\mathbf{x}) = \texttt{Linear}(\texttt{ReLU}(\texttt{LayerNorm}(\\
&\quad\mathbf{x}))), 
\end{aligned}
\end{equation}
where \texttt{LayerNorm} refers to layer normalization~\cite{ba2016layer}, while \texttt{MHSA} denotes the Multi-Head Self-Attention layer~\cite{vaswani_attention_2017} and we set \(n_{\text{heads}}\) = 8. 

\subsection{Detailed Formulation of of U-Net.} 
The following equations given the formal definition of the \texttt{DownSampleBlock}, \texttt{UpSampleBlock}, and the \texttt{BottleneckBlock}:
\begin{equation}
\begin{aligned} 
&\texttt{DecoderBlock}(\mathbf{x}, \mathbf{t}_{\text{emb}}) = \texttt{MHSA}(\texttt{ResBlock}_{\text{UNet}}(\\
&\quad\texttt{UpsampleBlock}(\mathbf{x}, \mathbf{t}_{\text{emb}}))),\\
&\texttt{EncoderBlock}(\mathbf{x}, \mathbf{t}_{\text{emb}}) = \texttt{MHSA}(\texttt{ResBlock}_{\text{UNet}}(\\
&\quad\texttt{DownsampleBlock}(\mathbf{x}, \mathbf{t}_{\text{emb}}))),\\
&\texttt{ResBlock}_{\text{UNet}}(\mathbf{x}) = \texttt{GroupNorm}(\mathbf{x}) + \mathbf{x}, \\
&\texttt{DownSampleBlock}(\mathbf{x}, \mathbf{t}_{\text{emb}}) = \texttt{SiLU}(\\
&\quad\texttt{GroupNorm}(\texttt{Conv1d}(\texttt{SiLU}(\texttt{TimeStepMLP}\\
&\quad(\texttt{GroupNorm}(\texttt{Conv1d}(\mathbf{x}))), \mathbf{t}_{\text{emb}})))) + \mathbf{x}\\
&\texttt{UpSampleBlock}(\texttt{Concat}(\mathbf{x},\texttt{DownSampleBlock}\\
&\quad(\mathbf{x}, \mathbf{t}_{\text{emb}})), \mathbf{t}_{\text{emb}}) = \texttt{SiLU}(\texttt{GroupNorm}(\texttt{Conv1d}(\\
&\quad\texttt{SiLU}(\texttt{TimeStepMLP}(\texttt{GroupNorm}(\texttt{Conv1d}(\\
&\quad\texttt{Concat}(\mathbf{x},\texttt{DownSampleBlock}(\mathbf{x}, \mathbf{t}_{\text{emb}}))))), \\
&\quad\mathbf{t}_{\text{emb}})))) +  \texttt{Concat}(\mathbf{x},\texttt{DownSampleBlock}(\mathbf{x}, \\
&\quad\mathbf{t}_{\text{emb}})) \\
&\texttt{BottleneckBlock}(\mathbf{x}) =  \texttt{MHSA}(\texttt{Res}_{\text{U-Net}}(\texttt{SiLU}(\\
&\quad\texttt{GroupNorm}(\texttt{Conv1d}(\texttt{SiLU}(\texttt{TimeStepMLP}(\\
&\quad\texttt{GroupNorm}(\texttt{Conv1d}(\mathbf{x}))), \mathbf{t}_{\text{emb}})))) + \mathbf{x})).
\end{aligned}
\end{equation}
where \texttt{GroupNorm} refers to Group Normalization~\cite{wu2018group}, while \texttt{Conv1d} signifies 1D convolution~\cite{kiranyaz20191d}. 
The \texttt{DownSampleBlock}, \texttt{UpSampleBlock}, and \texttt{BottleneckBlock} components, although distinct in their roles, share analogous layers with variations primarily in input and output channel sizes. 

Specifically, the \texttt{DownSampleBlock} commences with 64 channels, amplifying to 512, capturing intricate semantic information. 
In contrast, the \texttt{UpSampleBlock} initiates with 768 channels, tapering to 1, facilitating the restoration of feature map dimensions by harnessing the insights from the \texttt{DownSampleBlock}. 
This restoration is achieved through a skip connection, merging upsampled feature maps with their counterparts from the downsampling trajectory.
The \texttt{BottleneckBlock} serves as a conduit, preserving consistent input and output channel dimensions, and distilling pivotal features from the downsampling phase.

\subsection{Pseudo Code for the Training Stage.} The pseudo code of \texttt{DiffImpute} training is summarized in Alg.~\ref{alg:ap-train}.

\begin{algorithm}
	\caption{Pseudo code for the training stage of $\texttt{DiffImpute}$ on a complete dataset $\mathbf{x}$.}\label{alg:ap-train} 
	\begin{algorithmic}[1]
            \State \textbf{input:} Complete training data $\mathbf{x}\subseteq \mathbb{R}^k$, batch size $N$, time steps $T$, denoising network $f_{\theta}$, and smooth L1 loss scaling parameter $\beta_{\text{L1}} = 1$.
		\For {$epoch=1,2,\ldots$} 
			\For {sampled mini-batch $\{\mathbf{x}\}^N\}\in\mathcal{X}$}  
				\State $t \sim \operatorname{Uniform}(\{1, \ldots, T\})$  \Comment{Uniformly sample time steps for denoising model training}
                \State $\boldsymbol{\epsilon} \sim \mathcal{N}(\mathbf{0}, \mathbf{I})$ \Comment{Sample random noise from the Gaussian distribution}
                \State Compute the $\mathbf{x}_t$  based on $\mathbf{x}_0$ : $\sqrt{\bar{\alpha_t}} \mathbf{x}_0+\sqrt{1-\bar{\alpha_t}} \boldsymbol{\epsilon}$  \Comment{Diffuse $\mathbf{x}_0$ to the noisy data $\mathbf{x}_t$}
                \State Compute the predicted noise $\boldsymbol{\epsilon}_{\theta}=f(\mathbf{x}^i_t, t)$  
                \State Define the smooth L1 loss function $\mathcal{L}:=$  $\begin{cases}0.5\left(\boldsymbol{\epsilon}-\boldsymbol{\epsilon}_{\theta}\right)^2 / \beta_{\text{L1}}, & \text { if }\left|\boldsymbol{\epsilon}-\boldsymbol{\epsilon}_{\theta}\right|<\beta_{\text{L1}} \\ \left|\boldsymbol{\epsilon}-\boldsymbol{\epsilon}_{\theta}\right|-0.5 \cdot \beta_{\text{L1}} & \text { otherwise }\end{cases}$  \Comment{Calculate the loss between predicted noise $\boldsymbol{\epsilon}_{\theta}$ and ground truth noise $\boldsymbol{\epsilon}$}
                \State Update neural network \(f_{\theta}(\mathbf{x}_t, t)\) to minimize \(\mathcal{L}\) using AdamW optimizer.
			\EndFor
		\EndFor
            \State \textbf{return} denoising network \(f_{\theta}(\mathbf{x}_t, t)\)
	\end{algorithmic} 
\end{algorithm}

\subsection{Pseudo Code for the Sampling Stage.} The pseudo code of \texttt{DiffImpute} samping is summarized in Alg.~\ref{alg:1}.

\begin{algorithm}
	\caption{Pseudo code for the sampling stage of \texttt{DiffImpute} with \texttt{Harmonization}.}\label{alg:1}
	\begin{algorithmic}[1]
            \State \textbf{input:} Observed tabular data $\mathbf{x} \subseteq \mathbb{R}^k$, retraced step \(J\), precomputed noise variance \(\boldsymbol{\alpha}\), the Boolean mask for the known region \(\mathbf{m}\), time step \(T\), denoising network \(f_{\theta}(\mathbf{x}_t, t)\)
		\For {$t=T,\ldots,1$}  \Comment{Loop through every time step \(t\) reversely}
			\For {$j=1,\ldots,J$}  \Comment{\texttt{Harmonization} parameter: retraced steps}
                    \State $\boldsymbol{\epsilon} \sim \mathcal{N}(\mathbf{0}, \mathbf{I})$ if \(t>1\), else \(\boldsymbol{\epsilon}=0\) \Comment{Sampling random noise}
                    \State $\mathbf{x}^{\text{known}}_{t-1} = \sqrt{\bar{\alpha}_t}\cdot \mathbf{x}_0+\sqrt{1-\bar{\alpha}_t}\cdot \boldsymbol{\epsilon}$  \Comment{Calculate the noisy observation at time step \(t-1\)}
                    \State $\mathbf{x}^{\text{unknown}}_{t-1}=\frac{1}{\sqrt{\alpha_t}}\cdot\left(\mathbf{x}_t-\frac{1-\alpha_{t}}{\sqrt{1-\bar{\alpha}_t}}\cdot f_{\theta}(\mathbf{x}_t, t)\right)+\alpha_t\cdot\boldsymbol{\epsilon}$  \Comment{Sampling denoised data}
                    \State $\mathbf{x}_{t-1} = \mathbf{m} \cdot \mathbf{x}^{\text{known}}_{t-1} + (1-\mathbf{m}) \cdot \mathbf{x}^{\text{unknown}}_{t-1}$ \Comment{Combining known and unknown regions}
                    \If{\(j < J\) and \(t > 1\)} 
                        \State $\mathbf{x}_{t} = \sqrt{\alpha_{t}}\cdot \mathbf{x}_{t-1}+\sqrt{1-\alpha_{t}} \cdot\boldsymbol{\epsilon}$ \Comment{Diffuse \(\mathbf{x}_{t-1}\) back to \(\mathbf{x}_{t}\)}
                    \EndIf 
			\EndFor
		\EndFor
            \State \textbf{return} \(\mathbf{x}_0\)
	\end{algorithmic} 
\end{algorithm}

\subsection{Algorithms for \texttt{Impute-DDIM} Step Schedule.} Pseudo code for the \texttt{Impute-DDIM} skip type schedule function definition is depicted in code listing.~\ref{python_ddim}.
\begin{figure*}
\begin{mypython}[caption={\texttt{Impute-DDIM} skip type schedule function},label=python_ddim]
def skip_seq(num_timesteps, timesteps, skip_type="uniform"):
    if skip_type == "uniform":
        skip = num_timesteps // timesteps
        seq = range(0, num_timesteps, skip)
        return seq
    elif skip_type == "quad":
        seq = (
            np.linspace(
                0, np.sqrt(num_timesteps * 0.8), timesteps
            )
            ** 2
        )
        seq = [int(s) for s in list(seq)]
        return ddim_seq
    else:
        raise NotImplementedError
\end{mypython}
\end{figure*}

\subsection{Algorithms for \texttt{Harmonization} with \texttt{Impute-DDIM} Schedule.} Pseudo code for the \texttt{Harmonization} schedule function definition is illustrated in code listing.~\ref{python_harmonization}, where working with the \texttt{Impute-DDIM}.
The \texttt{ddim\_seq} argument is the output of the function of \texttt{skip\_seq} from the code listing.~\ref{python_ddim}.

\begin{figure*}
\begin{mypython}[caption={\texttt{Harmonization} schedule function definition},label=python_harmonization]

def get_schedule_jump_DDIM(ddim_seq, jump_length, jump_n_sample):
    jumps = {}
    for j in range(0, len(ddim_seq)-jump_length, jump_n_sample):
        jumps[ddim_seq[j]] = jump_n_sample - 1

    t = len(ddim_seq)
    ts = []

    while t >= 1:
        t = t-1
        ts.append(ddim_seq[t])

        if jumps.get(ddim_seq[t], 0) > 0:
            jumps[ddim_seq[t]] = jumps[ddim_seq[t]]-1
            for _ in range(jump_length):
                t = t + 1
                ts.append(ddim_seq[t])

    ts.append(-1)

    return ts
\end{mypython}
\end{figure*}

\subsection{Schematic Illustration.} 
\begin{figure*}[!ht]
\centering
\includegraphics[scale=0.40]{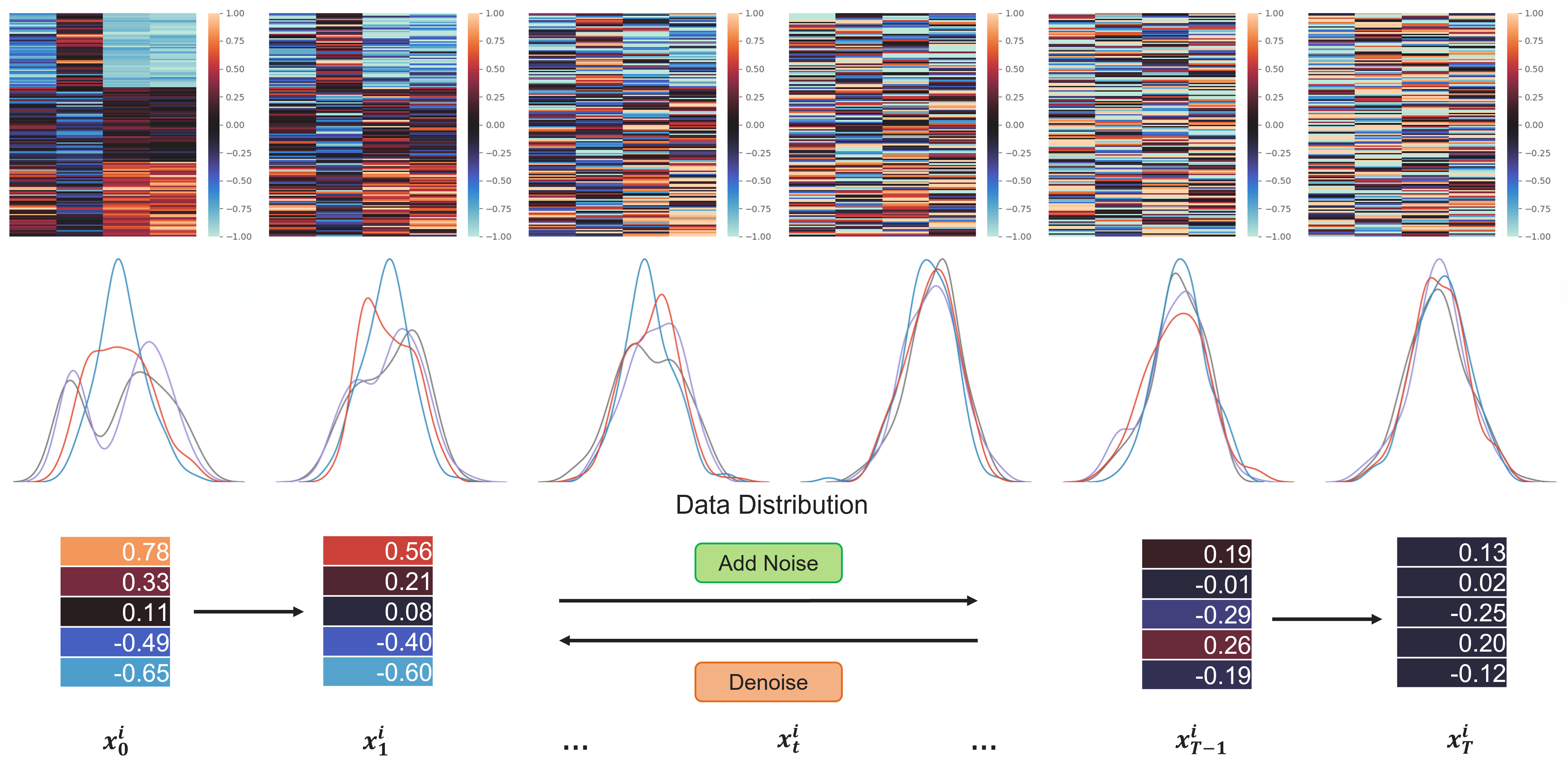}
\caption{This visualization captures the dual processes of noise addition and denoising. As noise is added, the data distribution converges towards a Gaussian shape, which is then reversed during the denoising phase.}\label{fig:ap-si}
\end{figure*}

To elucidate the diffusing and denoising process, we present a visual representation in Fig.~\ref{fig:ap-si}. This diagram captures the intricate dynamics of noise addition and subsequent denoising. Specifically, it illustrates how the data distribution gradually morphs into a Gaussian distribution during the noise addition phase and reverts during the denoising phase.

\section{Implementation Details}

\subsection{Hardware Platforms.} 
Our implementation followed a structured workflow:
\begin{itemize}
    \item We did the data preprocessing on any suitable hardware.
    \item Model training, inference, and evaluation were exclusively performed on an NVIDIA Tesla 3090 24GB GPU, boasting 35.6 TFLOPS. The software environment was consistent across all experiments, utilizing Python version 3.10.9 and Pytorch version 2.0.1+cu117.
\end{itemize}

\subsection{Training Settings.} 
While \texttt{DiffImpute} is trained on complete data, it performs imputation on test data, thereby leveraging insights from the complete dataset. To ensure a fair comparison, we also provided the training data as contextual information for all competing methods during their test data imputation.

\subsection{Hyper-parameters for \texttt{DiffImpute}.}
\texttt{DiffImpute} is trained over 20 epochs using batch sizes of 64.
Across all denoising network architectures and datasets, we employed an initial learning rate of $1e-3$, complemented by a learning rate decay of $1e-5$, optimized via AdamW.
A notable deviation is observed in the U-Net architecture for the YE dataset, which operates without feature learning rate decay and adopts an initial rate of 0.01.
During training, we designate the time step as $T_{\text{training}}=1000$.
Conversely, during the sampling phase, it's set to $T_{\text{sampling}=500}$, representing the reverse process steps.
The diffusion coefficient, $\alpha_{t}$, is derived from the forward process variance $\beta_{t}$, defined as $\alpha_{t} \coloneqq 1 - \beta_{t}$.
We adopt the $\beta_{t}$ schedule from a cosine schedule \cite{nichol2021improved}.
The posterior variance calculation follows: $\sigma_t = \frac{1-\bar{\alpha}_{t-1}}{1-\bar{\alpha}_t}\cdot\beta_t$~\cite{ho_denoising_2020}. 
For the \texttt{Impute-DDIM} acceleration, we partition the sampling step $T_{\text{sampling}}$ by a condensed time step $S$, uniformly distributing $T_{\text{sampling}}$ across $S$ steps.
For clarity, we set $\eta=0$, resulting in $\sigma_{t} = 0$, where $\sigma_{t}(\eta)=\eta \sqrt{\left(1-\alpha_{t-1}\right) /\left(1-\alpha_{t}\right)} \sqrt{1-\alpha_{t} / \alpha_{t-1}}$ \cite{song_denoising_2022}.
\Cref{MLP_para,ResNet_para,U-NET_para,Transformer_para} describe the implementation and configuration details of the four denoising networks.

\begin{table}[!ht]
\caption{MLP model hyper-parameters as denoising network architecture in \texttt{DiffImpute}.}
\label{MLP_para}
\begin{center}
\resizebox{\linewidth}{!}{\begin{tabular}{l|ccccccc}
\toprule
\multicolumn{1}{c|}{\bf Imputation Methods}  &\multicolumn{1}{c}{\bf CA}  &\multicolumn{1}{c}{\bf HE}  &\multicolumn{1}{c}{\bf JA}  &\multicolumn{1}{c}{\bf HI}  &\multicolumn{1}{c}{\bf AL}  &\multicolumn{1}{c}{\bf YE}  &\multicolumn{1}{c|}{\bf CO} \\
\hline
\textbf{Layer count}          &3 &3 &3 &3 &3 &3 &3  \\
\textbf{Feature embedding size}        &/ &/ &/ &/ &/ &/ &/  \\
\textbf{Head count}         &8 &8 &8 &8 &8 &8 &8  \\
\textbf{Activation \& FFN size factor}         &(ReLU, /) &(ReLU, /) &(ReLU, /) &(ReLU, /) &(ReLU, /) &(ReLU, /) &(ReLU, /)  \\
\textbf{Attention dropout}             &0.2 &0.2 &0.2 &0.2 &0.2 &0.2 &0.2  \\
\textbf{FFN dropout}             &0.1 &0.1 &0.1 &0.1 &0.1 &0.1 &0.1  \\
\textbf{Residual droupout}             &0 &0 &0 &0 &0 &0 &0  \\
\textbf{Initialization}         &/ &/ &/ &/ &/ &/ &/ \\
\hline
\textbf{Parameter count}         &2376 &18,279 &65,718 &19,516 &65,718 &174,330 &65,718  \\
\hline
\textbf{Optimizer}      &AdamW &AdamW &AdamW &AdamW &AdamW &AdamW &AdamW  \\
\textbf{Learning rate}   &1e-3 &1e-3 &1e-3 &1e-3 &1e-3 &1e-3 &1e-3  \\
\textbf{Weight decay}    &1e-5 &1e-5 &1e-5 &1e-5 &1e-5 &1e-5  &1e-5   \\
\bottomrule
\end{tabular}}
\end{center}
\end{table}

\begin{table}[!ht]
\caption{ResNet model hyper-parameters as denoising network architecture in \texttt{DiffImpute}.}
\label{ResNet_para}
\begin{center}
\resizebox{\linewidth}{!}{\begin{tabular}{l|ccccccc}
\toprule
\multicolumn{1}{c|}{\bf Imputation Methods}  &\multicolumn{1}{c}{\bf CA}  &\multicolumn{1}{c}{\bf HE}  &\multicolumn{1}{c}{\bf JA}  &\multicolumn{1}{c}{\bf HI}  &\multicolumn{1}{c}{\bf AL}  &\multicolumn{1}{c}{\bf YE}  &\multicolumn{1}{c|}{\bf CO} \\
\hline
\textbf{Layer count}          &3 &3 &3 &3 &3 &3 &3  \\
\textbf{Feature embedding size}        &192 &4.5 &3.0 &4.3 &3.3 &2.0 &5.8  \\
\textbf{Head count}         &8 &8 &8 &8 &8 &8 &8  \\
\textbf{Activation \& FFN size factor}         &(ReLU, /) &(ReLU, /) &(ReLU, /) &(ReLU, /) &(ReLU, /) &(ReLU, /) &(ReLU, /) \\
\textbf{Attention dropout}             &0.2 &0.2 &0.2 &0.2 &0.2 &0.2 &0.2  \\
\textbf{FFN dropout}             &0.1 &0.1 &0.1 &0.1 &0.1 &0.1 &0.1  \\
\textbf{Residual droupout}             &0 &0 &0 &0 &0 &0 &0  \\
\textbf{Initialization}         &/ &/ &/ &/ &/ &/ &/ \\
\hline
\textbf{Parameter count}         &3784 &22,119 &73,014 &23,484 &73,014 &186,234 &73,014  \\
\hline
\textbf{Optimizer}      &AdamW &AdamW &AdamW &AdamW &AdamW &AdamW &AdamW  \\
\textbf{Learning rate}   &1e-3 &1e-3 &1e-3 &1e-3 &1e-3 &1e-3 &1e-3  \\
\textbf{Weight decay}    &1e-5 &1e-5 &1e-5 &1e-5 &1e-5 &1e-5  &1e-5   \\
\bottomrule
\end{tabular}}
\end{center}
\end{table}

\begin{table}[!ht]
\caption{U-Net model hyper-parameters as denoising network architecture in \texttt{DiffImpute}.}
\label{U-NET_para}
\begin{center}
\resizebox{\linewidth}{!}{\begin{tabular}{l|ccccccc}
\toprule
\multicolumn{1}{c|}{\bf Imputation Methods}  &\multicolumn{1}{c}{\bf CA}  &\multicolumn{1}{c}{\bf HE}  &\multicolumn{1}{c}{\bf JA}  &\multicolumn{1}{c}{\bf HI}  &\multicolumn{1}{c}{\bf AL}  &\multicolumn{1}{c}{\bf YE}  &\multicolumn{1}{c|}{\bf CO} \\
\hline
\textbf{Layer count}          &3 &3 &3 &3 &3 &3 &3  \\
\textbf{Feature embedding size}        &/ &/ &/ &/ &/ &/ &/  \\
\textbf{Head count}         &8 &8 &8 &8 &8 &8 &8  \\
\textbf{Activation \& FFN size factor}         &(SiLU, /) &(SiLU, /) &(SiLU) &(SiLU, /) &(SiLU, /) &(SiLU, /) &(SiLU, /)  \\
\textbf{Attention dropout}             &0.2 &0.2 &0.2 &0.2 &0.2 &0.2 &0.2  \\
\textbf{FFN dropout}             &0.1 &0.1 &0.1 &0.1 &0.1 &0.1 &0.1  \\
\textbf{Residual droupout}             &0 &0 &0 &0 &0 &0 &0  \\
\textbf{Initialization}         &/ &/ &/ &/ &/ &/ &/ \\
\hline
\textbf{Parameter count}         &5,284,664 &5,590,792 &6,051,898 &5,607,324 &6,051,898 &6,714,334 &6,051,898  \\
\hline
\textbf{Optimizer}      &AdamW &AdamW &AdamW &AdamW &AdamW &AdamW &AdamW  \\
\textbf{Learning rate}   &1e-3 &1e-3 &1e-3 &1e-3 &1e-3 &1e-3 &1e-3  \\
\textbf{Weight decay}    &1e-2 &1e-2 &1e-2 &1e-2 &1e-2 &1e-2&1e-2  \\
\bottomrule
\end{tabular}}
\end{center}
\end{table}

\begin{table}[!ht]
\caption{Transformer model hyper-parameters as denoising network architecture in \texttt{DiffImpute}.}
\label{Transformer_para}
\begin{center}
\resizebox{\linewidth}{!}{\begin{tabular}{l|ccccccc}
\toprule
\multicolumn{1}{c|}{\bf Imputation Methods}  &\multicolumn{1}{c}{\bf CA}  &\multicolumn{1}{c}{\bf HE}  &\multicolumn{1}{c}{\bf JA}  &\multicolumn{1}{c}{\bf HI}  &\multicolumn{1}{c}{\bf AL}  &\multicolumn{1}{c}{\bf YE}  &\multicolumn{1}{c|}{\bf CO} \\
\hline
\textbf{Layer count}          &3 &3 &3 &3 &3 &3 &3  \\
\textbf{Feature embedding size}        &192 &192 &192 &192 &192 &192 &192  \\
\textbf{Head count}         &8 &8 &8 &8 &8 &8 &8  \\
\textbf{Activation \& FFN size factor}         &(ReGLU, 4/3) &(ReGLU, 4/3) &(ReGLU, 4/3) &(ReGLU, 4/3) &(ReGLU, 4/3) &(ReGLU, 4/3) &(ReGLU, 4/3)  \\
\textbf{Attention dropout}             &0.2 &0.2 &0.2 &0.2 &0.2 &0.2 &0.2  \\
\textbf{FFN dropout}             &0.1 &0.1 &0.1 &0.1 &0.1 &0.1 &0.1  \\
\textbf{Residual droupout}             &0 &0 &0 &0 &0 &0 &0  \\
\textbf{Initialization}         &kaiming &kaiming &kaiming &kaiming &kaiming &kaiming &kaiming \\
\hline
\textbf{Parameter count}         &3,997,448 &4,008,411 &4,023,990 &4,008,988 &4,023,990 &4,044,762 &4,023,990  \\
\hline
\textbf{Optimizer}      &AdamW &AdamW &AdamW &AdamW &AdamW &AdamW &AdamW  \\
\textbf{Learning rate}   &1e-3 &1e-3 &1e-3 &1e-3 &1e-3 &1e-3 &1e-3  \\
\textbf{Weight decay}    &1e-5 &1e-5 &1e-5 &1e-5 &1e-5 &1e-5  &1e-5   \\
\bottomrule
\end{tabular}}
\end{center}
\end{table}

\subsection{Evaluation Metrics.}
To assess imputation performance, we employ the following metrics. 
We first denote the imputed data as $\hat{\mathbf{x}} \in \mathbb{R}^k$ and the ground truth as $\mathbf{x} \in \mathbb{R}^k$. 
Here, $\hat{x}i$ represents the $i\text{-th}$ imputed value, and $x_i$ is the corresponding $i\text{-th}$ ground truth value. We use $N_{\text{miss}}$ to signify the total number of missing values.

To mitigate potential biases from randomness during mask generation, we instantiate five distinct random seeds for each missing percentage. 
Given the inherent variability in data masking and diffusion inference, each random setting undergoes 25 inferences, arising from 5 unique data masks and 5 independent inferences per mask. For each mask generated using a unique random seed, the imputed data is multiplied by one-fifth for each inference, and the results are accumulated over five inferences. Subsequently, the sum of these accumulated results is employed to calculate the MSE for the particular generated mask. The final outcome for each mask setting is determined by averaging the five MSE results obtained from each generated mask from the corresponding random seed.

\begin{itemize}
    \item \textbf{Mean Squared Error (MSE):} This metric quantifies the average squared discrepancy between the imputed and actual values.
    \begin{equation}
        \text{MSE}(\mathbf{x}, \hat{\mathbf{x}}) = \frac{\sum_{i=0}^{N_{\text{miss}} - 1} (x_i - \hat{x}_i)^2}{N_{\text{miss}}}
    \end{equation}
    \item \textbf{Pearson Correlation Coefficient:} This evaluates the linear relationship between the actual and imputed values.
    \begin{equation*}
      \text{R}(x, \hat{x}) = \frac{ \sum_{i=0}^{N_{\text{miss}} - 1} ((x_i - \bar{x})\cdot(\hat{x_i} - \bar{\hat{x}})) }{ \sqrt{ \sum_{i=0}^{N_{\text{miss}} - 1} (x_i - \bar{x})^2} \cdot\sqrt{ \sum_{i=0}^{N_{\text{miss}} - 1} (\hat{x_i} - \bar{\hat{x}})^2} }
    \end{equation*}
    
    \item \textbf{Downstream Tasks Performance:} For evaluating the performance on downstream tasks, we consistently use the same training and test sets. Depending on the nature of the downstream task, we employ either the root mean squared error (RMSE) for regression or the accuracy score for classification.
        \begin{itemize}
        \item \textbf{RMSE:} For regression tasks, the RMSE metric is used, where $y_{i}$ and $\hat{y}_{i}$ denote the $i\text{-th}$ actual and predicted values, respectively, and $N$ is the total number of values, defined as:
        \begin{equation}
        \text{RMSE}(\mathbf{y}, \hat{\mathbf{y}})=\sqrt{\frac{\sum_{i=0}^{N-1}\left(y_i-\hat{y}_i\right)^2}{N}}
        \end{equation}
        \item \textbf{Accuracy Score:} For classification tasks, we utilize the accuracy score, as defined in the \texttt{Scikit-learn} library~\cite{pedregosa_2011_scikitlearn}. Here, $\mathds{1}_{[\hat{y_i}=y_i]}$ is an indicator function that returns 1 if the condition $\hat{x_i}=y_i$ holds true.
        \begin{equation}
        \text{Accuracy Score}(\mathbf{y}, \hat{\mathbf{y}}) = \frac{\sum_{i=0}^{N} \mathds{1}_{[\hat{x_i}=y_i]}}{N}
        \end{equation}
        \end{itemize}
\end{itemize}

\subsection{Compared Methods.} 
Our research endeavors to benchmark various imputation techniques and model architectures across a suite of seven datasets. 
It's crucial to note that we refrained from fine-tuning model parameters or employing model-agnostic deep learning enhancements like pretraining, additional loss functions, or data augmentation. 
Although these methods can potentially elevate model performance, our core objective remains to gauge the intrinsic efficacy of the diverse model architectures under uniform conditions. 
Below, we elaborate on a concise synopsis of the methods under comparison:
\begin{itemize}
\item \textbf{Mean Imputation}: Substitutes missing values with the feature's mean.
\item \textbf{Median Imputation}: Uses the median of available values for imputation.
\item \textbf{Mode Imputation}: Fills missing slots with the most frequent value.
\item \textbf{0 Imputation}: Directly replaces missing values with 0.
\item \textbf{1 Imputation}: Uses 1 as the replacement.
\item \textbf{LOCF Imputation}: Fills gaps with the last observed value.
\item \textbf{NOCB Imputation}: Uses the subsequent observed value for imputation.
\item \textbf{MICE (linear) Imputation}: Employs multiple imputations based on regularized linear regression~\cite{Buuren_2011_MICE}.
\item \textbf{GAIN Imputation}: Leverages Generative Adversarial Nets for imputation~\cite{yoon_2018_gain}.
\end{itemize}

\subsection{Hyper-parameters for Compared Methods.} Below, we detail the hyper-parameters of the compared methods used in our experiments:
\begin{itemize}
    \item \textbf{MICE:} We fix and do not tune the following hyper-parameters:
\begin{itemize}
    \item \(n_{imputations} = 1\)
    \item \(max_{iter} = 100\)
    \item \(initial_{strategy} = 0\)
    \item \(imputation_{order} = 0\)
    \item \(random_{state}\) is set to the current time.
\end{itemize}
\item \textbf{GAIN:} We fix and do not tune the following hyper-parameters:
\begin{itemize}
    \item \(batch_{size} = 256\)
    \item \(n_{epochs} = 1000\)
    \item \(hint_{rate} = 0.9\)
    \item \(loss_{alpha} = 10\)
\end{itemize}
\end{itemize}

\section{More Results}
\subsection{Imputation Performance in Terms of MSE.}
We present the mean squared error (MSE) results for imputed data, evaluated under various missingness mechanisms across our seven benchmark datasets.

\textbf{Random Mask.}
In this segment, we focus on the imputation performance under the random mask settings. This mechanism aligns with the Missing Completely At Random (MCAR). The results for each of the seven datasets are detailed in the subsequent tables, referenced as \Cref{tab:ap-rm-ca,tab:ap-rm-he,tab:ap-rm-ja,tab:ap-rm-hi,tab:ap-rm-al,tab:ap-rm-ye,tab:ap-rm-co}.

\begin{table}[!ht]
\caption{Imputation performance comparison in terms of random mask setting, \textit{i.e.} Missing Completely At Random (MCAR), on CA using MSE. Optimal results are highlighted in \textbf{bold}.}
\label{tab:ap-rm-ca}
\begin{center}
\resizebox{\linewidth}{!}{
\begin{tabular}{l|ccccccccc}
    \toprule
       \textbf{Imputation Methods} & {10\%} & {20\%} & {30\%} & {40\%} & {50\%} & {60\%} & {70\%}& {80\%}& {90\%}\\
\hline
    Mean Imputation & 0.0210 & 0.0212 & 0.0214 & 0.0212 & 0.0213 & \textbf{0.0212} & \textbf{0.0212} & \textbf{0.0213} & \textbf{0.0212} \\
    Median Imputation& 0.0254 & 0.0256 & 0.0257 & 0.0254 & 0.0257 & 0.0256 & 0.0256 & 0.0256 & 0.0256    \\
    Mode Imputation& 0.0689 & 0.0843 & 0.0683 & 0.0689 & 0.0683 & 0.0681 & 0.0533 & 0.0536 & 0.0534  \\
    0 Imputation& 0.1055 & 0.1054 & 0.1067 & 0.1070 & 0.1073 & 0.1072 &0.1070 & 0.1069 & 0.1069  \\
    1 Imputation& 0.6892 & 0.6896 & 0.6881 & 0.6874 & 0.6868 & 0.6871 & 0.6872 & 0.6875 & 0.6876   \\
    LOCF Imputation& 0.0422 & 0.0421 & 0.0418 & 0.0426 & 0.0421 & 0.0422 & 0.0425 & 0.0425 & 0.0426   \\
    NOCB Imputation& 0.0420 & 0.0436 & 0.0438 & 0.0425 & 0.0437 & 0.0432 & 0.0429 & 0.0430 & 0.0431   \\
    MICE (linear) & 0.0192 & 0.0230 & 0.0252 & 0.0270 & 0.0314 & 0.0333 & 0.0367 & 0.0376 &0.0400 \\
    GAIN & 0.0224 & 0.0232 & 0.0238 & 0.0290 & 0.0422 & 0.0532 & 0.0739 & 0.0907 & 0.1024   \\
    \hline
    \textbf{\texttt{DiffImpute} w/ MLP} & 0.0495 & 0.0526 & 0.0554 & 0.0582 & 0.0609 & 0.0639 & 0.0670 & 0.0701 & 0.0734  \\
    \textbf{\texttt{DiffImpute} w/ ResNet} & 0.0160 & 0.0171 & \textbf{0.0182} & 0.0196 & 0.0218 & 0.0254 & 0.0321 & 0.0449 & 0.0680  \\
    \textbf{\texttt{DiffImpute} w/ Transformer} & \textbf{0.0155} & \textbf{0.0170} & 0.0184 & \textbf{0.0195} & \textbf{0.0210} & 0.0221 & 0.0233 & 0.0246 & 0.0259  \\
    \textbf{\texttt{DiffImpute} w/ U-Net} & 0.6323 & 0.6540 & 0.6759 & 0.6895 & 0.7005 & 0.7077 & 0.7155 & 0.7206 & 0.7252  \\
    \bottomrule
  \end{tabular}
  }
\end{center}
\end{table}

\begin{table}[!ht]
\caption{Imputation performance comparison in terms of random mask setting, \textit{i.e.} Missing Completely At Random (MCAR), on HE using MSE. Optimal results are highlighted in \textbf{bold}.}
\label{tab:ap-rm-he}
\begin{center}
\resizebox{\linewidth}{!}{
\begin{tabular}{l|ccccccccc}
    \toprule
       \textbf{Imputation Methods} & {10\%} & {20\%} & {30\%} & {40\%} & {50\%} & {60\%} & {70\%}& {80\%}& {90\%}\\
\hline
    Mean Imputation& 0.0285 & 0.0285 & 0.0285 & 0.0284 & 0.0285 & 0.0285 & 0.0285& 0.0285 & \textbf{0.0285} \\
    Median Imputation& 0.0294 & 0.0294 & 0.0293 & 0.0293 & 0.0293 & 0.0293 & 0.0293 &0.0293 &0.0293    \\
    Mode Imputation& 0.0965 & 0.0960 & 0.0961 & 0.0960 & 0.0960 & 0.0960 & 0.0959 & 0.0958 & 0.0965  \\
    0 Imputation& 0.2547 & 0.2547 & 0.2544 & 0.2543 & 0.2542 & 0.2542 & 0.2542 & 0.2543 & 0.2543  \\
    1 Imputation& 0.3942 & 0.3943 & 0.3949 & 0.3950 & 0.3951 & 0.3951 & 0.3951 & 0.3950 & 0.3950   \\
    LOCF Imputation& 0.0570 & 0.0573 & 0.0573 & 0.0572 & 0.0571 & 0.0571 & 0.0570 & 0.0570 & 0.0570   \\
    NOCB Imputation & 0.0573 & 0.0574 & 0.0573 & 0.0572 & 0.0572 & 0.0572 & 0.0572 & 0.0572& 0.0572   \\
    MICE (linear) & 0.0125 & 0.0137 & 0.0156 & 0.0180 & 0.0205 & 0.0246 & 0.0296 & 0.0365 &0.0453 \\
    GAIN & 0.0227 & 0.0220 & 0.0241 & 0.0298 & 0.0544 & 0.1342 & 0.1550 & 0.1401 & 0.2536   \\
    \hline
    \textbf{\texttt{DiffImpute} w/ MLP} & 0.1116 & 0.1292 & 0.1482 & 0.1684 & 0.1902 & 0.2130 & 0.2371 & 0.2619 & 0.2876  \\
    \textbf{\texttt{DiffImpute} w/ ResNet} & 0.0122 & 0.0136 & 0.0154 & 0.0178 & 0.0218 & 0.0291 & 0.0442 & 0.0757 & 0.1381  \\
    \textbf{\texttt{DiffImpute} w/ Transformer} & \textbf{0.0088} & \textbf{0.0101} & \textbf{0.0117} & \textbf{0.0137} & \textbf{0.0162} & \textbf{0.0193} & \textbf{0.0227} & \textbf{0.0268} & 0.0314  \\
    \textbf{\texttt{DiffImpute} w/ U-Net} & 0.2464 & 0.2579 & 0.2705 & 0.2894 & 0.3026 & 0.3233 & 0.3475 & 0.3759 & 0.4098  \\
    \bottomrule
  \end{tabular}
  }
\end{center}
\end{table}

\begin{table}[!ht]
\caption{Imputation performance comparison in terms of random mask setting, \textit{i.e.} Missing Completely At Random (MCAR), on JA using MSE. Optimal results are highlighted in \textbf{bold}.}
\label{tab:ap-rm-ja}
\begin{center}
\resizebox{\linewidth}{!}{
\begin{tabular}{l|ccccccccc}
    \toprule
       \textbf{Imputation Methods} & {10\%} & {20\%} & {30\%} & {40\%} & {50\%} & {60\%} & {70\%}& {80\%}& {90\%}\\
\hline
    Mean Imputation& 0.0295 & 0.0295 & 0.0295 & 0.0295 & 0.0295 & 0.0295 & \textbf{0.0295}& \textbf{0.0295} & \textbf{0.0295} \\
    Median Imputation& 0.0303 & 0.0303 & 0.0303 & 0.0303 & 0.0303 & 0.0303 & 0.0303 & 0.0303 & 0.0303    \\
    Mode Imputation& 0.1003 & 0.0998 & 0.0998 & 0.1009 & 0.1011 & 0.1012 & 0.1013 & 0.1033 & 0.1005  \\
    0 Imputation& 0.2262 & 0.2263 & 0.2261 & 0.2262 & 0.2262 & 0.2263 & 0.2262 & 0.2262 & 0.2262  \\
    1 Imputation& 0.4131 & 0.4128 & 0.4132 & 0.4130 & 0.4129 & 0.4128 & 0.4129 & 0.4128 & 0.4128   \\
    LOCF Imputation& 0.0590 & 0.0589 & 0.0588 & 0.0589 & 0.0588 & 0.0588 & 0.0588 & 0.0588 & 0.0589   \\
    NOCB Imputation& 0.0589 & 0.0588 & 0.0589 & 0.0588 & 0.0589 & 0.0588 & 0.0588 & 0.0587 & 0.0586   \\
    MICE (linear) & 0.0366 & 0.0376 & 0.0384 & 0.0396 & 0.0410 & 0.0428 & 0.0456 & 0.0487 &0.0533 \\
    GAIN & 0.0407 & 0.0375 & 0.0436 & 0.0538 & 0.0733 & 0.1355 & 0.0904 & 0.0804 & 0.2039   \\
    \hline
    \textbf\texttt{DiffImpute} w/ {MLP} & 0.2158 & 0.2521 & 0.2880 & 0.3230 & 0.3569 & 0.3902 & 0.4229 & 0.4547 & 0.4857  \\
    \textbf{\texttt{DiffImpute} w/ ResNet} & 0.0242 & 0.0253 & 0.0270 & 0.0301 & 0.0358 & 0.0470 & 0.0679 & 0.1035 & 0.1599  \\
    \textbf{\texttt{DiffImpute} w/ Transformer} & \textbf{0.0233} & \textbf{0.0240} & \textbf{0.0249} & \textbf{0.0260} & \textbf{0.0273} & \textbf{0.0288} & 0.0305 & 0.0325 & 0.0347  \\
    \textbf{\texttt{DiffImpute} w/ U-Net} & 0.3720 & 0.4570 & 0.5631 & 0.6937 & 0.8462 & 1.016 & 1.1949 & 1.3656 & 1.4972  \\
    \bottomrule
  \end{tabular}
  }
\end{center}
\end{table}

\begin{table}[!ht]
\caption{Imputation performance comparison in terms of random mask setting, \textit{i.e.} Missing Completely At Random (MCAR), on HI using MSE. Optimal results are highlighted in \textbf{bold}.}
\label{tab:ap-rm-hi}
\begin{center}
\resizebox{\linewidth}{!}{
\begin{tabular}{l|ccccccccc}
    \toprule
       \textbf{Imputation Methods} & {10\%} & {20\%} & {30\%} & {40\%} & {50\%} & {60\%} & {70\%}& {80\%}& {90\%}\\
\hline
    Mean Imputation& 0.0570 & 0.0572 & \textbf{0.0570} & \textbf{0.0570} & \textbf{0.0570} & \textbf{0.0570} &\textbf{0.0569} & \textbf{0.0569} &  \textbf{0.0568} \\
    Median Imputation& 0.0681 & 0.0698 & 0.0711 & 0.0739 & 0.0724 & 0.0738 & 0.0737 &0.0737  &0.0739\\
    Mode Imputation& 0.1028 & 0.1013 & 0.1015 & 0.1014 & 0.1004 & 0.0995 & 0.0977 & 0.1019 & 0.0984  \\
    0 Imputation& 0.1844 & 0.1849 & 0.1847 & 0.1845 & 0.1845 & 0.1844 & 0.1845 & 0.1845 & 0.1845  \\
    1 Imputation& 0.5811 & 0.5807 & 0.5806 & 0.5808 & 0.5808 & 0.5808 & 0.5807 & 0.5807 & 0.5806   \\
    LOCF Imputation& 0.1135 & 0.1144 & 0.1139 & 0.1140 & 0.1140 & 0.1138 & 0.1135 & 0.1135 & 0.1135   \\
    NOCB Imputation& 0.1135 & 0.1140 & 0.1137 & 0.1138 & 0.1141 & 0.1137 & 0.1137 & 0.1137& 0.1137   \\
    MICE (linear) & 0.0838 & 0.0875 & 0.0913 & 0.0956 & 0.0990 & 0.1022 & 0.1059 & 0.1088 &0.1114 \\
    GAIN & 0.0867 & 0.0811 & 0.0806 & 0.0955 & 0.1026 & 0.1330 & 0.1381 & 0.1483 & 0.1778   \\
    \hline
    \textbf{\texttt{DiffImpute} w/ MLP} & 0.1523 & 0.1652 & 0.1781 & 0.1921 & 0.2071 & 0.2226 & 0.2384 & 0.2544 & 0.2708  \\
    \textbf{\texttt{DiffImpute} w/ ResNet} & \textbf{0.0545} & \textbf{0.0568} & 0.0592 & 0.0626 & 0.0680 & 0.0767 & 0.0911 & 0.1142 & 0.1501  \\
    \textbf{\texttt{DiffImpute} w/ Transformer} & 0.0594 & 0.0613 & 0.0625 & 0.0638 & 0.0650 & 0.0661 & 0.0670 & 0.0680 & 0.0688  \\
    \textbf{\texttt{DiffImpute} w/ U-Net} & 0.7151 & 0.7265 & 0.7362 & 0.7465 & 0.7575 & 0.7676 & 0.7777 & 0.7877 & 0.7975  \\
    \bottomrule
  \end{tabular}
  }
\end{center}
\end{table}

\begin{table}[!ht]
\caption{Imputation performance comparison in terms of random mask setting, \textit{i.e.} Missing Completely At Random (MCAR), on AL using MSE. Optimal results are highlighted in \textbf{bold}.}
\label{tab:ap-rm-al}
\begin{center}
\resizebox{\linewidth}{!}{
\begin{tabular}{l|ccccccccc}
    \toprule
       \textbf{Imputation Methods} & {10\%} & {20\%} & {30\%} & {40\%} & {50\%} & {60\%} & {70\%}& {80\%}& {90\%}\\
\hline
    Mean Imputation& 0.0175 & 0.0176 & 0.0175 & 0.0175 & 0.0175 & 0.0175 & 0.0175 & 0.0175 & \textbf{0.0175} \\
    Median Imputation& 0.0209 & 0.0209 & 0.0209 & 0.0209 & 0.0209 & 0.0209 & 0.0209 & 0.0209  &0.0209    \\
    Mode Imputation& 0.0255 & 0.0255 & 0.0255 & 0.0255 & 0.0255 & 0.0255 & 0.0255 & 0.0255 & 0.0255  \\
    0 Imputation& 0.0386 & 0.0386 & 0.0387 & 0.0386 & 0.0386 & 0.0386 & 0.0386 & 0.0386 & 0.0386  \\
    1 Imputation& 0.8833 & 0.8832 & 0.8831 & 0.8831 & 0.8833 & 0.8831 & 0.8832 & 0.8832 & 0.8832   \\
    LOCF Imputation& 0.0351 & 0.0351 & 0.0351 & 0.0351 & 0.0351 & 0.0351 & 0.0351 & 0.0351 & 0.0351   \\
    NOCB Imputation& 0.0351 & 0.0351 & 0.0351 & 0.0351 & 0.0351 & 0.0351 & 0.0351 & 0.0351 & 0.0351   \\
    MICE (linear) & 0.0065 & 0.0071 & 0.0079 & 0.0087 & 0.0099 & 0.0114 & 0.0136 & 0.0169 & 0.0224 \\
    GAIN & 0.0067 & 0.0079 & 0.0126 & 0.0154 & 0.0183 & 0.0203 & 0.0257 & 0.0302 & 0.0343  \\
    \hline
    \textbf{\texttt{DiffImpute} w/ MLP} & 0.2710 & 0.3174 & 0.3541 & 0.3857 & 0.4129 & 0.4370 & 0.4584 & 0.4776 & 0.4949  \\
    \textbf{\texttt{DiffImpute} w/ ResNet} & 0.0098 & 0.0105 & 0.0115 & 0.0133 & 0.0168 & 0.0229 & 0.0327 & 0.0469 & 0.0652  \\
    \textbf{\texttt{DiffImpute} w/ Transformer} & \textbf{0.0048} & \textbf{0.0054} & \textbf{0.0062} & \textbf{0.0071} & \textbf{0.0083} & \textbf{0.0100} & \textbf{0.0120} & \textbf{0.0146} & 0.0177  \\
    \textbf{\texttt{DiffImpute} w/ U-Net} & 0.0130 & 0.0139 & 0.0148 & 0.0158 & 0.0169 & 0.0182 & 0.0197 & 0.0217 & 0.0242  \\
    \bottomrule
  \end{tabular}
  }
\end{center}
\end{table}

\begin{table}[!ht]
\caption{Imputation performance comparison in terms of random mask setting, \textit{i.e.} Missing Completely At Random (MCAR), on YE using MSE. Optimal results are highlighted in \textbf{bold}.}
\label{tab:ap-rm-ye}
\begin{center}
\resizebox{\linewidth}{!}{
\begin{tabular}{l|ccccccccc}
    \toprule
       \textbf{Imputation Methods} & {10\%} & {20\%} & {30\%} & {40\%} & {50\%} & {60\%} & {70\%}& {80\%}& {90\%}\\
\hline
    Mean Imputation& 0.0009 & 0.0009 & 0.0009 & 0.0009 & 0.0009 & 0.0009 & 0.0009 & \textbf{0.0009} & \textbf{0.0009} \\
    Median Imputation& 0.0009 & 0.0009 & 0.0009 & 0.0009 & 0.0009 & 0.0009 & 0.0009 & 0.0009 & 0.0009    \\
    Mode Imputation& 0.0010 & 0.0010 & 0.0010 & 0.0010 & 0.0010 & 0.0010 & 0.0010 & 0.0010 & 0.0010  \\
    0 Imputation& 0.2251 & 0.2252 & 0.2252 & 0.2251 & 0.2252 & 0.2251 & 0.2251 & 0.2251 & 0.2252  \\
    1 Imputation& 0.3553 & 0.3552 & 0.3552 & 0.3552 & 0.3552 & 0.3552 & 0.3552 & 0.3552 & 0.3552   \\
    LOCF Imputation& 0.0018 & 0.0018 & 0.0018 & 0.0018 & 0.0018 & 0.0018 & 0.0018 & 0.0018 & 0.0018  \\
    NOCB Imputation& 0.0018 & 0.0018 & 0.0018 & 0.0018 & 0.0018 & 0.0018 & 0.0018 & 0.0018 & 0.0018   \\
    MICE (linear) & 0.0001 & 0.0002 & 0.0003 & 0.0004 & 0.0005 & 0.0007 & 0.0012 & 0.0014 & 0.0016 \\
    GAIN & 0.0641 & 0.0015 & 0.0019 & 0.0032 & 0.0128 & 0.0843 & 0.0148 & 0.1877 & 0.2252   \\
    \hline
    \textbf{\texttt{DiffImpute} w/ MLP} & 0.2011 & 0.2672 & 0.3260 & 0.3795 & 0.4282 & 0.4729 & 0.5143 & 0.5526 & 0.5885  \\
    \textbf{\texttt{DiffImpute} w/ ResNet} & 0.0013 & 0.0014 & 0.0016 & 0.0023 & 0.0048 & 0.0132 & 0.0346 & 0.0759 & 0.1440  \\
    \textbf{\texttt{DiffImpute} w/ Transformer} & \textbf{0.0006} & \textbf{0.0006} & \textbf{0.0006} & \textbf{0.0007} & \textbf{0.0007} & \textbf{0.0008} & \textbf{0.0008} & \textbf{0.0009} & 0.0010  \\
    \textbf{\texttt{DiffImpute} w/ U-Net} & 0.0036 & 0.0045 & 0.0057 & 0.0750 & 0.0106 & 0.0171 & 0.0313 & 0.0606 & 0.1161  \\
    \bottomrule
  \end{tabular}
  }
\end{center}
\end{table}

\begin{table}[!ht]
\caption{Imputation performance comparison in terms of random mask setting, \textit{i.e.} Missing Completely At Random (MCAR), on CO using MSE. Optimal results are highlighted in \textbf{bold}.}
\label{tab:ap-rm-co}
\begin{center}
\resizebox{\linewidth}{!}{
\begin{tabular}{l|ccccccccc}
    \toprule
       \textbf{Imputation Methods} & {10\%} & {20\%} & {30\%} & {40\%} & {50\%} & {60\%} & {70\%}& {80\%}& {90\%}\\
\hline
    Mean Imputation& 0.0333 & 0.0333 & 0.0333 & 0.0333 & 0.0333 & 0.0333 & \textbf{0.0333} & \textbf{0.0333} & \textbf{0.0333} \\
    Median Imputation& 0.0425 & 0.0424 & 0.0424 & 0.0424 & 0.0424 & 0.0424 & 0.0425 & 0.0425 & 0.0425    \\
    Mode Imputation& 0.0472 & 0.0471 & 0.0471 & 0.0471 & 0.0471 & 0.0471 & 0.0471 & 0.0471 & 0.0471  \\
    0 Imputation& 0.0909 & 0.0908 & 0.0908 & 0.0908 & 0.0908 & 0.0908 & 0.0908 & 0.0908 & 0.0908  \\
    1 Imputation& 0.8479 & 0.8481 & 0.8480 & 0.8480 & 0.8480 & 0.8480 & 0.8480 & 0.8480 & 0.8480   \\
    LOCF Imputation& 0.0666 & 0.0665 & 0.0666 & 0.0665 & 0.0666 & 0.0666 & 0.0666 & 0.0666 & 0.0666   \\
    NOCB Imputation& 0.0665 & 0.0664 & 0.0665 & 0.0664 & 0.0665 & 0.0665 & 0.0665 & 0.0666 & 0.0667   \\
    MICE (linear) & 29550 & 33301 & 880.73 & 7965.1 & 154.84 & 5.7013 & 0.46 & 8976.6 & 4148.3 \\
    GAIN & 0.0290 & 0.0292 & 0.0314 & 0.0405 & 0.0663 & 0.0768 & 0.0751 & 0.784 & 0.0893   \\
    \hline
    \textbf{\texttt{DiffImpute} w/ MLP} & 0.1555 & 0.1827 & 0.2100 & 0.2373 & 0.2642 & 0.2910 & 0.3180 & 0.3443 & 0.3701  \\
    \textbf{\texttt{DiffImpute} w/ ResNet} & 0.0200 & 0.0220 & 0.0243 & 0.0268 & 0.0300 & 0.0342 & 0.0368 & 0.0388 & 0.0407  \\
    \textbf{\texttt{DiffImpute} w/ Transformer} & \textbf{0.0176} & \textbf{0.0206} & \textbf{0.0235} & \textbf{0.0263} & \textbf{0.0290} & \textbf{0.0315} & 0.0345 & 0.0368 & 0.0390  \\
    \textbf{\texttt{DiffImpute} w/ U-Net} & 0.1098 & 0.1249 & 0.1447 & 0.1703 & 0.2047 & 0.2504 & 0.3122 & 0.3949 & 0.5069  \\
    \bottomrule
  \end{tabular}
  }
\end{center}
\end{table}

\textbf{Column Mask.}
In this segment, we assess the imputation performance under the column mask settings, aligning with the Missing At Random (MAR). The results for each of the seven datasets are detailed in the subsequent tables, referenced as \Cref{tab:ap-cm-ca,tab:ap-cm-he,tab:ap-cm-ja,tab:ap-cm-hi,tab:ap-cm-al,tab:ap-cm-ye,tab:ap-cm-co}. It's important to highlight that the NOCB imputation method is not suitable for the column mask setting, given the absence of a subsequent observation to utilize for imputation.

\begin{table}[!ht]
\caption{Imputation performance comparison in terms of column mask setting, \textit{i.e.} Missing  At Random (MAR), on CA using MSE. The best results are in \textbf{bold}.}
\label{tab:ap-cm-ca}
\begin{center}
\resizebox{\linewidth}{!}{
\begin{tabular}{l|cccc}
    \toprule
       \textbf{Imputation Methods} & {1} & {2} & {3} & {4}\\
\hline
    Mean Imputation& 0.0228 & 0.0245 & 0.0220 & 0.0137\\
    Median Imputation& 0.0273 & 0.0314 & 0.0266 & 0.0162 \\
    Mode Imputation& 0.0702 & 0.0544 & 0.0565 & 0.0281   \\
    0 Imputation & 0.1043 & 0.1214 & 0.0944 & 0.0818   \\
    1 Imputation & 0.7275 & 0.6591 & 0.7035 & 0.7407 \\
    LOCF Imputation & 0.0419 & 0.0453 & 0.0462 & 0.0246 \\
    NOCB Imputation & / & / & / & / \\
    MICE (linear) & 0.1012 & \textbf{0.0009} & \textbf{0.0111} & \textbf{0.0030}\\
    GAIN & 0.0610 & 0.0011 & 0.0062 & 0.0067 \\
    \hline
    \textbf{\texttt{DiffImpute} w/ MLP} & 0.0492 & 0.0586 & 0.0550 & 0.0469\\
    \textbf{\texttt{DiffImpute} w/ ResNet} & 0.0849 & 0.0225 & 0.0846 & 0.0902 \\
    \textbf{\texttt{DiffImpute} w/ Transformer} & \textbf{0.0184} & 0.0208 & 0.0173 & 0.0088 \\
    \textbf{\texttt{DiffImpute} w/ U-Net} & 0.6117 & 0.6188 & 0.7210 & 0.7079 \\
    \bottomrule
  \end{tabular}
  }
\end{center}
\end{table}

\begin{table}[!ht]
\caption{Imputation performance comparison in terms of column mask setting, \textit{i.e.} Missing  At Random (MAR), on HE using MSE. The best results are in \textbf{bold}.}
\label{tab:ap-cm-he}
\begin{center}
\resizebox{\linewidth}{!}{
\begin{tabular}{l|cccc}
    \toprule
       \textbf{Imputation Methods} & {1} & {2} & {3} & {4}\\
\hline
    Mean Imputation& 0.0225 & 0.0200 & 0.0351 & 0.0421 \\
    Median Imputation& 0.0231 & 0.0202 & 0.0360 & 0.0437 \\
    Mode Imputation& 0.1043 & 0.0239 & 0.1337 & 0.1733   \\
    0 Imputation& 0.2856 & 0.3412 & 0.2279 & 0.2301   \\
    1 Imputation& 0.3066 & 0.3333 & 0.4127 & 0.4674 \\
    LOCF Imputation& 0.0266 & 0.0316 & 0.0469 & 0.0504 \\
    NOCB Imputation& / & / & / & / \\
    MICE (linear) & 0.0015 & \textbf{0.0014} & 0.0207 & 0.0321\\
    GAIN & \textbf{0.0009} & 0.0024 & \textbf{0.0143} & 0.0286 \\
    \hline
    \textbf{\texttt{DiffImpute} w/ MLP} & 0.0983 & 0.1067 & 0.1234 & 0.1322\\
    \textbf{\texttt{DiffImpute} w/ ResNet} & 0.2633 & 0.3497 & 0.2640 & 0.2210 \\
    \textbf{\texttt{DiffImpute} w/ Transformer} & 0.0021 & 0.0149 & 0.0151 & \textbf{0.0149} \\
    \textbf{\texttt{DiffImpute} w/ U-Net} & 0.1920 & 0.3147 & 0.2874 & 0.2284 \\
    \bottomrule
  \end{tabular}}
\end{center}
\end{table}

\begin{table}[!ht]
\caption{Imputation performance comparison in terms of column mask setting, \textit{i.e.} Missing  At Random (MAR), on JA using MSE. The best results are in \textbf{bold}.}
\label{tab:ap-cm-ja}
\begin{center}\resizebox{\linewidth}{!}{
\begin{tabular}{l|cccc}
    \toprule
       \textbf{Imputation Methods} & {1} & {2} & {3} & {4}\\
\hline
    Mean Imputation& 0.0347 & 0.0279 & 0.0294 & 0.0379 \\
    Median Imputation& 0.0358 & 0.0281 & 0.0303 & 0.0389 \\
    Mode Imputation& 0.0550 & 0.0776 & 0.0880 & 0.0719   \\
    0 Imputation& 0.1891 & 0.1987 & 0.2332 & 0.3026   \\
    1 Imputation& 0.4190 & 0.4063 & 0.3930 & 0.3380 \\
    LOCF Imputation& 0.0582 & 0.0338 & 0.0631 & 0.0846 \\
    NOCB Imputation& / & / & / & / \\
    MICE (linear) & 0.0568 & 0.0561 & 0.0205 & 0.0272\\
    GAIN & 0.0303 & 0.0348 & 0.0164 & \textbf{0.0190} \\
    \hline
    \textbf{\texttt{DiffImpute} w/ MLP} & 0.2041 & 0.2014 & 0.1993 & 0.2253\\
    \textbf{\texttt{DiffImpute} w/ ResNet} & 0.2059 & 0.3091 & 0.2522 & 0.2880 \\
    \textbf{\texttt{DiffImpute} w/ Transformer} & \textbf{0.0299} & \textbf{0.0253} & \textbf{0.0114} & 0.0197 \\
    \textbf{\texttt{DiffImpute} w/ U-Net} & 0.3295 & 0.3412 & 0.3179 & 0.4265 \\
    \bottomrule
  \end{tabular}}
\end{center}
\end{table}

\begin{table}[!ht]
\caption{Imputation performance comparison in terms of column mask setting, \textit{i.e.} Missing At Random (MAR), on HI using MSE. The best results are in \textbf{bold}.}
\label{tab:ap-cm-hi}
\begin{center}\resizebox{\linewidth}{!}{
\begin{tabular}{l|cccc}
    \toprule
       \textbf{Imputation Methods} & {1} & {2} & {3} & {4}\\
\hline
    Mean Imputation& 0.0263 & 0.0492 & 0.0635 & 0.0534 \\
    Median Imputation& 0.0264 & 0.0701 & 0.0842 & 0.0536 \\
    Mode Imputation& 0.0707 & 0.0768 & 0.1187 & 0.1066   \\
    0 Imputation& 0.1307 & 0.1545 & 0.1905 & 0.1830   \\
    1 Imputation& 0.6354 & 0.6198 & 0.5846 & 0.5696 \\
    LOCF Imputation& 0.0664 & 0.1227 & 0.1460 & 0.1021 \\
    NOCB Imputation& / & / & / & / \\
    MICE (linear) & \textbf{0.0018} & 0.0043 & 0.0543 & 0.1110\\
    GAIN & \textbf{0.0018} & \textbf{0.0030} & \textbf{0.0314} & 0.0723 \\
    \hline
    \textbf{\texttt{DiffImpute} w/ MLP} & 0.1090 & 0.1334 & 0.1473 & 0.1437\\
    \textbf{\texttt{DiffImpute} w/ ResNet} & 0.0788 & 0.1824 & 0.1983 & 0.1881 \\
    \textbf{\texttt{DiffImpute} w/ Transformer} & 0.0301 & 0.0536 & 0.0676 & \textbf{0.0562} \\
    \textbf{\texttt{DiffImpute} w/ U-Net} & 0.6449 & 0.6786 & 0.7392 & 0.7265 \\
    \bottomrule
  \end{tabular}}
\end{center}
\end{table}

\begin{table}[!ht]
\caption{Imputation performance comparison in terms of column mask setting, \textit{i.e.} Missing  At Random (MAR), on AL using MSE. The best results are in \textbf{bold}.}
\label{tab:ap-cm-al}
\begin{center}\resizebox{\linewidth}{!}{
\begin{tabular}{l|cccc}
    \toprule
       \textbf{Imputation Methods} & {1} & {2} & {3} & {4}\\
\hline
    Mean Imputation& 0.0086 & 0.0214 & 0.0138 & 0.0185 \\
    Median Imputation& 0.0102 & 0.0265 & 0.0171 & 0.0227 \\
    Mode Imputation& 0.0102 & 0.0287 & 0.0171 & 0.0233   \\
    0 Imputation& 0.0102 & 0.0331 & 0.0171 & 0.0368   \\
    1 Imputation& 0.9433 & 0.8718 & 0.9328 & 0.8881 \\
    LOCF Imputation& 0.0102 & 0.1004 & 0.0509 & 0.0752 \\
    NOCB Imputation& / & / & / & / \\
    MICE (linear) & 0.0106 & 0.0208 & 0.0068 & 0.0101\\
    GAIN & 0.0099 & 0.0201 & 0.0058 & 0.0086 \\
    \hline
    \textbf{\texttt{DiffImpute} w/ MLP} & 0.1989 & 0.2244 & 0.2204 & 0.2348\\
    \textbf{\texttt{DiffImpute} w/ ResNet} & 0.0507 & 0.0476 & 0.0225 & 0.0791 \\
    \textbf{\texttt{DiffImpute} w/ Transformer} & \textbf{0.0029} & \textbf{0.0069} & \textbf{0.0037} & \textbf{0.0064} \\
    \textbf{\texttt{DiffImpute} w/ U-Net} & 0.0068 & 0.0057 & 0.0166 & 0.0142 \\
    \bottomrule
  \end{tabular}}
\end{center}
\end{table}

\begin{table}[!ht]
\caption{Imputation performance comparison in terms of column mask setting, \textit{i.e.} Missing  At Random (MAR), on YE using MSE. The best results are in \textbf{bold}.}
\label{tab:ap-cm-ye}
\begin{center}\resizebox{\linewidth}{!}{
\begin{tabular}{l|cccc}
    \toprule
       \textbf{Imputation Methods} & {1} & {2} & {3} & {4}\\
\hline
    Mean Imputation& 0.0007 & 0.0011 & 0.0010 & 0.0013 \\
    Median Imputation& 0.0007 & 0.0011 & 0.0010 & 0.0013 \\
    Mode Imputation& 0.0007 & 0.0014 & 0.0012 & 0.0015   \\
    0 Imputation& 0.3638 & 0.2321 & 0.2263 & 0.2119   \\
    1 Imputation& 0.2126 & 0.4028 & 0.3276 & 0.3496 \\
    LOCF Imputation& 0.0009 & 0.0016 & 0.0011 & 0.0017 \\
    NOCB Imputation& / & / & / & / \\
    MICE (linear) & 0.0008 & 0.0012 & 0.0004 & 0.0007\\
    GAIN & 0.0006 & 0.0019 & 0.0003 & 0.0011 \\
    \hline
    \textbf{\texttt{DiffImpute} w/ MLP} & 0.1465 & 0.1535 & 0.1629 & 0.1756\\
    \textbf{\texttt{DiffImpute} w/ ResNet} & 0.3666 & 0.3285 & 0.2516 & 0.2469 \\
    \textbf{\texttt{DiffImpute} w/ Transformer} & \textbf{0.0004} & \textbf{0.0007} & \textbf{0.0007} & \textbf{0.0009} \\
    \textbf{\texttt{DiffImpute} w/ U-Net} & 0.0013 & 0.0011 & 0.0015 & 0.0014 \\
    \bottomrule
  \end{tabular}}
\end{center}
\end{table}

\begin{table}[!ht]
\caption{Imputation performance comparison in terms of column mask setting, \textit{i.e.} Missing  At Random (MAR), on CO using MSE. The best results are in \textbf{bold}.}
\label{tab:ap-cm-co}
\begin{center}\resizebox{\linewidth}{!}{
\begin{tabular}{l|cccc}
    \toprule
       \textbf{Imputation Methods} & {1} & {2} & {3} & {4}\\
\hline
    Mean Imputation& 0.0378 & 0.0333 & 0.0323 & 0.0303 \\
    Median Imputation& 0.0409 & 0.0353 & 0.0341 & 0.0321 \\
    Mode Imputation& 0.0595 & 0.0394 & 0.0509 & 0.0566   \\
    0 Imputation& 0.0622 & 0.1206 & 0.0633 & 0.0759   \\
    1 Imputation& 0.8684 & 0.7813 & 0.8494 & 0.7801 \\
    LOCF Imputation& 0.0444 & 0.2175 & 0.1031 & 0.0499 \\
    NOCB Imputation& / & / & / & / \\
    MICE (linear) & NaN & NaN & NaN & NaN\\
    GAIN & NaN & NaN & NaN & NaN \\
    \hline
    \textbf{\texttt{DiffImpute} w/ MLP} & 0.1474 & 0.1451 & 0.1396 & 0.1430\\
    \textbf{\texttt{DiffImpute} w/ ResNet} & 0.0366 & 0.0322 & 0.0325 & 0.0292 \\
    \textbf{\texttt{DiffImpute} w/ Transformer} & \textbf{0.0245} & \textbf{0.0213} & \textbf{0.0253} & \textbf{0.0230} \\
    \textbf{\texttt{DiffImpute} w/ U-Net} & 0.1034 & 0.1022 & 0.0926 & 0.1111 \\
    \bottomrule
  \end{tabular}}
\end{center}
\end{table}

\textbf{Imputation Performance Rankings.}
In this segment, we showcase the consolidated rankings of imputation performance, measured by mean squared error (MSE), under various masking mechanisms, specifically Missing Completely At Random (MCAR) and Missing At Random (MAR). These rankings span seven datasets, as detailed in \Cref{MSE ranking,MSE ranking_MAR,MSE ranking_MCAR}. Within each dataset, the performance metrics are sorted to determine the rankings. The column labeled ``rank'' represents the average ranking across the different missingness settings.

\begin{table}[ht]
\caption{Overall imputation performance rankings under the random mask setting (MCAR) evaluated by MSE. \texttt{DiffImpute} with Transformer has the best overall performance. The \texttt{DiffImpute} with the Transformer architecture outperform other methods in six datasets out of seven datasets. The best results are in \textbf{bold}.}
\label{MSE ranking_MCAR}
\begin{center}
\resizebox{\linewidth}{!}{\begin{tabular}{l|ccccccc|cc}
\toprule
\multicolumn{1}{c|}{\bf Imputation Methods}  &\multicolumn{1}{c}{\bf CA}  &\multicolumn{1}{c}{\bf HE}  &\multicolumn{1}{c}{\bf JA}  &\multicolumn{1}{c}{\bf HI}  &\multicolumn{1}{c}{\bf AL}  &\multicolumn{1}{c}{\bf YE}  &\multicolumn{1}{c|}{\bf CO} &\multicolumn{1}{c}{\bf Mean} &\multicolumn{1}{c}{\bf Std} \\
\toprule
Mean Imputation         &2.1	&3.4	&2.0	&\textbf{1.2}	&4.3	&2.4	&2.6   &2.6   &0.9 \\
Median Imputation       &4.2	&4.6	&3.1	&3.6	&5.9	&2.4	&4.4   &4.0   &1.0 \\
Mode Imputation         &9.2	&8.4	&8.6	&5.8	&7.3	&4.6	&5.4   &7.0   &1.7 \\
0 Imputation            &11.0	&10.8	&10.1	&10.3	&10.8	&11.0	&9.0 &10.4 &0.7 \\
1 Imputation            &12.3	&12.9	&11.8	&12.0	&13.0	&12.3	&12.1    &12.3    &0.4 \\
LOCF Imputation         &6.4	&6.3	&6.7	&7.8	&8.8	&6.6	&7.3   &7.1   &0.8 \\
NOCB Imputation         &7.1	&7.1	&6.3	&7.8	&8.8	&6.6	&6.8   &7.2   &0.8 \\
MICE                    &4.4	&3.1	&4.7	&5.6	&2.2	&2.3	&12.9  &5.0  &3.4 \\
GAIN                    &7.2	&6.8	&7.1	&7.1	&5.1	&8.8	&5.7   &6.8   &1.1 \\
\hline
\textbf{\texttt{DiffImpute} w/ MLP}      &9.0	&10.2	&11.2	&10.7	&12.0	&12.6	&10.8 &10.9 &1.1 \\
\textbf{\texttt{DiffImpute} w/ ResNet}   &3.6	&3.9	&4.7	&3.7	&6.0	&7.8	&2.4   &4.6   &1.7 \\
\textbf{\texttt{DiffImpute} w/ Transformer}     &\textbf{1.7}	&\textbf{1.2}	&\textbf{1.7}	&2.4	&\textbf{1.1}	&\textbf{1.9}	&\textbf{1.3}   &1.6   &0.4\\
\textbf{\texttt{DiffImpute} w/ U-Net}      &12.7	&12.0	&12.9	&13.0	&4.7	&9.2	&10.2  &10.7  &2.8\\
\bottomrule
\end{tabular}}
\end{center}
\end{table}

\begin{table}[ht]
\caption{Overall imputation performance rankings under the column mask setting (MAR) evaluated by MSE. \texttt{DiffImpute} with Transformer has the best overall performance. The \texttt{DiffImpute} with the Transformer architecture outperform other methods in six datasets out of seven datasets. The best results are in \textbf{bold}.}
\label{MSE ranking_MAR}
\begin{center}
\resizebox{\linewidth}{!}{\begin{tabular}{l|ccccccc|cc}
\toprule
\multicolumn{1}{c|}{\bf Imputation Methods}  &\multicolumn{1}{c}{\bf CA}  &\multicolumn{1}{c}{\bf HE}  &\multicolumn{1}{c}{\bf JA}  &\multicolumn{1}{c}{\bf HI}  &\multicolumn{1}{c}{\bf AL}  &\multicolumn{1}{c}{\bf YE}  &\multicolumn{1}{c|}{\bf CO} &\multicolumn{1}{c}{\bf Mean} &\multicolumn{1}{c}{\bf Std} \\
\toprule
Mean Imputation         &3.8	&4.0	&3.3	&\textbf{2.5}	&4.3	&3.3	&2.8	&3.4	&0.6  \\
Median Imputation       &4.8	&5.0	&4.3	&4.0	&5.8	&3.3	&4.0	&4.4	&0.7  \\
Mode Imputation         &7.5	&7.5	&6.3	&6.3	&6.3	&5.8	&5.5	&6.4	&0.7  \\
0 Imputation            &9.8	&10.5	&8.8	&9.3	&6.8	&10.3	&6.8	&8.8	&1.4  \\
1 Imputation            &11.8	&11.5	&11.8	&11.0	&12.0	&11.5	&10.0	&11.4	&0.6  \\
LOCF Imputation         &5.8	&6.3	&6.0	&6.3	&8.5	&7.0	&6.8	&6.6	&0.9  \\
NOCB Imputation         &    /& /& /& /& /& /& /& /&/		 \\
MICE                    &3.3	&2.3	&4.5	&3.0	&4.8	&3.5	&NaN	&3.5	&0.9  \\
GAIN                    &2.8	&1.5	&2.5	&1.8	&2.8	&3.5	&NaN	&2.5	&0.7  \\
\hline
\textbf{\texttt{DiffImpute} w/ MLP}      &7.3	&7.3	&8.5	&8.3	&11.0	&9.0	&8.8	&8.5	&1.2  \\
\textbf{\texttt{DiffImpute} w/ ResNet}   &7.8	&10.3	&9.8	&9.5	&9.5	&11.3	&2.3	&8.8	&2.8  \\
\textbf{\texttt{DiffImpute} w/ Transformer}  &\textbf{2.5}	&\textbf{2.3}	&\textbf{1.3}	&4.0	&\textbf{1.3}	&\textbf{1.8}	&\textbf{1.0}	&2.0	&1.0\\
\textbf{\texttt{DiffImpute} w/ U-Net}      &11.3	&9.8	&11.3	&12.0	&3.0	&6.0	&7.3	&8.6	&3.1  \\
\bottomrule
\end{tabular}}
\end{center}
\end{table}

\begin{table}[ht]
\caption{Overall imputation performance rankings under the random mask (MCAR) and the column mask (MAR) settings  evaluated by MSE. \texttt{DiffImpute} with Transformer has the best overall performance. The \texttt{DiffImpute} with the Transformer architecture outperform other methods in six datasets out of seven datasets. The best results are in \textbf{bold}.}
\label{MSE ranking}
\begin{center}
\resizebox{\linewidth}{!}{\begin{tabular}{l|ccccccc|cc}
\toprule
\multicolumn{1}{c|}{\bf Imputation Methods}  &\multicolumn{1}{c}{\bf CA}  &\multicolumn{1}{c}{\bf HE}  &\multicolumn{1}{c}{\bf JA}  &\multicolumn{1}{c}{\bf HI}  &\multicolumn{1}{c}{\bf AL}  &\multicolumn{1}{c}{\bf YE}  &\multicolumn{1}{c|}{\bf CO} &\multicolumn{1}{c}{\bf Mean} &\multicolumn{1}{c}{\bf Std} \\
\toprule
Mean Imputation         &2.6	&3.6	&2.4	&\textbf{1.6}	&4.3	&2.7	&2.6 &2.8    &0.9 \\
Median Imputation       &4.4	&4.7	&3.5	&3.7	&5.8	&2.7	&4.3 &4.2    &1.0 \\
Mode Imputation         &8.7	&8.2	&7.8	&5.9	&7.0	&4.9	&5.5 &6.9    &1.4 \\
0 Imputation            &10.6	&10.7	&9.7	&10.0	&9.5	&10.8	&8.3&9.9 &0.9 \\
1 Imputation            &12.2	&12.5	&11.8	&11.7	&12.7	&12.1	&11.5 &12.0 &0.4 \\
LOCF Imputation         &6.2	&6.3	&6.5	&7.3	&8.7	&6.7	&7.2 &7.0    &0.9 \\
NOCB Imputation         &7.1	&7.1	&6.3	&7.8	&8.8	&6.6	&6.8 &7.2    &0.8 \\
MICE                    &4.1	&2.8	&4.6	&4.8	&3.0	&2.7	&12.9 &5.0   &3.6 \\
GAIN                    &5.8	&5.2	&5.7	&5.5	&4.4	&7.2	&5.7 &5.6    &0.8 \\
\hline
\textbf{\texttt{DiffImpute} w/ MLP}      &8.5	&9.3	&10.4	&9.9	&11.7	&11.5	&10.2 &10.2   &1.1\\
\textbf{\texttt{DiffImpute} w/ ResNet}   &4.8	&5.8	&6.2	&5.5	&7.1	&8.8	&2.4 &5.8   &2.0\\
\textbf{\texttt{DiffImpute} w/ Transformer}     &\textbf{1.9}	&\textbf{1.5}	&\textbf{1.5}	&2.9	&\textbf{1.2}	&\textbf{1.8}	&\textbf{1.2} &1.7   &0.6\\
\textbf{\texttt{DiffImpute} w/ U-Net}    &12.2	&11.3	&12.4	&12.7	&4.2	&8.2	&9.3 &10.0   &3.1\\
\bottomrule
\end{tabular}}
\end{center}
\end{table}

\begin{figure*}[ht]
\centering
\includegraphics[scale=0.45]{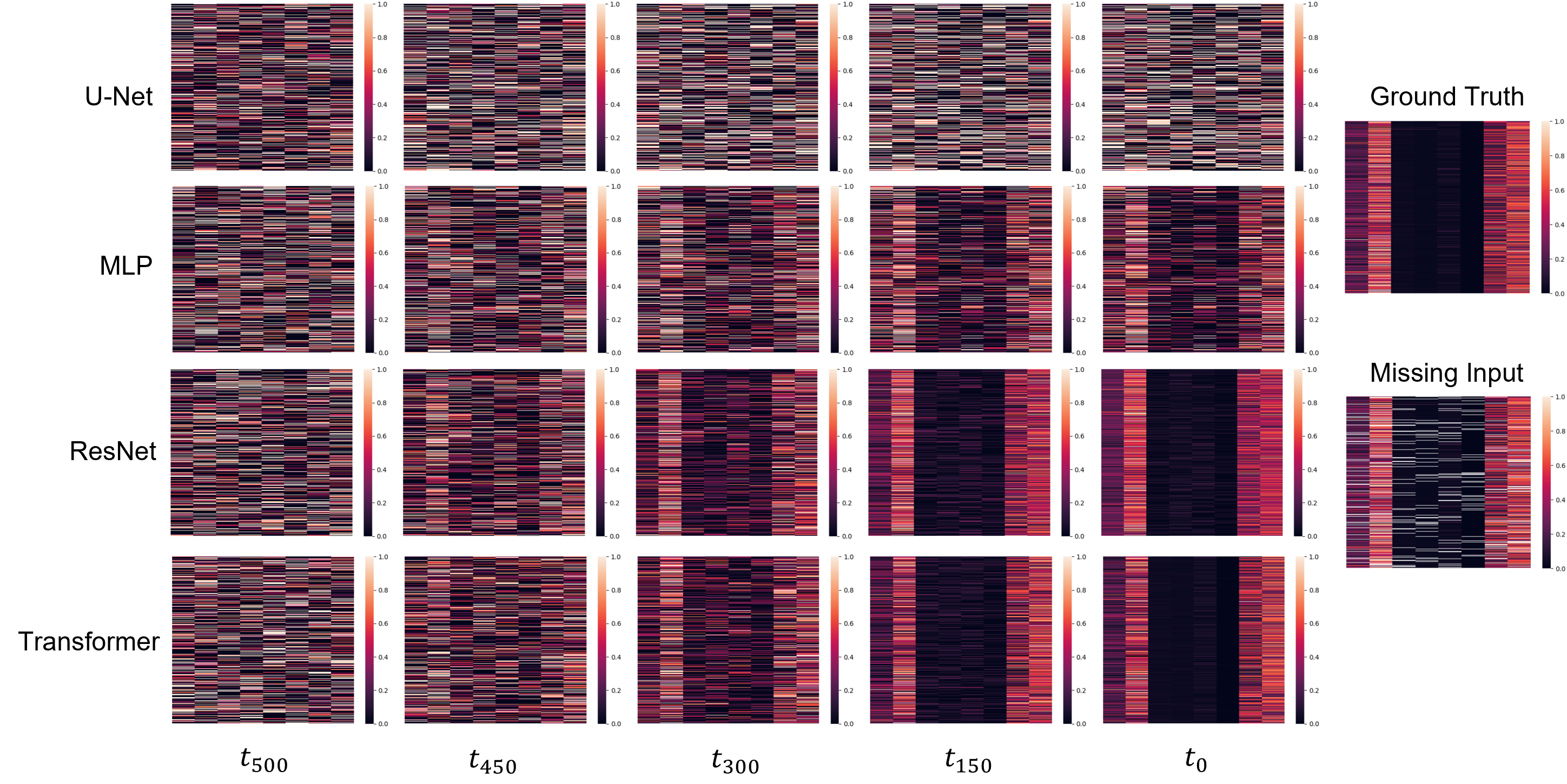}
\caption{Sampling Process on CA dataset at 90\% random mask of different model architectures.}
\label{fig5}
\end{figure*}

\textbf{Visualization of the Imputation Performance.}
Fig.~\ref{fig5} demonstrates the imputation process through the time steps of four denoising networks for the CA dataset with 90\% random mask. The ResNet and Transformer architectures utilized in \texttt{DiffImpute} exhibit superior imputation capability.

\subsection{Imputation Performance in Terms of Pearson Correlation.}
The following tables display the Pearson correlation performance between the ground truth data and the imputed data under different missingness mechanisms across seven datasets.

 \textbf{Random Mask.}
This section presents the evaluation of imputation performance using Pearson correlation under random mask settings, which correspond to the Missing Completely At Random (MCAR) mechanism, across seven datasets (\Cref{tab:pc-rm-ca,tab:pc-rm-he,tab:pc-rm-ja,tab:pc-rm-hi,tab:pc-rm-al,tab:pc-rm-ye,tab:pc-rm-co}).

\begin{table}[!ht]
\caption{Imputation performance comparison in terms of random mask setting, \textit{i.e.} Missing Completely At Random (MCAR), on CA using Pearson correlation. The best results are in \textbf{bold}.}
\label{tab:pc-rm-ca}
\begin{center}
\resizebox{\linewidth}{!}{
\begin{tabular}{l|ccccccccc}
    \toprule
       \textbf{Imputation Methods} & {10\%} & {20\%} & {30\%} & {40\%} & {50\%} & {60\%} & {70\%}& {80\%}& {90\%}\\
\hline
    Mean Imputation& 0.8138 & 0.8118 & 0.8126 & 0.8145 & 0.8142 & \textbf{0.8144} & \textbf{0.8139} & \textbf{0.8136} & \textbf{0.8140} \\
    Median Imputation& 0.7780 & 0.7763 & 0.7777 & 0.7788 & 0.7782 & 0.7789 & 0.7784 &0.7783  & 0.7787    \\
    Mode Imputation& 0.7162 & 0.6895 & 0.7154 & 0.6926 & 0.7154 & 0.7156 & 0.7383 & 0.7376 & 0.7385  \\
    LOCF Imputation& 0.6597 & 0.6649 & 0.6683 & 0.6620 & 0.6673 & 0.6654 & 0.6651 & 0.6640 & 0.6608  \\
    NOCB Imputation& 0.6649 & 0.6528 & 0.6538 & 0.6613 & 0.6555 & 0.6584 & 0.6608 & 0.6599 & 0.6571   \\
    MICE & 0.8418 & 0.8222 & 0.8012 & 0.7848 & 0.7527 & 0.7369 & 0.7125 & 0.7009 & 0.6832   \\
    GAIN & 0.8211 & 0.8160 & 0.7947 & 0.7480 & 0.6207 & 0.5500 & 0.4414 & 0.3555 & 0.4523   \\
    \hline
    \textbf{\texttt{DiffImpute} w/ MLP} & 0.5824 & 0.5649 & 0.5529 & 0.5408 & 0.5290 & 0.5151 & 0.5015 & 0.4881 & 0.4758  \\
    \textbf{\texttt{DiffImpute} w/ ResNet} & 0.8628 & 0.8533 & \textbf{0.8461} & \textbf{0.8380} & \textbf{0.8262} & 0.8104 & 0.7804 & 0.7105 & 0.5272  \\
    \textbf{\texttt{DiffImpute} w/ Transformer} & \textbf{0.8680} & \textbf{0.8543} & 0.8429 & 0.8325 & 0.8187 & 0.8072 & 0.7950 & 0.7814 & 0.7687  \\
    \textbf{\texttt{DiffImpute} w/ U-Net} & -0.0219 & -0.0392 & -0.0573 & -0.0678 & -0.0730 & -0.0775 & -0.0803 & -0.0839 & -0.0827  \\
    \bottomrule
  \end{tabular}
  }
\end{center}
\end{table}

\begin{table}[!ht]
\caption{Imputation performance comparison in terms of random mask setting, \textit{i.e.} Missing Completely At Random (MCAR), on HE using Pearson correlation. The best results are in \textbf{bold}.}
\label{tab:pc-rm-he}
\begin{center}
\resizebox{\linewidth}{!}{
\begin{tabular}{l|ccccccccc}
    \toprule
       \textbf{Imputation Methods} & {10\%} & {20\%} & {30\%} & {40\%} & {50\%} & {60\%} & {70\%}& {80\%}& {90\%}\\
\hline
    Mean Imputation& 0.7682 & 0.7685 & 0.7690 & 0.7692 & 0.7691 & 0.7692 & 0.7692 & 0.7691 & \textbf{0.7692} \\
    Median Imputation& 0.7646 & 0.7653 & 0.7656 & 0.7656 & 0.7655 & 0.7656 & 0.7656 & 0.7655 & 0.7656    \\
    Mode Imputation& 0.3891 & 0.3929 & 0.3966 & 0.3962 & 0.3953 & 0.4021 & 0.4066 & 0.4101 & 0.4056  \\
    LOCF Imputation& 0.5907 & 0.5884 & 0.5888 & 0.5893 & 0.5899 & 0.5905 & 0.5910 & 0.5910 & 0.5910   \\
    NOCB Imputation& 0.5891 & 0.5885 & 0.5893 & 0.5896 & 0.5896 & 0.5897 & 0.5899 & 0.5899 & 0.5899   \\
    MICE (linear) & 0.9100 & 0.9019 & 0.8882 & 0.8711 & 0.8528 & 0.8236 & 0.7575 & 0.7381 & 0.6740 \\
    GAIN & 0.8414 & 0.8390 & 0.8259 & 0.7960 & 0.7095 & 0.3360 & 0.2034 & 0.3752 & 0.2500   \\
    \hline
    \textbf{\texttt{DiffImpute} w/ MLP} & 0.3977 & 0.3533 & 0.3138 & 0.2783 & 0.2464 & 0.2183 & 0.1938 & 0.1728 & 0.1542  \\
    \textbf{\texttt{DiffImpute} w/ ResNet} & 0.9092 & 0.8987 & 0.8858 & 0.8689 & 0.8450 & 0.8107 & 0.7560 & 0.6594 & 0.4552  \\
    \textbf{\texttt{DiffImpute} w/ Transformer} & \textbf{0.9354} & \textbf{0.9252} & \textbf{0.9130} & \textbf{0.8971} & \textbf{0.8769} & \textbf{0.8521} & \textbf{0.8229} & \textbf{0.7882} & 0.7425  \\
    \textbf{\texttt{DiffImpute} w/ U-Net} & 0.2709 & 0.2677 & 0.2671 & 0.2648 & 0.2613 & 0.2562 & 0.2520 & 0.2474 & 0.2437  \\
    \bottomrule
  \end{tabular}
  }
\end{center}
\end{table}

\begin{table}[!ht]
\caption{Imputation performance comparison in terms of random mask setting, \textit{i.e.} Missing Completely At Random (MCAR), on JA using Pearson correlation. The best results are in \textbf{bold}.}
\label{tab:pc-rm-ja}
\begin{center}
\resizebox{\linewidth}{!}{
\begin{tabular}{l|ccccccccc}
    \toprule
       \textbf{Imputation Methods} & {10\%} & {20\%} & {30\%} & {40\%} & {50\%} & {60\%} & {70\%}& {80\%}& {90\%}\\
\hline
    Mean Imputation& 0.7182 & 0.7179 & 0.7182 & 0.7180 & 0.7182 & 0.7181 & \textbf{0.7180} & \textbf{0.7179} & \textbf{0.7179} \\
    Median Imputation& 0.7140 & 0.7137 & 0.7140 & 0.7137 & 0.7141 & 0.7140 & 0.7139 & 0.7138 & 0.7139    \\
    Mode Imputation& 0.3168 & 0.3176 & 0.3155 & 0.3127 & 0.3094 & 0.3062 & 0.3042 & 0.3034 & 0.3013  \\
    LOCF Imputation& 0.5163 & 0.5162 & 0.5166 & 0.5164 & 0.5165 & 0.5163 & 0.5162 & 0.5160 & 0.5162   \\
    NOCB Imputation& 0.5162 & 0.5164 & 0.5164 & 0.5164 & 0.5160 & 0.5165 & 0.5167 & 0.5167 & 0.5174   \\
    MICE (linear) & 0.6996 & 0.6916 & 0.6855 & 0.6759 & 0.6638 & 0.6489 & 0.6262 & 0.6004 & 0.5631 \\
    GAIN & 0.6658 & 0.6803 & 0.6514 & 0.6283 & 0.5944 & 0.3190 & 0.4965 & 0.4867 & 0.0952   \\
    \hline
    \textbf{\texttt{DiffImpute} w/ MLP} & 0.1892 & 0.1691 & 0.1509 & 0.1359 & 0.1236 & 0.1130 & 0.1032 & 0.0943 & 0.0864  \\
    \textbf{\texttt{DiffImpute} w/ ResNet} & 0.7773 & 0.7672 & 0.7530 & 0.7310 & 0.6974 & 0.6442 & 0.5630 & 0.4463 & 0.2906  \\
    \textbf{\texttt{DiffImpute} w/ Transformer} & \textbf{0.7904} & \textbf{0.7827} & \textbf{0.7739} & \textbf{0.7630} & \textbf{0.7503} & \textbf{0.7351} & 0.7176 & 0.6970 & 0.6743  \\
    \textbf{\texttt{DiffImpute} w/ U-Net} & 0.1525 & 0.1473 & 0.1472 & 0.1500 & 0.1536 & 0.1561 & 0.1571 & 0.1568 & 0.1563  \\
    \bottomrule
  \end{tabular}
  }
\end{center}
\end{table}

\begin{table}[!ht]
\caption{Imputation performance comparison in terms of random mask setting, \textit{i.e.} Missing Completely At Random (MCAR), on HI using Pearson correlation. The best results are in \textbf{bold}.}
\label{tab:pc-rm-hi}
\begin{center}
\resizebox{\linewidth}{!}{
\begin{tabular}{l|ccccccccc}
    \toprule
       \textbf{Imputation Methods} & {10\%} & {20\%} & {30\%} & {40\%} & {50\%} & {60\%} & {70\%}& {80\%}& {90\%}\\
\hline
    Mean Imputation& 0.6248 & 0.6240 & \textbf{0.6243} & \textbf{0.6243} & \textbf{0.6240} & \textbf{0.6243} & \textbf{0.6248} & \textbf{0.6250} & \textbf{0.6251} \\
    Median Imputation& 0.5513 & 0.5419 & 0.5329 & 0.5157 & 0.5246 & 0.5161 & 0.5171 & 0.5172 & 0.5176    \\
    Mode Imputation& 0.3910 & 0.3947 & 0.3880 & 0.3860 & 0.3922 & 0.3977 & 0.4012 & 0.3907 & 0.4091  \\
    LOCF Imputation& 0.3912 & 0.3878 & 0.3896 & 0.3899 & 0.3899 & 0.3911 & 0.3922 & 0.3922 & 0.3922  \\
    NOCB Imputation& 0.3919 & 0.3911 & 0.3920 & 0.3913 & 0.3900 & 0.3913 & 0.3907 & 0.3907 & 0.3907   \\
    MICE (linear) & 0.5469 & 0.5316 & 0.5093 & 0.4861 & 0.4688 & 0.4521 & 0.4329 & 0.4180 & 0.4032 \\
    GAIN & 0.4461 & 0.4754 & 0.4829 & 0.4305 & 0.4532 & 0.3956 & 0.3797 & 0.4449 & 0.2699   \\
    \hline
    \textbf{\texttt{DiffImpute} w/ MLP} & 0.2938 & 0.2781 & 0.2667 & 0.2552 & 0.2437 & 0.2341 & 0.2261 & 0.2189 & 0.2129  \\
    \textbf{\texttt{DiffImpute} w/ ResNet} & \textbf{0.6475} & \textbf{0.6317} & 0.6138 & 0.5914 & 0.5593 & 0.5124 & 0.4430 & 0.3457 & 0.2271  \\
    \textbf{\texttt{DiffImpute} w/ Transformer} & 0.6133 & 0.5994 & 0.5885 & 0.5774 & 0.5671 & 0.5574 & 0.5496 & 0.5416 & 0.5339  \\
    \textbf{\texttt{DiffImpute} w/ U-Net} & 0.0052 & 0.0041 & 0.0036 & 0.0031 & 0.0014 & -0.0001 & -0.0012 & -0.0024 & -0.0036  \\
    \bottomrule
  \end{tabular}
  }
\end{center}
\end{table}

\begin{table}[!ht]
\caption{Imputation performance comparison in terms of random mask setting, \textit{i.e.} Missing Completely At Random (MCAR), on AL using Pearson correlation. The best results are in \textbf{bold}.}
\label{tab:pc-rm-al}
\begin{center}
\resizebox{\linewidth}{!}{
\begin{tabular}{l|ccccccccc}
    \toprule
       \textbf{Imputation Methods} & {10\%} & {20\%} & {30\%} & {40\%} & {50\%} & {60\%} & {70\%}& {80\%}& {90\%}\\
\hline
    Mean Imputation& 0.6797 & 0.6793 & 0.6802 & 0.6800 & 0.6798 & 0.6798 & 0.6798 & 0.6797 & 0.6796 \\
    Median Imputation& 0.6310 & 0.6304 & 0.6313 & 0.6311 & 0.6309 & 0.6310 & 0.6310 & 0.6309  & 0.6308    \\
    Mode Imputation& 0.5520 & 0.5508 & 0.5519 & 0.5515 & 0.5514 & 0.5551 & 0.5515 & 0.5514 & 0.5513  \\
    LOCF Imputation& 0.4617 & 0.4617 & 0.4617 & 0.4617 & 0.4617 & 0.4617 & 0.4617 & 0.4617 & 0.4617   \\
    NOCB Imputation& 0.4612 & 0.4612 & 0.4612 & 0.4612 & 0.4612 & 0.4612 & 0.4612 & 0.4612 & 0.4612   \\
    MICE (linear) & 0.9006 & 0.8912 & 0.8789 & \textbf{0.8660} & 0.8486 & 0.8248 & 0.7907 & 0.7395 & 0.6545 \\
    GAIN & 0.8993 & 0.8804 & 0.7993 & 0.7464 & 0.6900 & 0.6576 & 0.5322 & 0.4854 & 0.4521   \\
    \hline
    \textbf{\texttt{DiffImpute} w/ MLP} & 0.0752 & 0.0546 & 0.0406 & 0.0304 & 0.0227 & 0.0161 & 0.0112 & 0.0069 & 0.0034  \\
    \textbf{\texttt{DiffImpute} w/ ResNet} & 0.8360 & 0.8239 & 0.8049 & 0.7705 & 0.7035 & 0.5849 & 0.4190 & 0.2437 & 0.0939  \\
    \textbf{\texttt{DiffImpute} w/ Transformer} & \textbf{0.9233} & \textbf{0.9133} & \textbf{0.9009} & 0.8845 & \textbf{0.8627} & \textbf{0.8335} & \textbf{0.7952} & \textbf{0.7448} & \textbf{0.6819}  \\
    \textbf{\texttt{DiffImpute} w/ U-Net} & 0.7762 & 0.7597 & 0.7428 & 0.7241 & 0.7035 & 0.6788 & 0.6509 & 0.6156 & 0.5722  \\
    \bottomrule
  \end{tabular}
  }
\end{center}
\end{table}

\begin{table}[!ht]
\caption{Imputation performance comparison in terms of random mask setting, \textit{i.e.} Missing Completely At Random (MCAR), on YE using Pearson correlation. The best results are in \textbf{bold}.}
\label{tab:pc-rm-ye}
\begin{center}
\resizebox{\linewidth}{!}{
\begin{tabular}{l|ccccccccc}
    \toprule
       \textbf{Imputation Methods} & {10\%} & {20\%} & {30\%} & {40\%} & {50\%} & {60\%} & {70\%}& {80\%}& {90\%}\\
\hline
    Mean Imputation& 0.9877 & 0.9876 & 0.9876 & 0.9876 & 0.9876 & 0.9876 & 0.9876 & \textbf{0.9876} & \textbf{0.9876} \\
    Median Imputation& 0.9876 & 0.9875 & 0.9875 & 0.9875 & 0.9875 & 0.9875 & 0.9875 & 0.9875 & 0.9875    \\
    Mode Imputation& 0.9864 & 0.9863 & 0.9863 & 0.9864 & 0.9863 & 0.9864 & 0.9864 & 0.9864 & 0.9865  \\
    LOCF Imputation& 0.9755 & 0.9754 & 0.9754 & 0.9754 & 0.9754 & 0.9754 & 0.9755 & 0.9755 & 0.9755   \\
    NOCB Imputation& 0.9754 & 0.9754 & 0.9754 & 0.9754 & 0.9754 & 0.9754 & 0.9754 & 0.9754 & 0.9754   \\
    MICE (linear) & \textbf{0.9989} & \textbf{0.9977} & \textbf{0.9963} & \textbf{0.9947} & \textbf{0.9928} & \textbf{0.9906} & 0.9829 & 0.9810 & 0.9782 \\
    GAIN & 0.9830 & 0.9815 & 0.9777 & 0.9719 & 0.9384 & 0.7138 & 0.9052 & 0.2598 & 0.2119   \\
    \hline
    \textbf{\texttt{DiffImpute} w/ MLP} & 0.2620 & 0.2079 & 0.1741 & 0.1506 & 0.1340 & 0.1218 & 0.1124 & 0.1052 & 0.0994  \\
    \textbf{\texttt{DiffImpute} w/ ResNet} & 0.9818 & 0.9809 & 0.9789 & 0.9740 & 0.9602 & 0.9206 & 0.8206 & 0.6173 & 0.2984  \\
    \textbf{\texttt{DiffImpute} w/ Transformer} & 0.9921 & 0.9917 & 0.9912 & 0.9906 & 0.9900 & 0.9892 & \textbf{0.9883} & 0.9874 & 0.9862  \\
    \textbf{\texttt{DiffImpute} w/ U-Net} & 0.9499 & 0.9375 & 0.9229 & 0.9045 & 0.8818 & 0.8512 & 0.8102 & 0.7602 & 0.7064  \\
    \bottomrule
  \end{tabular}
  }
\end{center}
\end{table}

\begin{table}[!ht]
\caption{Imputation performance comparison in terms of random mask setting, \textit{i.e.} Missing Completely At Random (MCAR), on CO using Pearson correlation. The best results are in \textbf{bold}.}
\label{tab:pc-rm-co}
\begin{center}
\resizebox{\linewidth}{!}{
\begin{tabular}{l|ccccccccc}
    \toprule
       \textbf{Imputation Methods} & {10\%} & {20\%} & {30\%} & {40\%} & {50\%} & {60\%} & {70\%}& {80\%}& {90\%}\\
\hline
    Mean Imputation& 0.7499 & 0.7499 & 0.7500 & 0.7499 & 0.7500 & 0.7501 & \textbf{0.7501} & \textbf{0.7501} & \textbf{0.7501} \\
    Median Imputation& 0.6827 & 0.6828 & 0.6828 & 0.6828 & 0.6827 & 0.6829 & 0.6828 & 0.6827 & 0.6828    \\
    Mode Imputation& 0.6520 & 0.6520 & 0.6520 & 0.6520 & 0.6520 & 0.6521 & 0.6521 & 0.6520 &  0.6521  \\
    LOCF Imputation& 0.5622 & 0.5628 & 0.5625 & 0.5626 & 0.5623 & 0.5622 & 0.5624 & 0.5627 & 0.5623   \\
    NOCB Imputation& 0.5628 & 0.5631 & 0.5631 & 0.5632 & 0.5630 & 0.5630 & 0.5627 & 0.5625 & 0.5618   \\
    MICE (linear) & -0.0150 & -0.0070 & 0.0036 & -0.0510 & -0.1070 & -0.0390 & 0.1820 & 0.0021 & -0.0020 \\
    GAIN & 0.7928 & 0.7928 & 0.7874 & 0.7475 & 0.5077 & 0.3772 & 0.4619 & 0.4580 & 0.2975   \\
    \hline
    \textbf{\texttt{DiffImpute} w/ MLP} & 0.2707 & 0.2231 & 0.1846 & 0.1526 & 0.1263 & 0.1044 & 0.0852 & 0.0693 & 0.0556  \\
    \textbf{\texttt{DiffImpute} w/ ResNet} & 0.8604 & 0.8441 & 0.8267 & 0.8064 & 0.7815 & 0.7475 & 0.7218 & 0.7054 & 0.6888  \\
    \textbf{\texttt{DiffImpute} w/ Transformer} & \textbf{0.8780} & \textbf{0.8543} & \textbf{0.8317} & \textbf{0.8094} & \textbf{0.7880} & \textbf{0.7671} & 0.7425 & 0.7223 & 0.7034  \\
    \textbf{\texttt{DiffImpute} w/ U-Net} & 0.3785 & 0.3344 & 0.2863 & 0.2348 & 0.1806 & 0.1257 & 0.0703 & 0.0167 & -0.0340  \\
    \bottomrule
  \end{tabular}
  }
\end{center}
\end{table}

\textbf{Column Mask.}
In this segment, we delve into the imputation performance assessment using the Pearson correlation metric under the column mask settings. This approach aligns with the Missing At Random (MAR) paradigm. The detailed results for each of the seven datasets are provided in \Cref{tab:pr-cm-ca,tab:pr-cm-he,tab:pr-cm-ja,tab:pr-cm-hi,tab:pr-cm-al,tab:pr-cm-ye,tab:pr-cm-co}. It's pertinent to mention that the column mask setting renders the NOCB imputation method inapplicable, given the lack of a subsequent observation for imputation purposes.

\begin{table}[!ht]
\caption{Imputation performance comparison in terms of column mask setting, \textit{i.e.} Missing Completely At Random (MCAR), on CA using Pearson correlation. The best results are in \textbf{bold}.}
\label{tab:pr-cm-ca}
\begin{center}\resizebox{\linewidth}{!}{
\begin{tabular}{l|cccc}
    \toprule
       \textbf{Imputation Methods} & {1} & {2} & {3} & {4}\\
\hline
    Mean Imputation& \textbf{0.8140} & \textbf{0.8140} & \textbf{0.8140} & \textbf{0.8140} \\
    Median Imputation& 0.7787 & 0.7787 & 0.7787 & 0.7787 \\
    Mode Imputation& 0.7385 & 0.7385 & 0.7385 & 0.7385   \\
    LOCF Imputation& 0.6615 & 0.6615 & 0.6615 & 0.6615 \\
    NOCB Imputation& / & / & / & / \\
    MICE (linear) & 0.1814 & 0.2818 & 0.6596 & 0.9691\\
    GAIN & 0.0323 & 0.2640 & 0.7887 & 0.9685 \\
    \hline
    \textbf{\texttt{DiffImpute} w/ MLP} & 0.0317 & 0.3627 & 0.3746 & 0.5798\\
    \textbf{\texttt{DiffImpute} w/ ResNet} & 0.1733 & -0.0002 & 0.3057 & -0.0469 \\
    \textbf{\texttt{DiffImpute} w/ Transformer} & 0.2575 & 0.6394 & 0.7743 & 0.9175 \\
    \textbf{\texttt{DiffImpute} w/ U-Net} & -0.0022 & -0.0640 & -0.0143 & -0.0897 \\
    \bottomrule
  \end{tabular}}
\end{center}
\end{table}

\begin{table}[!ht]
\caption{Imputation performance comparison in terms of column mask setting, \textit{i.e.} Missing Completely At Random (MCAR), on HE using Pearson correlation. The best results are in \textbf{bold}.}
\label{tab:pr-cm-he}
\begin{center}\resizebox{\linewidth}{!}{
\begin{tabular}{l|cccc}
    \toprule
       \textbf{Imputation Methods} & {1} & {2} & {3} & {4}\\
\hline
    Mean Imputation& 0.7692 & 0.7692 & 0.7692 & 0.7692 \\
    Median Imputation& 0.7656 & 0.7656 & 0.7656 & 0.7656 \\
    Mode Imputation& 0.4056 & 0.4056 & 0.4056 & 0.4056   \\
    LOCF Imputation& 0.5911 & 0.5911 & 0.5911 & 0.5911 \\
    NOCB Imputation& / & / & / & / \\
    MICE (linear) & 0.0797 & \textbf{0.9836} & 0.7218 & 0.7660 \\
    GAIN & 0.0509 & 0.9713 & 0.7839 & 0.7937 \\
    \hline
    \textbf{\texttt{DiffImpute} w/ MLP} & 0.0457 & 0.2731 & 0.1824 & 0.3239\\
    \textbf{\texttt{DiffImpute} w/ ResNet} & 0.5779 & 0.2973 & 0.5354 & 0.5509 \\
    \textbf{\texttt{DiffImpute} w/ Transformer} & \textbf{0.7734} & 0.8365 & \textbf{0.8169} & \textbf{0.8914} \\
    \textbf{\texttt{DiffImpute} w/ U-Net} & 0.0572 & 0.2208 & 0.0971 & 0.1945 \\
    \bottomrule
  \end{tabular}}
\end{center}
\end{table}

\begin{table}[!ht]
\caption{Imputation performance comparison in terms of column mask setting, \textit{i.e.} Missing Completely At Random (MCAR), on JA using Pearson correlation. The best results are in \textbf{bold}.}
\label{tab:pr-cm-ja}
\begin{center}\resizebox{\linewidth}{!}{
\begin{tabular}{l|cccc}
    \toprule
       \textbf{Imputation Methods} & {1} & {2} & {3} & {4}\\
\hline
    Mean Imputation& \textbf{0.7179} & \textbf{0.7179} & 0.7179 & 0.7179 \\
    Median Imputation& 0.7138 & 0.7139 & 0.7139 & 0.7139 \\
    Mode Imputation& 0.3013 & 0.3013 & 0.3013 & 0.3013   \\
    LOCF Imputation& 0.5162 & 0.5162 & 0.5162 & 0.5162 \\
    NOCB Imputation& / & /& / & /\\
    MICE (linear) & -0.0090 & 0.2864 & 0.8519 & 0.7471 \\
    GAIN & 0.0141 & 0.3346 & \textbf{0.8844} & 0.8060 \\
    \hline
    \textbf{\texttt{DiffImpute} w/ MLP} & 0.0138 & 0.0548 & 0.0941 & 0.1827\\
    \textbf{\texttt{DiffImpute} w/ ResNet} & 0.1849 & 0.2422 & 0.3213 & 0.4518 \\
    \textbf{\texttt{DiffImpute} w/ Transformer} & 0.1979 & 0.3747 & 0.8622 & \textbf{0.8505} \\
    \textbf{\texttt{DiffImpute} w/ U-Net} & -0.0180 & 0.0699 & 0.0916 & 0.2285 \\
    \bottomrule
  \end{tabular}}
\end{center}
\end{table}

\begin{table}[!ht]
\caption{Imputation performance comparison in terms of column mask setting, \textit{i.e.} Missing Completely At Random (MCAR), on HI using Pearson correlation. The best results are in \textbf{bold}.}
\label{tab:pr-cm-hi}
\begin{center}\resizebox{\linewidth}{!}{
\begin{tabular}{l|cccc}
    \toprule
       \textbf{Imputation Methods} & {1} & {2} & {3} & {4}\\
\hline
    Mean Imputation& \textbf{0.6251} & \textbf{0.6251} & 0.6251 & 0.6251 \\
    Median Imputation& 0.5176 & 0.5176 & 0.5176 & 0.5176 \\
    Mode Imputation& 0.4091 & 0.4091 & 0.4091 & 0.4091   \\
    LOCF Imputation& 0.3911 & 0.3911 & 0.3911 & 0.3911 \\
    NOCB Imputation& / & / & / & / \\
    MICE (linear) & 0.5234 & 0.3278 & 0.5956 & 0.5393\\
    GAIN & 0.3119 & 0.2392 & \textbf{0.7328} & \textbf{0.6325} \\
    \hline
    \textbf{\texttt{DiffImpute} w/ MLP} & 0.0024 & 0.1180 & 0.2406 & 0.2306\\
    \textbf{\texttt{DiffImpute} w/ ResNet} & 0.2010 & -0.0460 & 0.3727 & 0.1481 \\
    \textbf{\texttt{DiffImpute} w/ Transformer} & 0.4956 & 0.3981 & 0.5255 & 0.5383 \\
    \textbf{\texttt{DiffImpute} w/ U-Net} & -0.0030 & 0.0196 & 0.0095 & 0.5255 \\
    \bottomrule
  \end{tabular}}
\end{center}
\end{table}

\begin{table}[!ht]
\caption{Imputation performance comparison in terms of column mask setting, \textit{i.e.} Missing Completely At Random (MCAR), on AL using Pearson correlation. The best results are in \textbf{bold}.}
\label{tab:pr-cm-al}
\begin{center}\resizebox{\linewidth}{!}{
\begin{tabular}{l|cccc}
    \toprule
       \textbf{Imputation Methods} & {1} & {2} & {3} & {4}\\
\hline
    Mean Imputation& 0.6796 & 0.6796 & 0.6796 & 0.6796 \\
    Median Imputation& 0.6308 & 0.6308 & 0.6308 & 0.6308 \\
    Mode Imputation& 0.5513 & 0.5513 & 0.5513 & 0.5513   \\
    LOCF Imputation& 0.4617 & 0.4617 & 0.4617 & 0.4617 \\
    NOCB Imputation& / & / & / & / \\
    MICE (linear) & 0.7555 & 0.8037 & 0.8102 & 0.8228\\
    GAIN & 0.7392 & 0.7910 & 0.8314 & 0.8373 \\
    \hline
    \textbf{\texttt{DiffImpute} w/ MLP} & 0.0329 & 0.0131 & 0.0256 & 0.0562\\
    \textbf{\texttt{DiffImpute} w/ ResNet} & 0.4733 & 0.4771 & 0.4027 & 0.3236 \\
    \textbf{\texttt{DiffImpute} w/ Transformer} & \textbf{0.8375} & \textbf{0.8549} & \textbf{0.8666} & \textbf{0.8738} \\
    \textbf{\texttt{DiffImpute} w/ U-Net} & 0.5533 & 0.6889 & 0.6778 & 0.7592 \\
    \bottomrule
  \end{tabular}}
\end{center}
\end{table}

\begin{table}[!ht]
\caption{Imputation performance comparison in terms of column mask setting, \textit{i.e.} Missing Completely At Random (MCAR), on YE using Pearson correlation. The best results are in \textbf{bold}.}
\label{tab:pr-cm-ye}
\begin{center}\resizebox{\linewidth}{!}{
\begin{tabular}{l|cccc}
    \toprule
       \textbf{Imputation Methods} & {1} & {2} & {3} & {4}\\
\hline
    Mean Imputation& \textbf{0.9876} & \textbf{0.9876} & 0.9876 & 0.9876 \\
    Median Imputation& 0.9875 & 0.9875 & 0.9875 & 0.9875 \\
    Mode Imputation& 0.9863 & 0.9863 & 0.9863 & 0.9863   \\
    LOCF Imputation& 0.9839 & 0.9839 & 0.9839 & 0.9839 \\
    NOCB Imputation& / & / & / & / \\
    MICE (linear) & 0.3292 & 0.8135 & \textbf{0.9912} & \textbf{0.9918}\\
    GAIN & 0.0309 & 0.6259 & 0.9925 & 0.9883 \\
    \hline
    \textbf{\texttt{DiffImpute} w/ MLP} & -0.0009 & 0.3019 & 0.1618 & 0.1459\\
    \textbf{\texttt{DiffImpute} w/ ResNet} & 0.0516 & 0.0828 & -0.2318 & -0.2155 \\
    \textbf{\texttt{DiffImpute} w/ Transformer} & 0.5469 & 0.9382 & 0.9049 & 0.9478 \\
    \textbf{\texttt{DiffImpute} w/ U-Net} & 0.0254 & 0.7423 & 0.9545 & 0.9638 \\
    \bottomrule
  \end{tabular}}
\end{center}
\end{table}

\begin{table}[!ht]
\caption{Imputation performance comparison in terms of column mask setting, \textit{i.e.} Missing Completely At Random (MCAR), on CO using Pearson correlation. The best results are in \textbf{bold}.}
\label{tab:pr-cm-co}
\begin{center}\resizebox{\linewidth}{!}{
\begin{tabular}{l|cccc}
    \toprule
       \textbf{Imputation Methods} & {1} & {2} & {3} & {4}\\
\hline
    Mean Imputation& \textbf{0.7501} & \textbf{0.7501} & \textbf{0.7501} & \textbf{0.7501} \\
    Median Imputation& 0.6828 & 0.6828 & 0.6828 & 0.6828 \\
    Mode Imputation& 0.6521 & 0.6521 & 0.6521 & 0.6521   \\
    LOCF Imputation& 0.5621 & 0.5621 & 0.5621 & 0.5621 \\
    NOCB Imputation& / & / & / & / \\
    MICE (linear) & NaN & NaN & NaN & NaN\\
    GAIN & NaN & NaN & NaN & NaN \\
    \hline
    \textbf{\texttt{DiffImpute} w/ MLP} & 0.0121 & 0.1933 & 0.1223 & 0.1786 \\
    \textbf{\texttt{DiffImpute} w/ ResNet} & 0.1872 & 0.5201 & 0.4335 & 0.6617 \\
    \textbf{\texttt{DiffImpute} w/ Transformer} & 0.4553 & 0.7481 & 0.6273 & 0.7497 \\
    \textbf{\texttt{DiffImpute} w/ U-Net} & -0.0028 & 0.1780 & 0.2590 & 0.2288 \\
    \bottomrule
  \end{tabular}}
\end{center}
\end{table}

\textbf{Pearson Correlation Performance Rankings.}
This section presents overall Pearson correlation performance rankings under different mask settings (MCAR, and MAR) across seven datasets, as shown in \Cref{Pearson Correlation ranking,Pearson Correlation ranking_MCAR,Pearson Correlation ranking_MAR}.

\begin{table}[ht]
\caption{Overall Pearson correlation rankings under the random mask setting (MCAR). \texttt{DiffImpute} with Transformer outperform other methods in six datasets out of seven datasets. The best results are in \textbf{bold}.}
\label{Pearson Correlation ranking_MCAR}
\begin{center}
\resizebox{\linewidth}{!}{\begin{tabular}{l|lllllll|ll}
\toprule
\multicolumn{1}{c|}{\bf Imputation Methods}  &\multicolumn{1}{c}{\bf CA}  &\multicolumn{1}{c}{\bf HE}  &\multicolumn{1}{c}{\bf JA}  &\multicolumn{1}{c}{\bf HI}  &\multicolumn{1}{c}{\bf AL}  &\multicolumn{1}{c}{\bf YE}  &\multicolumn{1}{c|}{\bf CO} &\multicolumn{1}{c}{\bf Mean} &\multicolumn{1}{c}{\bf Std}\\
\toprule
Mean Imputation         &2.6	&3.7	&2.1	&\textbf{1.2}	&4.6	&2.4	&2.6	&2.7	&1.0  \\
Median Imputation       &4.4	&4.7	&3.2	&3.6	&6.0	&3.4	&4.4	&4.3	&0.9  \\
Mode Imputation         &5.9	&8.7	&8.8	&7.1	&7.3	&4.6	&5.4	&6.8	&1.5  \\
LOCF Imputation         &7.4	&6.8	&6.9	&7.8	&8.6	&6.7	&7.2	&7.3	&0.6  \\
NOCB Imputation         &8.2	&7.1	&6.7	&7.4	&9.6	&7.2	&6.7	&7.6	&1.0  \\
MICE                    &4.6	&2.7	&4.6	&5.0	&2.1	&2.3	&10.7	&4.6	&2.7  \\
GAIN                    &7.4	&6.6	&7.0	&6.4	&5.2	&8.3	&5.9	&6.7	&0.9  \\
\hline
\textbf{\texttt{DiffImpute} w/ MLP}      &9.7	&10.4	&10.7	&10.0	&11.0	&11.0	&9.8	&10.4	&0.5  \\
\textbf{\texttt{DiffImpute} w/ ResNet}   &2.8	&3.9	&4.2	&3.9	&6.0	&7.9	&2.4	&4.4	&1.8  \\
\textbf{\texttt{DiffImpute} w/ Transformer}   &\textbf{2.0}	&\textbf{1.2}	&\textbf{1.6}	&2.4	&\textbf{1.0}	&\textbf{2.2}	&\textbf{1.3}	&1.7	&0.5   \\  
\textbf{\texttt{DiffImpute} w/ U-Net}    &11.0	&10.3	&10.2	&11.0	&4.6	&9.4	&9.6	&9.4	&2.1  \\
\bottomrule
\end{tabular}}
\end{center}
\end{table}

\begin{table}[ht]
\caption{Overall Pearson correlation rankings under the random mask setting (MCAR). \texttt{DiffImpute} with Transformer outperform other methods in two datasets out of seven datasets. The mean imputaion methods outperform other methods in five datasets. The best results are in \textbf{bold}.}
\label{Pearson Correlation ranking_MAR}
\begin{center}
\resizebox{\linewidth}{!}{\begin{tabular}{l|lllllll|ll}
\toprule
\multicolumn{1}{c|}{\bf Imputation Methods}  &\multicolumn{1}{c}{\bf CA}  &\multicolumn{1}{c}{\bf HE}  &\multicolumn{1}{c}{\bf JA}  &\multicolumn{1}{c}{\bf HI}  &\multicolumn{1}{c}{\bf AL}  &\multicolumn{1}{c}{\bf YE}  &\multicolumn{1}{c|}{\bf CO} &\multicolumn{1}{c}{\bf Mean} &\multicolumn{1}{c}{\bf Std}\\
\toprule
Mean Imputation         &\textbf{1.8}	&3.0	&\textbf{2.5}	&\textbf{1.5}	&4.5	&\textbf{2.0}	&\textbf{1.0}	&2.3	&1.1 \\
Median Imputation       &3.0	&4.3	&3.5	&4.0	&5.8	&3.0	&2.5	&3.7	&1.0 \\
Mode Imputation         &4.3	&7.3	&6.5	&5.3	&7.0	&4.0	&3.8	&5.4	&1.4 \\
LOCF Imputation         &5.3	&5.5	&4.5	&6.5	&8.5	&5.0	&5.0	&5.8	&1.3 \\
NOCB Imputation         &/    &/   &/   &/   &/   &/   &/   &/   &/			\\
MICE                    &5.3	&4.3	&5.5	&3.5	&2.5	&3.8	&NaN	&4.1	&1.0 \\
GAIN                    &5.0	&3.8	&3.8	&4.0	&2.5	&4.8	&NaN	&4.0	&0.8 \\
\hline
\textbf{\texttt{DiffImpute} w/ MLP}      &7.8	&9.3	&9.3	&8.8	&10.0	&9.3	&7.5	&8.8	&0.8 \\
\textbf{\texttt{DiffImpute} w/ ResNet}   &8.5	&6.8	&7.0	&9.0	&8.5	&9.3	&5.5	&7.8	&1.3 \\
\textbf{\texttt{DiffImpute} w/ Transformer}   &4.3	&\textbf{1.5}	&3.0	&4.0	&\textbf{1.0}	&6.5	&3.3	&3.4	&1.7  \\  
\textbf{\texttt{DiffImpute} w/ U-Net}    &10.0	&9.5	&9.5	&8.5	&4.8	&7.5	&7.5	&8.2	&1.7  \\
\bottomrule
\end{tabular}}
\end{center}
\end{table}

\begin{table}[ht]
\caption{Overall Pearson correlation rankings of MCAR and MAR (MSE). \texttt{DiffImpute} with Transformer outperform other methods in four datasets and the mean imputation methods outperform other methods in three datasets. The best results are in \textbf{bold}.}
\label{Pearson Correlation ranking}
\begin{center}
\resizebox{\linewidth}{!}{\begin{tabular}{l|lllllll|ll}
\toprule
\multicolumn{1}{c|}{\bf Imputation Methods}  &\multicolumn{1}{c}{\bf CA}  &\multicolumn{1}{c}{\bf HE}  &\multicolumn{1}{c}{\bf JA}  &\multicolumn{1}{c}{\bf HI}  &\multicolumn{1}{c}{\bf AL}  &\multicolumn{1}{c}{\bf YE}  &\multicolumn{1}{c|}{\bf CO} &\multicolumn{1}{c}{\bf Mean} &\multicolumn{1}{c}{\bf Std}\\
\toprule
Mean Imputation         &\textbf{2.3}	&3.5	&2.2	&\textbf{1.3}	&4.5	&\textbf{2.3}	&2.1	&2.6	&1.1\\
Median Imputation       &4.0	&4.5	&3.3	&3.7	&5.9	&3.3	&3.8	&4.1	&0.9\\
Mode Imputation         &5.4	&8.2	&8.1	&6.5	&7.2	&4.4	&4.9	&6.4	&1.5\\
LOCF Imputation         &6.8	&6.4	&6.2	&7.4	&8.5	&6.2	&6.5	&6.8	&0.9\\
NOCB Imputation         &8.2	&7.1	&6.7	&7.4	&9.6	&7.2	&6.7	&7.6	&1.0\\
MICE                    &4.8	&3.2	&4.8	&4.5	&2.2	&2.8	&10.7	&4.7	&2.8\\
GAIN                    &6.7	&5.7	&6.0	&5.7	&4.4	&7.2	&5.9	&5.9	&0.9\\
\hline
\textbf{\texttt{DiffImpute} w/ MLP}      &9.1	&10.1	&10.2	&9.6	&10.7	&10.5	&9.1	&9.9	&0.6\\
\textbf{\texttt{DiffImpute} w/ ResNet}   &4.5	&4.8	&5.1	&5.5	&6.8	&8.3	&3.4	&5.5	&1.6\\
\textbf{\texttt{DiffImpute} w/ Transformer}     &2.7	&\textbf{1.3}	&\textbf{2.0}	&2.9	&\textbf{1.0}	&3.5	&\textbf{1.9}	&2.2	&0.9\\
\textbf{\texttt{DiffImpute} w/ U-Net}    &10.7	&10.1	&10.0	&10.2	&4.6	&8.8	&8.9	&9.1	&2.1\\
\bottomrule
\end{tabular}}
\end{center}
\end{table}

\subsection{Performance on Downstream Tasks.}
In this section, we present the performance metrics of downstream tasks for imputed data, considering various missingness mechanisms across our seven benchmark datasets. Specifically, for regression tasks, we employ the root mean squared error (RMSE) as the evaluation metric, while classification tasks are gauged using the accuracy score. Our focus here is on the random mask settings, which align with the Missing Completely At Random (MCAR) setting.

\textbf{Random Mask.}
Delving deeper into the random mask settings, we evaluate the downstream task performance in the context of the Missing Completely At Random (MCAR). Detailed results for each of the seven datasets are provided in \Cref{tab:ds-rm-ca,tab:ds-rm-he,tab:ds-rm-ja,tab:ds-rm-hi,tab:ds-rm-al,tab:ds-rm-ye,tab:ds-rm-co}.

\begin{table}[!ht]
\caption{Downstream task performance comparison in random mask setting (MCAR) on CA, evaluated by RMSE. For each missing setting, the best results are in \textbf{bold}.}
\label{tab:ds-rm-ca}
\begin{center}
\resizebox{\linewidth}{!}{
\begin{tabular}{l|ccccccccc}
    \toprule
       \textbf{Imputation Methods} & {10\%} & {20\%} & {30\%} & {40\%} & {50\%} & {60\%} & {70\%}& {80\%}& {90\%}\\
\hline
    Mean Imputation& 0.8707 & 1.0113 & 1.0974 & 1.1683 & 1.2189 & 1.2532 &  1.2680 & 1.2615 &  1.2461 \\
    Median Imputation& 0.8986 & 1.0449 & 1.1319 & 1.2037 & 1.2480 & 1.2753 & 1.2795 &1.2527 & \textbf{1.2150}    \\
    Mode Imputation& 0.9982 & 1.3324 & 1.3552 & 1.6428 & 1.5582 & 1.6270 & 1.3985 & 1.3580 &  1.2889  \\
    0 Imputation& 1.1661 & 1.4571 & 1.6696 & 1.8366 & 1.9694 & 2.073 & 2.1479 & 2.2096 & 2.2443  \\
    1 Imputation& 1.3509 & 1.6528 & 1.8069 & 1.8886 & 1.9268 & 1.9520 & 1.9805 & 2.0049 & 2.0774   \\
    LOCF Imputation& 1.5345 & 1.6405 & 1.6802 & 1.4143 & 1.7231 & 1.7528 & 1.7746 & 1.787 & 1.8204   \\
    NOCB Imputation& 1.5317 & 1.6512 & 1.6996 & 1.4195 & 1.7400 & 1.7782 & 1.8056 & 1.8163 & 1.8216  \\
    MICE(linear) & 0.7643 & 0.8571 & 0.9543 & 1.0534 & 1.1461 & 1.2349 & 1.3023 & 1.3927 & 1.4240 \\
    GAIN  & 0.8464 & 0.9473 & 0.9991 & 1.1548 & 1.2405 & 1.3517 & 1.8428 & 2.1072 & 2.2291  \\
    \hline
    \textbf{\texttt{DiffImpute} w/ MLP} & 0.9986 & 1.2324 & 1.4155 & 1.5677 & 1.7011 & 1.8234 & 1.9264 & 2.0195 & 2.1030  \\
    \textbf{\texttt{DiffImpute} w/ ResNet} & 0.7917 & 0.8916 & 0.9637 & 1.0388 & 1.1239 & 1.2563 & 1.5024 & 1.9100 & 2.2878  \\
    \textbf{\texttt{DiffImpute} w/ Transformer} & \textbf{0.7614} & \textbf{0.8365} & \textbf{0.8951} &\textbf{0.9633} & \textbf{1.0286} & \textbf{1.0874} & \textbf{1.1465} & \textbf{1.1994} & 1.2527  \\
    \textbf{\texttt{DiffImpute} w/ U-Net} & 1.2736 & 1.6123 & 1.8475 & 2.0147 & 2.1314 & 2.2267 & 2.2929 & 2.3461 & 2.3812  \\
    \bottomrule
  \end{tabular}
  }
\end{center}
\end{table}

\begin{table}[!ht]
\caption{Downstream task performance comparison in random mask setting (MCAR) on HE, evaluated by accuracy score, the best results are in \textbf{bold}.}
\label{tab:ds-rm-he}
\begin{center}
\resizebox{\linewidth}{!}{
\begin{tabular}{l|ccccccccc}
    \toprule
       \textbf{Imputation Methods} & {10\%} & {20\%} & {30\%} & {40\%} & {50\%} & {60\%} & {70\%}& {80\%}& {90\%}\\
\hline
    Mean Imputation& 0.3172 & 0.2723 & 0.2291 & 0.1874 & 0.1484 & 0.1149 & 0.0866 & 0.0643 & 0.0511 \\
    Median Imputation& 0.3160 & 0.2705 & 0.2288 & 0.1874 & 0.1481 & 0.1131 & 0.0832 & 0.0567 & 0.0344    \\
    Mode Imputation& 0.2931 & 0.2361 & 0.1877 & 0.1484 & 0.1176 & 0.0914 & 0.0694 & 0.0531 &  0.0412  \\
    0 Imputation& 0.2295 & 0.1584 & 0.1203 & 0.0975 & 0.0810 & 0.0706 & 0.0646 & 0.0606 & 0.0596  \\
    1 Imputation& 0.2238 & 0.1453 & 0.0963 & 0.0692 & 0.0524 & 0.0400 & 0.0323 & 0.0261 & 0.0207   \\
    LOCF Imputation& 0.0234 & 0.0266 & 0.0252 & 0.0260 & 0.0256 & 0.0250 & 0.0240 & 0.0240 & 0.0240   \\
    NOCB Imputation& 0.0243 & 0.0270 & 0.0262 & 0.0256 & 0.0266 & 0.0246 & 0.0256 & 0.0256 & 0.0256   \\
    MICE (linear) & 0.3345 & 0.3083 & 0.2812 & 0.2433 & \textbf{0.2036} & \textbf{0.1600} &\textbf{0.1206} & \textbf{0.0875} & 0.0538 \\
    GAIN & 0.3246 & 0.2798 & 0.2425 & 0.1968 & 0.1304 & 0.0937 & 0.0747 & 0.0655 & 0.0601   \\
    \hline
    \textbf{\texttt{DiffImpute} w/ MLP} & 0.2695 & 0.2007 & 0.1486 & 0.1115 & 0.0866 & 0.0701 & 0.0579 & 0.0499 & 0.0440  \\
    \textbf{\texttt{DiffImpute} w/ ResNet} & 0.3313 & 0.2980 & 0.2621 & 0.2199 & 0.1726 & 0.1266 & 0.0868 & 0.0671 & \textbf{0.0610}  \\
    \textbf{\texttt{DiffImpute} w/ Transformer} & \textbf{0.3397} & \textbf{0.3145} & \textbf{0.2826} & \textbf{0.2439} & 0.1986 & 0.1567 & 0.1148 & 0.0780 & 0.0485  \\
    \textbf{\texttt{DiffImpute} w/ U-Net} & 0.2531 & 0.1800 & 0.1327 & 0.1036& 0.0826 & 0.0685 & 0.0578 & 0.0518 & 0.0474  \\
    \bottomrule
  \end{tabular}
  }
\end{center}
\end{table}

\begin{table}[!ht]
\caption{Downstream task performance comparison in random mask setting (MCAR) on JA, evaluated by accuracy score, the best results are in \textbf{bold}.}
\label{tab:ds-rm-ja}
\begin{center}
\resizebox{\linewidth}{!}{
\begin{tabular}{l|ccccccccc}
    \toprule
       \textbf{Imputation Methods} & {10\%} & {20\%} & {30\%} & {40\%} & {50\%} & {60\%} & {70\%}& {80\%}& {90\%}\\
\hline
    Mean Imputation& 0.6863 & 0.6547 & 0.6173 & 0.5762 & 0.5307 & 0.4782 & 0.4215 & 0.3579 & 0.2875 \\
    Median Imputation& 0.6829 & 0.6497 & 0.6144 & 0.5743 & 0.5279 & 0.4776 & 0.4228 & 0.3693 & 0.3327    \\
    Mode Imputation& 0.6577 & 0.6150 & 0.5813 & 0.5532 & 0.5299 & 0.5119 & 0.4983 & 0.4840 & \textbf{0.4736}  \\
    0 Imputation& 0.6243 & 0.5681 & 0.5342 & 0.5127 & 0.4979 & 0.4867 & 0.4767 & 0.4717 & 0.4664  \\
    1 Imputation& 0.6289 & 0.5728 & 0.5317 & 0.5023 & 0.4816 & 0.4618 & 0.4449 & 0.4285 & 0.4037  \\
    LOCF Imputation& 0.3759 & 0.3803 & 0.3839 & 0.3864 & 0.3858 & 0.3904 & 0.3907 & 0.3902 & 0.3935   \\
    NOCB Imputation& 0.3766 & 0.3794 & 0.3831 & 0.3847 & 0.3880 & 0.3894 & 0.3921 & 0.3922 & 0.3932   \\
    MICE (linear) & 0.6975 & 0.6780 & 0.6578 & 0.6291 & 0.5969 & \textbf{0.5699} & \textbf{0.5283} & \textbf{0.4902} & 0.4397\\
    GAIN & 0.6658 & 0.6803 & 0.6302 & 0.5909 & 0.5436 & 0.5054 & 0.4931 & 0.4697 & 0.4669   \\
    \hline
    \textbf{\texttt{DiffImpute} w/ MLP} & 0.6461 & 0.5903 & 0.5494 & 0.5183 & 0.4972 & 0.4797 & 0.4664 & 0.4585 & 0.4541  \\
    \textbf{\texttt{DiffImpute} w/ ResNet} & 0.6905 & 0.6658 & 0.6409 & 0.5183 & 0.5724 & 0.5308 & 0.4937 & 0.4707 & 0.4572  \\
    \textbf{\texttt{DiffImpute} w/ Transformer} & \textbf{0.6998} & \textbf{0.6838} & \textbf{0.6624} & \textbf{0.6379} & \textbf{0.6045} & 0.5637 & 0.5177 & 0.4608 & 0.3970  \\
    \textbf{\texttt{DiffImpute} w/ U-Net} & 0.6421 & 0.5881 & 0.5477 & 0.5197 & 0.4973 & 0.4797 & 0.4651 & 0.4586 & 0.4527  \\
    \bottomrule
  \end{tabular}
  }
\end{center}
\end{table}

\begin{table}[!ht]
\caption{Downstream task performance comparison in random mask setting on HI, evaluated by accuracy score, the best results are in \textbf{bold}.}
\label{tab:ds-rm-hi}
\begin{center}
\resizebox{\linewidth}{!}{
\begin{tabular}{l|ccccccccc}
    \toprule
       \textbf{Imputation Methods} & {10\%} & {20\%} & {30\%} & {40\%} & {50\%} & {60\%} & {70\%}& {80\%}& {90\%}\\
\hline
    Mean Imputation& 0.6931 & 0.6713 & 0.6515 & 0.6305 & \textbf{0.6135} & \textbf{0.5950} & \textbf{0.5786} & \textbf{0.5629} & \textbf{0.5463} \\
    Median Imputation& 0.6929 & 0.6708 & 0.6506 & 0.6305 & 0.6114 & 0.5907 & 0.5736 & 0.5573 & 0.5430    \\
    Mode Imputation& 0.6915 & 0.6670 & 0.6441 & 0.6232 & 0.6026 & 0.5840 & 0.5671 & 0.5528 & 0.5409  \\
    0 Imputation& 0.6823 & 0.6507 & 0.6242 & 0.5984 & 0.5741 & 0.5516 & 0.5276 & 0.5040 & 0.4867  \\
    1 Imputation& 0.6385 & 0.5844 & 0.5447 & 0.5188 & 0.5004 & 0.4872 & 0.4791 & 0.4747 & 0.4724   \\
    LOCF Imputation& 0.5014 & 0.4994 & 0.4997 & 0.4976 & 0.5006 & 0.5013 & 0.5017 & 0.5017 & 0.5017   \\
    NOCB Imputation& 0.4994 & 0.4978 & 0.4977 & 0.4990 & 0.4986 & 0.4973 & 0.4974 & 0.4974 & 0.4974   \\
    MICE (linear) & 0.6890 & 0.6669 & 0.6453 & 0.6114 & 0.5906 & 0.5645 & 0.5480 & 0.5286 & 0.5119 \\
    GAIN & 0.6849 &0.6527 & 0.6280 & 0.6105 & 0.5945 & 0.5544 & 0.5378 & 0.5102 & 0.4874   \\
    \hline
    \textbf{\texttt{DiffImpute} w/ MLP} & 0.6768 & 0.6394 & 0.6120 & 0.5881 & 0.5674 & 0.5483 & 0.5340 & 0.5175 & 0.5050  \\
    \textbf{\texttt{DiffImpute} w/ ResNet} & 0.6909 & 0.6664 & 0.6420 & 0.6176 & 0.5917 & 0.5670 & 0.5383 & 0.5044 & 0.4836  \\
    \textbf{\texttt{DiffImpute} w/ Transformer} & \textbf{0.6979} & \textbf{0.6767} & \textbf{0.6545} & \textbf{0.6340} & 0.6097 & 0.5870 & 0.5652 & 0.5406 & 0.5196  \\
    \textbf{\texttt{DiffImpute} w/ U-Net} & 0.6665 & 0.6243 & 0.5922 & 0.5681 & 0.5459 & 0.5284 & 0.5139 & 0.5016 & 0.4939  \\
    \bottomrule
  \end{tabular}
  }
\end{center}
\end{table}

\begin{table}[!ht]
\caption{Downstream task performance comparison in random mask setting on AL, evaluated by accuracy score, the best results are in \textbf{bold}.}
\label{tab:ds-rm-al}
\begin{center}
\resizebox{\linewidth}{!}{
\begin{tabular}{l|ccccccccc}
    \toprule
       \textbf{Imputation Methods} & {10\%} & {20\%} & {30\%} & {40\%} & {50\%} & {60\%} & {70\%}& {80\%}& {90\%}\\
\hline
    Mean Imputation& 0.8002 & 0.6321 & 0.4549 & 0.2964 & 0.1756 & 0.0927 & 0.0421 & 0.0160 & 0.0052 \\
    Median Imputation& 0.8325 & 0.7148 & 0.5730 & 0.4247 & 0.2877 & 0.1724 & 0.0891 & 0.0359 & 0.0098    \\
    Mode Imputation& 0.8080 & 0.6604 & 0.4953 & 0.3371 & 0.2104 & 0.1155 & 0.0557 & 0.0229 & 0.0072  \\
    0 Imputation& 0.7102 & 0.4903 & 0.3057 & 0.1729 & 0.0915 & 0.0448 & 0.0211 & 0.0092 & 0.0036  \\
    1 Imputation& 0.1194 & 0.0272 & 0.0064 & 0.0021 & 0.0013 & 0.0012 & 0.0011 & 0.0011 & 0.0011   \\
    LOCF Imputation& 0.0009 & 0.0009 & 0.0009 & 0.0009 & 0.0009 & 0.0009 & 0.0009 & 0.0009 & 0.0009   \\
    NOCB Imputation& 0.0010 & 0.0010 & 0.0010 & 0.0010 & 0.0010 & 0.0010 & 0.0010 & 0.0010 & 0.0010  \\
    MICE (linear) & 0.8724 & 0.7969 & 0.6883 & 0.5724 & 0.4309 & \textbf{0.2951} & \textbf{0.1693} & \textbf{0.0788} & \textbf{0.0202} \\
    GAIN & 0.8724 & 0.7575 & 0.5574 & 0.3936 & 0.2470 & 0.1364 & 0.0551 & 0.0168 & 0.0040   \\
    \hline
    \textbf{\texttt{DiffImpute} w/ MLP} & 0.4176 & 0.1748 & 0.0751 & 0.0344 & 0.0169 & 0.0085 & 0.0045 & 0.0029 & 0.0019  \\
    \textbf{\texttt{DiffImpute} w/ ResNet} & 0.8519 & 0.7366 & 0.5801 & 0.3987 & 0.2309 & 0.1063 & 0.0390 & 0.0125 & 0.0039  \\
    \textbf{\texttt{DiffImpute} w/ Transformer} & \textbf{0.8875} & \textbf{0.8301} & \textbf{0.7386} & \textbf{0.6070} & \textbf{0.4427} & 0.2702 & 0.1313 & 0.0469 & 0.0103  \\
    \textbf{\texttt{DiffImpute} w/ U-Net} & 0.8321 & 0.7061 & 0.5542 & 0.3925 & 0.2477 & 0.1345 & 0.0598 & 0.0221 & 0.0060  \\
    \bottomrule
  \end{tabular}
  }
\end{center}
\end{table}

\begin{table}[!ht]
\caption{Downstream task performance comparison in random mask setting on YE, evaluated by RMSE, the best results are in \textbf{bold}.}
\label{tab:ds-rm-ye}
\begin{center}
\resizebox{\linewidth}{!}{
\begin{tabular}{l|ccccccccc}
    \toprule
       \textbf{Imputation Methods} & {10\%} & {20\%} & {30\%} & {40\%} & {50\%} & {60\%} & {70\%}& {80\%}& {90\%}\\
\hline
    Mean Imputation& 9.6483 & 9.9895 & 10.3056 & 10.5864 & 10.8372 & 11.0496 & \textbf{11.2184} & \textbf{11.3274} & \textbf{11.3629}\\
    Median Imputation& 9.6279 & 9.9625 & 10.2814 & 10.5704 & 10.8363 & 11.0667 & 11.2547 & 11.3902 & 11.4502    \\
    Mode Imputation& 9.7211 & 10.1028 & 10.4576 & 10.7646 & 11.1054 & 11.4239 & 11.7990 & 12.3536 & 13.2784  \\
    0 Imputation& 10.2651 & 10.8614 & 11.2272 & 11.4203 & 11.5104 & 11.5486 & 11.5515 & 11.5434 & 11.5288  \\
    1 Imputation& 10.4652 & 11.0338 & 11.329 & 11.4941 & 11.5855 & 11.6359 & 11.6544 & 11.6536 & 11.6344   \\
    LOCF Imputation& 12.4934 & 12.4969 & 12.4953 & 12.5030 & 12.5114 & 12.5117 & 12.4934 & 12.4934 & 12.4934   \\
    NOCB Imputation& 12.4883 & 12.4909 & 12.4963 & 12.5015 & 12.5267 & 12.5402 & 12.4883 & 12.4883 & 12.4883   \\
    MICE (linear) & 9.9231 & 9.8463 & 10.1061 & 10.4166 & 10.7099 & \textbf{11.0431} & 11.3486 & 11.6996 & 11.9950 \\
    GAIN & 9.9231 & 9.8463 & 10.8024 & 10.4166 & 11.3067 & 11.5499 & 11.4964 & 11.5453 & 11.5261  \\
    \hline
    \textbf{\texttt{DiffImpute} w/ MLP} & 10.2733 & 10.8953 & 11.2566 & 11.4651 & 11.5683 & 11.6109 & 11.6202 & 11.6075 & 11.5891  \\
    \textbf{\texttt{DiffImpute} w/ ResNet} & 9.6229 & 9.9908 & 10.3924 & 10.8053 & 11.0905 & 11.2565 & 11.4274 & 11.4886 & 11.4806  \\
    \textbf{\texttt{DiffImpute} w/ Transformer} & \textbf{9.5022} & \textbf{9.7544} & \textbf{10.0342} & \textbf{10.3449} & \textbf{10.6919} & 11.0639 & 11.4675 & 11.8724 & 12.2635  \\
    \textbf{\texttt{DiffImpute} w/ U-Net} & 9.8339 & 10.2640 & 10.5960 & 10.8568 & 11.0618 & 11.2840 & 11.5149 & 11.6760 & 11.7223  \\
    \bottomrule
  \end{tabular}
  }
\end{center}
\end{table}

\begin{table}[!ht]
\caption{Downstream task performance comparison in random mask setting on CO, evaluated by accuracy score, the best results are in \textbf{bold}.}
\label{tab:ds-rm-co}
\begin{center}
\resizebox{\linewidth}{!}{
\begin{tabular}{l|ccccccccc}
    \toprule
       \textbf{Imputation Methods} & {10\%} & {20\%} & {30\%} & {40\%} & {50\%} & {60\%} & {70\%}& {80\%}& {90\%}\\
\hline
    Mean Imputation& 0.8379 & 0.7526 & 0.6826 & 0.6252 & 0.5801 & 0.5447 & 0.5172 & 0.4978 & \textbf{0.4869} \\
    Median Imputation& 0.8397 & 0.7549 & 0.6850 & 0.6280 & 0.5827 & 0.5471 & 0.5206 & 0.5015 & 0.4905    \\
    Mode Imputation& 0.8270 & 0.7327 & 0.6549 & 0.5896 & 0.5343 & 0.4877 & 0.4473 & 0.4132 & 0.3853  \\
    0 Imputation& 0.8118 & 0.7020 & 0.6093 & 0.5284 & 0.4587 & 0.3985 & 0.3458 & 0.2982 & 0.2300  \\
    1 Imputation& 0.6544 & 0.5354 & 0.4691 & 0.4253 & 0.3940 & 0.3734 & 0.3633 & 0.3711 & 0.3942   \\
    LOCF Imputation& 0.4004 & 0.3872 & 0.3918 & 0.3951 & 0.3979 & 0.4001 & 0.4015 & 0.4035 & 0.4047   \\
    NOCB Imputation& 0.4001 & 0.3877 & 0.3927 & 0.3956 & 0.3982 & 0.3994 & 0.4015 & 0.4035 & 0.4043   \\
    MICE (linear) & 0.7608 & 0.6504 & 0.5881 & 0.4820 & 0.4332 & 0.3916 & 0.3852 & 0.4534 & 0.3657  \\
    GAIN & 0.8502 & 0.7707 & 0.6961 & 0.5760 & 0.4926 & 0.3988 & 0.3396 & 0.3098 & 0.2302   \\
    \hline
    \textbf{\texttt{DiffImpute} w/ MLP} & 0.7997 & 0.6870 & 0.6032 & 0.5397 & 0.4905 & 0.4522 & 0.4180 & 0.3898 & 0.3639  \\
    \textbf{\texttt{DiffImpute} w/ ResNet} & 0.8557 & 0.7796 & 0.7114 & 0.6523 & 0.6008 & 0.5556 & 0.5165 & 0.4889 & 0.4630  \\
    \textbf{\texttt{DiffImpute} w/ Transformer} & \textbf{0.8622} & \textbf{0.7904} & \textbf{0.7244} & \textbf{0.6646} & \textbf{0.6144} & \textbf{0.5710} & \textbf{0.5351} & \textbf{0.5031} & 0.4766  \\
    \textbf{\texttt{DiffImpute} w/ U-Net} & 0.8086 & 0.7027 & 0.6185 & 0.5505 & 0.4963 & 0.4490 & 0.4073 & 0.3700 & 0.3373  \\
    \bottomrule
  \end{tabular}
  }
\end{center}
\end{table}

\textbf{Column Mask.}
In this section, we assess the imputation performance using the Pearson correlation metric, specifically under the column mask settings. These settings are representative of the Missing at Random (MAR). Our evaluation spans across seven benchmark datasets, as detailed in \Cref{tab:ds-cm-ca,tab:ds-cm-he,tab:ds-cm-ja,tab:ds-cm-hi,tab:ds-cm-al,tab:ds-cm-ye,tab:ds-cm-co}. It's important to highlight that the NOCB imputation method is not applicable in this context, given the absence of a subsequent observation for backfilling missing values.

\begin{table}[!ht]
\caption{Downstream task performance comparison in column mask setting (MAR) on CA, evaluated by RMSE, the best results are in \textbf{bold}.}
\label{tab:ds-cm-ca}
\begin{center}\resizebox{\linewidth}{!}{
\begin{tabular}{l|cccc}
    \toprule
       \textbf{Imputation Methods} & {1} & {2} & {3} & {4}\\
      \hline
    Mean Imputation& 0.8321 & 0.9880 & 1.2584 & 1.2831 \\
    Median Imputation& 0.8474 & 1.0118 & 1.3759 & 1.2288 \\
    Mode Imputation& 0.925 & 1.0672 & 1.5293 & 1.2891   \\
    0 Imputation& 0.9295 & 1.7578 & 1.5229 & 1.7217   \\
    1 Imputation& 1.1986 & 1.1815 & 1.8452 & 1.8931 \\
    LOCF Imputation& 0.9175 & 1.0747 & 1.5489 & 1.3072 \\
    NOCB Imputation& / & / & / & / \\
    MICE (linear) & 0.7302 & \textbf{0.6850} & 1.2246 & \textbf{0.8795}\\
    GAIN & 0.7107 & 0.6862 & \textbf{0.9819} & 1.1849 \\
    \hline
    \textbf{\texttt{DiffImpute} w/ MLP} & 0.8775 & 1.2106 & 1.6318 & 1.6548\\
    \textbf{\texttt{DiffImpute} w/ ResNet} & 0.9440 & 1.8283 & 1.5211 & 1.8269 \\
    \textbf{\texttt{DiffImpute} w/ Transformer} & \textbf{0.7228} & 0.7790 & 1.0002 & 1.0263 \\
    \textbf{\texttt{DiffImpute} w/ U-Net} & 1.0677 & 1.9387 & 1.8962 & 2.0328 \\
    \bottomrule
  \end{tabular}}
\end{center}
\end{table}

\begin{table}[!ht]
\caption{Downstream task performance comparison in column mask setting (MAR) on HE, evaluated by accuracy score, the best results are in \textbf{bold}}
\label{tab:ds-cm-he}
\begin{center}\resizebox{\linewidth}{!}{
\begin{tabular}{l|cccc}
    \toprule
       \textbf{Imputation Methods} & {1} & {2} & {3} & {4}\\
      \hline
    Mean Imputation& 0.3547 & 0.3279 & 0.2832 & 0.2696 \\
    Median Imputation& 0.3550 & 0.3277 & 0.2816 & 0.2681 \\
    Mode Imputation& 0.3489 & 0.3141 & 0.2297 & 0.2364   \\
    0 Imputation& 0.3160 & 0.2528 & 0.1727 & 0.1808   \\
    1 Imputation& 0.3428 & 0.2626 & 0.1436 & 0.1376 \\
    LOCF Imputation& 0.3546 & 0.3250 & 0.2646 & 0.2667 \\
    NOCB Imputation& / & / & / & / \\
    MICE (linear) & 0.3576 & 0.3567 & 0.3232 & 0.2657\\
    GAIN & 0.3574 & \textbf{0.3571} & \textbf{0.3346} & \textbf{0.2809} \\
    \hline
    \textbf{\texttt{DiffImpute} w/ MLP} & 0.3416 & 0.2945 & 0.2186 & 0.2137\\
    \textbf{\texttt{DiffImpute} w/ ResNet} & 0.3340 & 0.2900 & 0.1712 & 0.1888 \\
    \textbf{\texttt{DiffImpute} w/ Transformer} & \textbf{0.3566} & 0.3393 & 0.3199 & 0.3117 \\
    \textbf{\texttt{DiffImpute} w/ U-Net} & 0.3352 & 0.2561 & 0.2634 & 0.2160 \\
    \bottomrule
  \end{tabular}}
\end{center}
\end{table}

\begin{table}[!ht]
\caption{Downstream task performance comparison in column mask setting (MAR) on JA, evaluated by accuracy score, the best results are in \textbf{bold}.}
\label{tab:ds-cm-ja}
\begin{center}\resizebox{\linewidth}{!}{
\begin{tabular}{l|cccc}
    \toprule
       \textbf{Imputation Methods} & {1} & {2} & {3} & {4}\\
      \hline
    Mean Imputation& 0.7108 & 0.7060 & 0.7005 & 0.6783 \\
    Median Imputation& 0.7103 & 0.7056 & 0.7009 & 0.6774 \\
    Mode Imputation& 0.7011 & 0.6987 & 0.6857 & 0.6532   \\
    0 Imputation& 0.7100 & 0.6806 & 0.6793 & 0.6158   \\
    1 Imputation& 0.6897 & 0.6862 & 0.6716 & 0.6021 \\
    LOCF Imputation& 0.7101 & 0.7056 & 0.6960 & 0.6608 \\
    NOCB Imputation& / & / & / & / \\
    MICE (linear) & \textbf{0.7131} & 0.6706 & 0.7097 & 0.6915\\
    GAIN & 0.7129 & 0.6843 & 0.6980 & \textbf{0.6995} \\
    \hline
    \textbf{\texttt{DiffImpute} w/ MLP} & 0.7082 & 0.6919 & 0.6908 & 0.6524\\
    \textbf{\texttt{DiffImpute} w/ ResNet} & 0.7103 & 0.6781 & 0.6825 & 0.6158 \\
    \textbf{\texttt{DiffImpute} w/ Transformer} & 0.7123 & \textbf{0.7078} & \textbf{0.7108} & 0.6937 \\
    \textbf{\texttt{DiffImpute} w/ U-Net} & 0.7061 & 0.6732 & 0.6755 & 0.6815 \\
    \bottomrule
  \end{tabular}}
\end{center}
\end{table}

\begin{table}[!ht]
\caption{Downstream task performance comparison in column mask setting (MAR) on HI, evaluated by accuracy score, the best results are in \textbf{bold}.}
\label{tab:ds-cm-hi}
\begin{center}\resizebox{\linewidth}{!}{
\begin{tabular}{l|cccc}
    \toprule
       \textbf{Imputation Methods} & {1} & {2} & {3} & {4}\\
      \hline
    Mean Imputation& 0.6964 & \textbf{0.6998} & 0.7022 & 0.6914 \\
    Median Imputation& 0.6970 & 0.6961 & 0.7006 & 0.6873 \\
    Mode Imputation& 0.6961 & 0.6961 & 0.7009 & 0.6856   \\
    0 Imputation& 0.6842 & 0.6832 & \textbf{0.7035} & 0.6718   \\
    1 Imputation& 0.6842 & 0.6263 & 0.6367 & 0.5959 \\
    LOCF Imputation& 0.6558 & 0.6954 & 0.6918 & 0.6888 \\
    NOCB Imputation& / & / & / & / \\
    MICE (linear) & 0.6350 & 0.6950 & 0.6840 & \textbf{0.6981}\\
    GAIN & 0.6473 & 0.6943 & 0.6849 & 0.6898 \\
    \hline
    \textbf{\texttt{DiffImpute} w/ MLP} & 0.6764 & 0.6669 & 0.6895 & 0.6544\\
    \textbf{\texttt{DiffImpute} w/ ResNet} & 0.6773 & 0.6756 & 0.7030 & 0.6647 \\
    \textbf{\texttt{DiffImpute} w/ Transformer} & \textbf{0.7032} & 0.6989 & 0.7027 & 0.6910 \\
    \textbf{\texttt{DiffImpute} w/ U-Net} & 0.6726 & 0.6434 & 0.6564 & 0.6562 \\
    \bottomrule
  \end{tabular}}
\end{center}
\end{table}

\begin{table}[!ht]
\caption{Downstream task performance comparison in column mask setting (MAR) on AL, evaluated by accuracy score, the best results are in \textbf{bold}.}
\label{tab:ds-cm-al}
\begin{center}\resizebox{\linewidth}{!}{
\begin{tabular}{l|cccc}
    \toprule
       \textbf{Imputation Methods} & {1} & {2} & {3} & {4}\\
      \hline
    Mean Imputation& 0.9164 & 0.9045 & 0.9047 & 0.8852\\
    Median Imputation& 0.9167 & 0.905 & 0.9052 & 0.8820 \\
    Mode Imputation& 0.9167 & 0.9023 & 0.9052 & 0.8830   \\
    0 Imputation& 0.9167 & 0.8954 & 0.9052 & 0.8458   \\
    1 Imputation& 0.7638 & 0.5757 & 0.4265 & 0.3502 \\
    LOCF Imputation& 0.9167 & 0.8247 & 0.8547 & 0.7762 \\
    NOCB Imputation& / & / & / & / \\
    MICE (linear) & 0.9108 & 0.9003 & 0.9116 & 0.9000\\
    GAIN & 0.9157 & 0.8958 & 0.9121 & 0.9011 \\
    \hline
    \textbf{\texttt{DiffImpute} w/ MLP} & 0.8783 & 0.8264 & 0.7922 & 0.7469\\
    \textbf{\texttt{DiffImpute} w/ ResNet} & 0.9161 & 0.8753 & 0.8917 & 0.8200 \\
    \textbf{\texttt{DiffImpute} w/ Transformer} & \textbf{0.9177} & 0.9124 & \textbf{0.9146} & \textbf{0.9047} \\
    \textbf{\texttt{DiffImpute} w/ U-Net} & 0.9161 & \textbf{0.9141} & 0.8978 & 0.8879 \\
    \bottomrule
  \end{tabular}}
\end{center}
\end{table}

\begin{table}[!ht]
\caption{Downstream task performance comparison in column mask setting (MAR) on YE, evaluated by RMSE (MAR), the best results are in \textbf{bold}.}
\label{tab:ds-cm-ye}
\begin{center}\resizebox{\linewidth}{!}{
\begin{tabular}{l|cccc}
    \toprule
       \textbf{Imputation Methods} & {1} & {2} & {3} & {4}\\
      \hline
    Mean Imputation& 9.2610 & 9.4197 & 9.3024 & 9.4945 \\
    Median Imputation& 9.2610 & 9.3982 & 9.2909 & 9.4762 \\
    Mode Imputation& 9.2610 & 9.3931 & 9.2842 & 9.4635   \\
    0 Imputation& 9.2610 & 9.6935 & 9.3141 & 10.1599   \\
    1 Imputation& 9.2606 & 10.1696 & 9.6535 & 10.2094 \\
    LOCF Imputation& 9.2610 & 9.4576 & 9.2906 & 9.6248 \\
    NOCB Imputation& / & / & / & / \\
    MICE (linear) & 9.261 & 9.4699 & \textbf{9.2610} & \textbf{9.3314}\\
    GAIN & 9.261 & 9.5965 & \textbf{9.2610} & 9.3885 \\
    \hline
    \textbf{\texttt{DiffImpute} w/ MLP} & \textbf{9.2609} & 9.9062 & 9.4708 & 10.2741\\
    \textbf{\texttt{DiffImpute} w/ ResNet} & 9.2611 & 9.6901 & 9.3116 & 10.1554 \\
    \textbf{\texttt{DiffImpute} w/ Transformer} & \textbf{9.2609} & 9.3298 & 9.2727 & 9.4193 \\
    \textbf{\texttt{DiffImpute} w/ U-Net} & \textbf{9.2609} & \textbf{9.2640} & 9.4906 & 9.3764 \\
    \bottomrule
  \end{tabular}}
\end{center}
\end{table}

\begin{table}[!ht]
\caption{Downstream task performance comparison in column mask setting (MAR) on CO, evaluated by accuracy score, the best results are in \textbf{bold}.}
\label{tab:ds-cm-co}
\begin{center}\resizebox{\linewidth}{!}{
\begin{tabular}{l|cccc}
    \toprule
       \textbf{Imputation Methods} & {1} & {2} & {3} & {4}\\
      \hline
    Mean Imputation& 0.8919 & 0.8890 & 0.8187 & 0.7491 \\
    Median Imputation& 0.8951 & 0.8924 & 0.8257 & 0.7610 \\
    Mode Imputation& 0.8799 & 0.8875 & 0.8099 & 0.7271   \\
    0 Imputation& 0.8784 & 0.8807 & 0.8064 & 0.7159   \\
    1 Imputation& 0.8370 & 0.7896 & 0.6767 & 0.6398 \\
    LOCF Imputation& 0.8939 & 0.8717 & 0.8223 & 0.7630 \\
    NOCB Imputation& / & / & / & / \\
    MICE (linear) & NaN & NaN & NaN & NaN\\
    GAIN & NaN & NaN & NaN & NaN \\
    \hline
    \textbf{\texttt{DiffImpute} w/ MLP} & 0.8836 & 0.8703 & 0.8077 & 0.7247\\
    \textbf{\texttt{DiffImpute} w/ ResNet} & 0.8938 & 0.8882 & 0.8233 & 0.7564 \\
    \textbf{\texttt{DiffImpute} w/ Transformer} & \textbf{0.8988} & 0.8962 & 0.8318 & 0.7745 \\
    \textbf{\texttt{DiffImpute} w/ U-Net} & 0.8870 & \textbf{0.9281} & \textbf{0.8746} & \textbf{0.7861} \\
    \bottomrule
  \end{tabular}}
\end{center}
\end{table}

\textbf{Downstream Tasks Performance Rankings.}
This section presents overall downstream tasks performance rankings under different mask settings (MCAR, and MAR) across seven datasets (\Cref{Downstream Metric ranking_MCAR,Downstream Metric ranking_MAR}).

\begin{table}[!ht]
\caption{Downstream task performance comparison under the random mask setting (MCAR) across the seven datasets. As different datasets apply different metrics, we report the performance rankings as the measurement. \texttt{DiffImpute} with Transformer has the best overall performance, the best results are in \textbf{bold}.}
\label{Downstream Metric ranking_MCAR}
\begin{center}
\resizebox{\linewidth}{!}{\begin{tabular}{l|lllllll|ll}
\toprule
\multicolumn{1}{c|}{\bf Imputation Methods}  &\multicolumn{1}{c}{\bf CA}  &\multicolumn{1}{c}{\bf HE}  &\multicolumn{1}{c}{\bf JA}  &\multicolumn{1}{c}{\bf HI}  &\multicolumn{1}{c}{\bf AL}  &\multicolumn{1}{c}{\bf YE}  &\multicolumn{1}{c|}{\bf CO} &\multicolumn{1}{c}{\bf Mean} &\multicolumn{1}{c}{\bf Std}\\
\hline
Mean Imputation         &3.8	&4.7	&7.7	&\textbf{1.4}	&7.6	&\textbf{3.0}	&4.0	&4.6	&2.1 \\
Median Imputation       &5.3	&6.1	&8.1	&2.3	&3.6	&3.0	&\textbf{3.0}	&4.5	&1.9 \\
Mode Imputation         &6.4	&7.3	&4.8	&3.8	&6.0	&8.1	&5.3	&6.0	&1.4 \\
0 Imputation            &10.7	&8.3	&7.6	&8.6	&9.0	&7.8	&10.0	&8.8 &1.1	\\
1 Imputation            &10.8	&11.2	&10.0	&12.0	&11.0	&10.0	&10.9	&10.8 &0.6	\\
LOCF Imputation         &8.4	&12.7	&12.0	&11.0	&13.0	&12.6	&10.1	&11.4	&1.5 \\
NOCB Imputation         &9.2	&12.1	&12.1	&12.0	&12.0	&12.2	&10.0	&11.4	&1.1 \\
MICE                    &3.1	&1.8	&\textbf{2.3}	&5.4	&1.6	&4.1	&9.6	&4.0	&2.6 \\
GAIN                    &6.3	&4.4	&4.0	&7.1	&4.9	&5.7	&7.7	&5.7	&1.3 \\
\hline
\textbf{\texttt{DiffImpute} w/ MLP}      &8.7	&8.4	&7.7	&8.2	&10.0	&9.0	&8.2	&8.6	&0.7 \\
\textbf{\texttt{DiffImpute} w/ ResNet }   &4.8	&2.8	&4.0	&6.6	&5.8	&4.6	&2.8	&4.5	&1.3 \\
\textbf{\texttt{DiffImpute} w/ Transformer}    &\textbf{1.2}	&\textbf{2.0}	&2.7	&2.4	&\textbf{1.4}	&3.7	&\textbf{1.3}	&2.1	&0.8 \\
\textbf{\texttt{DiffImpute} w/ U-Net}     &12.2	&9.0	&7.9	&10.0	&5.2	&7.0	&7.9	&8.5	&2.1 \\
\bottomrule
\end{tabular}}
\end{center}
\end{table}

\begin{table}[!ht]
\caption{Downstream task performance comparison under the column mask setting (MAR) across the seven datasets. As different datasets apply different metrics, we report the performance rankings as the measurement. \texttt{DiffImpute} with Transformer has the best overall performance, the best results are in \textbf{bold}.}
\label{Downstream Metric ranking_MAR}
\begin{center}
\resizebox{\linewidth}{!}{\begin{tabular}{l|lllllll|ll}
\toprule
\multicolumn{1}{c|}{\bf Imputation Methods}  &\multicolumn{1}{c}{\bf CA}  &\multicolumn{1}{c}{\bf HE}  &\multicolumn{1}{c}{\bf JA}  &\multicolumn{1}{c}{\bf HI}  &\multicolumn{1}{c}{\bf AL}  &\multicolumn{1}{c}{\bf YE}  &\multicolumn{1}{c|}{\bf CO} &\multicolumn{1}{c}{\bf Mean} &\multicolumn{1}{c}{\bf Std}\\
\hline
Mean Imputation         &4.3	&4.0	&3.8	&2.5	&5.5	&6.0	&5.3	&4.5	&1.1 \\
Median Imputation       &4.8	&4.5	&4.3	&4.3	&4.0	&5.3	&3.0	&4.3	&0.6 \\
Mode Imputation         &7.0	&7.3	&8.0	&4.8	&4.3	&4.3	&7.0	&6.1	&1.5 \\
0 Imputation            &8.8	&11.3	&9.3	&5.5	&5.5	&8.5	&8.5	&8.2	&1.9  \\
1 Imputation            &10.5	&10.5	&10.8	&10.3	&12.0	&9.0	&10.0	&10.4	&0.8 \\
LOCF Imputation         &7.5	&5.8	&5.8	&6.8	&8.3	&6.0	&4.8	&6.4	&1.1 \\
NOCB Imputation         & /&	&/ &/ &/ &/ &/ &/ &/	\\
MICE                    &2.0	&2.8	&4.5	&7.3	&5.5	&3.5	&NaN	&4.3	&1.8 \\
GAIN                    &\textbf{1.8}	&\textbf{1.5}	&4.0	&7.8	&5.0	&4.3	&NaN	&4.0	&2.1 \\
\hline
\textbf{\texttt{DiffImpute} w/ MLP}      &8.3	&8.8	&7.8	&9.3	&10.8	&8.8	&8.0	&8.8	&0.9  \\
\textbf{\texttt{DiffImpute} w/ ResNet }   &9.3	&10.3	&8.5	&6.8	&8.5	&9.5	&4.5	&8.2	&1.8 \\
\textbf{\texttt{DiffImpute} w/ Transformer}    &2.3	&2.5	&\textbf{1.8}	&\textbf{2.3}	&\textbf{1.3}	&\textbf{2.8}	&\textbf{1.8}	&2.1	&0.5 \\
\textbf{\texttt{DiffImpute} w/ U-Net}     &11.8	&9.0	&9.0	&10.3	&5.0	&4.0	&2.3	&7.3	&3.3  \\
\bottomrule
\end{tabular}}
\end{center}
\end{table}

\subsection{Time Performance.}
\textbf{Training Time.} 
In the subsequent tables, we present the training durations associated with various denoising models employed in our study. 
Notably, these durations exclude the time taken for \texttt{Harmonization} and \texttt{Impute-DDIM} processes. 
All time measurements are provided in seconds, as detailed in \Cref{Training Time}.

\begin{table}[ht]
\caption{The training time performance, measured in seconds, reveals that the U-Net model exhibits the longest training duration.}
\label{Training Time}
\begin{center}\resizebox{\linewidth}{!}{
\begin{tabular}{l|ccccccc}
\toprule
\multicolumn{1}{c|}{\bf Methods}  &\multicolumn{1}{c}{\bf CA}  &\multicolumn{1}{c}{\bf HE}  &\multicolumn{1}{c}{\bf JA}  &\multicolumn{1}{c}{\bf HI}  &\multicolumn{1}{c}{\bf AL}  &\multicolumn{1}{c}{\bf YE}  &\multicolumn{1}{c}{\bf CO} \\
\hline
\textbf{\texttt{DiffImpute} w/ MLP}      &16 &58 &54 &78 &72 &343 &488 \\
\textbf{\texttt{DiffImpute} w/ ResNet}   &26 &92 &82 &122 &107 &526 &743 \\
\textbf{\texttt{DiffImpute} w/ Transformer}     &88 &295 &267 &404 &386 &1762 &2428 \\
\textbf{\texttt{DiffImpute} w/ U-Net}    &267 &926 &856 &1252 &1180 &5555 &7572 \\
\bottomrule
\end{tabular}}
\end{center}
\end{table}

\textbf{Inference Time.} 
The subsequent tables detail the inference durations for the various denoising models incorporated in our research. 
It's noteworthy to mention that, based on the \texttt{Harmonization} algorithm (as seen in code snippet.~\ref{python_harmonization}), the inference time for models utilizing \texttt{Harmonization} witnessed an approximately fivefold increase. 
All durations are quantified in seconds, as elaborated in \Cref{Inference Time}.

\begin{table}[!ht]
\caption{The inference time performance, measured in seconds, reveals that the U-Net model exhibits the longest training duration.}
\label{Inference Time}
\begin{center}\resizebox{\linewidth}{!}{
\begin{tabular}{l|ccccccc}
\toprule
\multicolumn{1}{c|}{\bf Methods}  &\multicolumn{1}{c}{\bf CA}  &\multicolumn{1}{c}{\bf HE}  &\multicolumn{1}{c}{\bf JA}  &\multicolumn{1}{c}{\bf HI}  &\multicolumn{1}{c}{\bf AL}  &\multicolumn{1}{c}{\bf YE}  &\multicolumn{1}{c}{\bf CO} \\
\hline
\textbf{\texttt{DiffImpute} w/ MLP}      &3 &9 &19 &12 &13 &36 &71 \\
\textbf{\texttt{DiffImpute} w/ ResNet}   &4 &12 &24 &15 &16 &42 &89 \\
\textbf{\texttt{DiffImpute} w/ Transformer}     &11 &74 &298 &107 &553 &677 &913 \\
\textbf{\texttt{DiffImpute} w/ U-Net}    &27 &157 &869 &236 &959 &1827 &2519 \\
\bottomrule
\end{tabular}}
\end{center}
\end{table}

\subsection{Ablation Results without \texttt{Time Step Tokenizer}.}
This section demonstrates the ablation results after excluding the \texttt{Time Step Tokenizer}. The evaluations are specifically conducted under various missingness mechanisms, focusing on the CA dataset.

\textbf{Random Mask.}
Below, we present tables detailing the imputation outcomes under random mask settings. These outcomes are quantified using three metrics: mean squared error (MSE), Pearson correlation, and performance on downstream tasks. The respective results can be referenced in \Cref{tab:ap-rm-ca-wot,tab:pc-rm-ca-wot,tab:pc-rm-ca-wot}.

\begin{table}[!ht]
\caption{Imputation MSE performance comparison without \texttt{Time Step Tokenizer} in random mask (MCAR) setting on CA. The best results are in \textbf{bold}.}
\label{tab:ap-rm-ca-wot}
\begin{center}
\resizebox{\linewidth}{!}{
\begin{tabular}{l|ccccccccc}
    \toprule
       \textbf{Imputation Methods} & {10\%} & {20\%} & {30\%} & {40\%} & {50\%} & {60\%} & {70\%}& {80\%}& {90\%}\\
\hline
    \textbf{MLP w/o \texttt{Time Step Tokenizer}} & 0.0173	&0.0187	&0.0199	&0.0212	&0.0226	&0.0238	&0.0251	&0.0263	&0.0275  \\
    \textbf{ResNet w/o \texttt{Time Step Tokenizer}} & \textbf{0.0157}	&\textbf{0.0171}	&\textbf{0.0184}	&\textbf{0.0198}	&\textbf{0.0220}	&0.0255	&0.0321	&0.0448	&0.0658\\
    \textbf{Transformer w/o \texttt{Time Step Tokenizer}} & 0.0169	&0.0184	&0.0199	&0.0210	&0.0224	&0.0236	&0.0250	&0.0264	&0.0277 \\
    \textbf{U-Net w/o \texttt{Time Step Tokenizer}} & 0.0176	&0.0189	&0.0200	&0.0212	&0.0224	&\textbf{0.0234}	&\textbf{0.0245}	&\textbf{0.0257}	&\textbf{0.0266}  \\
    \bottomrule
  \end{tabular}
  }
\end{center}
\end{table}

\begin{table}[!ht]
\caption{Pearson correlation performance comparison without \texttt{Time Step Tokenizer} in random mask (MCAR) setting on CA. The best results are in \textbf{bold}.}
\label{tab:pc-rm-ca-wot}
\begin{center}
\resizebox{\linewidth}{!}{
\begin{tabular}{l|ccccccccc}
    \toprule
       \textbf{Imputation Methods} & {10\%} & {20\%} & {30\%} & {40\%} & {50\%} & {60\%} & {70\%}& {80\%}& {90\%}\\
\hline
    \textbf{MLP w/o \texttt{Time Step Tokenizer}} & 0.8515	&0.8379	&0.8284	&0.8167	&0.8035	&0.7920	&0.7797	&0.7678	&0.7569  \\
    \textbf{ResNet w/o \texttt{Time Step Tokenizer}} &\textbf{0.8648}&\textbf{0.8527}&\textbf{0.8426}	&\textbf{0.8332}	&\textbf{0.8180}	&\textbf{0.7984}	&0.7602	&0.6794	&0.5192\\
    \textbf{Transformer w/o \texttt{Time Step Tokenizer}} & 0.8531	&0.8389	&0.8268	&0.8174	&0.8041	&0.7931	&0.7790	&0.7651	&0.7527\\
    \textbf{U-Net w/o \texttt{Time Step Tokenizer}} & 0.8493	&0.8372	&0.8286	&0.8188	&0.8074	&0.7981	&\textbf{0.7865}	&\textbf{0.7756}	&\textbf{0.7661}\\
    \bottomrule
  \end{tabular}
  }
\end{center}
\end{table}

\begin{table}[!ht]
\caption{Downstream task performance comparison without \texttt{Time Step Tokenizer} in random mask (MCAR) setting on CA, evaluated by RMSE. The best results are in \textbf{bold}.}
\label{tab:ds-rm-ca-wot}
\begin{center}
\resizebox{\linewidth}{!}{
\begin{tabular}{l|ccccccccc}
    \toprule
       \textbf{Imputation Methods} & {10\%} & {20\%} & {30\%} & {40\%} & {50\%} & {60\%} & {70\%}& {80\%}& {90\%}\\
\hline
    \textbf{MLP w/o \texttt{Time Step Tokenizer}} & 0.7916	&0.8922	&0.9683	&1.0452	&1.1141	&1.1766	&1.2294	&1.2723	&1.3099  \\
    \textbf{ResNet w/o \texttt{Time Step Tokenizer}} &0.7909	&0.8914	&0.9656	&1.0409	&1.1169	&1.2139	&1.3766	&1.6312	&1.8705\\
    \textbf{Transformer w/o \texttt{Time Step Tokenizer}} & \textbf{0.7844}	&\textbf{0.8816}	&\textbf{0.9588}	&\textbf{1.0334}	&\textbf{1.1041}	&\textbf{1.1665}	&1.2242	&1.2687	&1.3095\\
    \textbf{U-Net w/o \texttt{Time Step Tokenizer}} & 0.7975	&0.8994	&0.9713	&1.0449	&1.1101	&1.1680	&\textbf{1.2166}	&\textbf{1.2536}	&\textbf{1.2892}\\
    \bottomrule
  \end{tabular}
  }
\end{center}
\end{table}

\textbf{Column Mask.}
Below, we present tables detailing the imputation outcomes under random mask settings. These outcomes are quantified using three metrics: mean squared error (MSE), Pearson correlation, and performance on downstream tasks. The respective results can be referenced in \Cref{tab:ap-cm-ca-wot,tab:pr-cm-ca-wot,tab:ds-cm-ca-wot}.

\begin{table}[!ht]
\caption{Imputation performance comparison without \texttt{Time Step Tokenizer} in column mask (MAR) setting on CA, evaluated by MSE. The best results are in \textbf{bold}.}
\label{tab:ap-cm-ca-wot}
\begin{center}\resizebox{\linewidth}{!}{
\begin{tabular}{l|cccc}
    \toprule
       \textbf{Imputation Methods} & {1} & {2} & {3} & {4}\\
\hline
    \textbf{MLP w/o \texttt{Time Step Tokenizer}} & 0.0196	&0.0223	&0.0198 	&0.0112\\
    \textbf{ResNet w/o \texttt{Time Step Tokenizer}} & 0.0741	&0.0951	&0.0914	&0.0722 \\
    \textbf{Transformer w/o \texttt{Time Step Tokenizer}} & 0.0191 	&0.0224	&0.0193	&0.0106
\\
    \textbf{U-Net w/o \texttt{Time Step Tokenizer}} & 0.2000 	&0.0180	&0.0268	&0.0205\\
    \bottomrule
  \end{tabular}}
\end{center}
\end{table}

\begin{table}[!ht]
\caption{Pearson correlation performance comparison without \texttt{Time Step Tokenizer} in column mask (MAR) setting on CA. The best results are in \textbf{bold}.}
\label{tab:pr-cm-ca-wot}
\begin{center}\resizebox{\linewidth}{!}{
\begin{tabular}{l|cccc}
    \toprule
       \textbf{Imputation Methods} & {1} & {2} & {3} & {4}\\
\hline
    \textbf{MLP w/o \texttt{Time Step Tokenizer}} & 0.1728	&0.5812	&0.7376	&0.8899\\
    \textbf{ResNet w/o \texttt{Time Step Tokenizer}} & 0.1983	&0.3260	&-0.0072 	&0.3305\\
    \textbf{Transformer w/o \texttt{Time Step Tokenizer}} & 0.1908	&0.5899	&0.7426	&0.8977
\\
    \textbf{U-Net w/o \texttt{Time Step Tokenizer}} & 0.1658	&0.5232	&0.7896	&0.7782\\
    \bottomrule
  \end{tabular}}
\end{center}
\end{table}

\begin{table}[!ht]
\caption{Downstream task performance comparison without \texttt{Time Step Tokenizer} in column mask (MAR) setting on CA, evaluated by RMSE. The best results are in \textbf{bold}.}
\label{tab:ds-cm-ca-wot}
\begin{center}\resizebox{\linewidth}{!}{
\begin{tabular}{l|cccc}
    \toprule
       \textbf{Imputation Methods} & {1} & {2} & {3} & {4}\\
\hline
    \textbf{MLP w/o \texttt{Time Step Tokenizer}} & 0.7566	&0.8494	&1.0102 	&1.1316\\
    \textbf{ResNet w/o \texttt{Time Step Tokenizer}} & 0.9282	&1.7319	&1.5977	&1.5334 \\
    \textbf{Transformer w/o \texttt{Time Step Tokenizer}} & 0.7498 	&0.8399	&1.0759	&1.0995
\\
    \textbf{U-Net w/o \texttt{Time Step Tokenizer}} & 0.7637 	&0.9363	&0.9413	&1.1991\\
    \bottomrule
  \end{tabular}}
\end{center}
\end{table}

\subsection{Ablation Results of \texttt{Harmonization}.}
This section delves into the imputation efficacy of four distinct denoising models when integrated with the \texttt{Harmonization} technique. The evaluations are specifically conducted under various missingness mechanisms, focusing on the CA dataset.

\textbf{Random Mask.}
Below, we present tables detailing the imputation outcomes under random mask settings. These outcomes are quantified using three metrics: mean squared error (MSE), Pearson correlation, and performance on downstream tasks. The respective results can be referenced in \Cref{tab:ap-rm-ca-har,tab:pc-rm-ca-har,tab:ds-rm-ca-har}.

\begin{table}[!ht]
\caption{Imputation MSE performance comparison with \texttt{Harmonization} in random mask (MCAR) setting on CA. The best results are in \textbf{bold}.}
\label{tab:ap-rm-ca-har}
\begin{center}
\resizebox{\linewidth}{!}{
\begin{tabular}{l|ccccccccc}
    \toprule
       \textbf{Imputation Methods} & {10\%} & {20\%} & {30\%} & {40\%} & {50\%} & {60\%} & {70\%}& {80\%}& {90\%}\\
\hline
    \textbf{\texttt{Harmonization} w/ MLP} & 0.0253 & 0.0258 & 0.0265 & 0.0268 & 0.0274 & 0.0280 & 0.0285 & 0.0292 & 0.0298  \\
    \textbf{\texttt{Harmonization} w/ ResNet} & \textbf{0.0146} & \textbf{0.0157} & \textbf{0.0169} & \textbf{0.0178} & \textbf{0.0189} & \textbf{0.0198} & \textbf{0.0208} & \textbf{0.0218} & \textbf{0.0229}  \\
    \textbf{\texttt{Harmonization} w/ Transformer} & 0.0155 & 0.0168 & 0.0180 & 0.0191 & 0.0206 & 0.0219 & 0.0232 & 0.0246 & 0.0258  \\
    \textbf{\texttt{Harmonization} w/ U-Net} & 2.0681 & 2.6099 & 3.1769 & 3.9142 & 4.7691 & 5.6382 & 6.6880 & 7.9535 & 9.2977  \\
    \bottomrule
  \end{tabular}
  }
\end{center}
\end{table}

\begin{table}[!ht]
\caption{Pearson correlation performance comparison with \texttt{Harmonization} in random mask (MCAR) setting on CA. The best results are in \textbf{bold}.}
\label{tab:pc-rm-ca-har}
\begin{center}
\resizebox{\linewidth}{!}{
\begin{tabular}{l|ccccccccc}
    \toprule
       \textbf{Imputation Methods} & {10\%} & {20\%} & {30\%} & {40\%} & {50\%} & {60\%} & {70\%}& {80\%}& {90\%}\\
\hline
    \textbf{\texttt{Harmonization} w/ MLP} & 0.7817 & 0.7774 & 0.7747 & 0.7733 & 0.7694 & 0.7655 & 0.7614 & 0.7569 & 0.7533  \\
    \textbf{\texttt{Harmonization} w/ ResNet} & 0.8752 & 0.8645 & 0.8554 & \textbf{0.8474} & \textbf{0.8373} & \textbf{0.8287} & \textbf{0.8184} & \textbf{0.8085} & \textbf{0.7986}  \\
    \textbf{\texttt{Harmonization} w/ Transformer} & \textbf{0.8772} & \textbf{0.8662} & \textbf{0.8566} & 0.8473 & 0.8352 & 0.8240 & 0.8115 & 0.7994 & 0.7883 \\
    \textbf{\texttt{Harmonization} w/ U-Net} & 0.0781 & 0.0726 & 0.0683 & 0.0677 & 0.0663 & 0.0668 & 0.0671 & 0.0652 &  0.656 \\
    \bottomrule
  \end{tabular}
  }
\end{center}
\end{table}

\begin{table}[!ht]
\caption{Downstream task performance comparison with \texttt{Harmonization} in MCAR setting on CA, evaluated by RMSE. The best results are in \textbf{bold}.
}
\label{tab:ds-rm-ca-har}
\begin{center}
\resizebox{\linewidth}{!}{
\begin{tabular}{l|ccccccccc}
    \toprule
       \textbf{Imputation Methods} & {10\%} & {20\%} & {30\%} & {40\%} & {50\%} & {60\%} & {70\%}& {80\%}& {90\%}\\
\hline
    \textbf{\texttt{Harmonization} w/ MLP} & 0.8692 & 1.0101 & 1.1407 & 1.1852 & 1.2500 & 1.3100 & 1.3547 & 1.3843 & 1.4057  \\
    \textbf{\texttt{Harmonization} w/ ResNet} & 0.7679 & 0.8574 & 0.9255 & 1.0000 & 1.0723 & 1.1369 & 1.2031 & 1.2612 & 1.3190 \\
    \textbf{\texttt{Harmonization} w/ Transformer} & \textbf{0.7486} & \textbf{0.8162} & \textbf{0.8705} & \textbf{0.9335} & \textbf{0.9943} & \textbf{1.0509} & \textbf{1.1076} & \textbf{1.1657} & \textbf{1.2264} \\
    \textbf{\texttt{Harmonization} w/ U-Net} & 1.1727 & 1.4774 & 1.6634 & 1.7959 & 1.8834 & 1.9391 & 1.9825 & 2.0146 & 2.0634 \\
    \bottomrule
  \end{tabular}
  }
\end{center}
\end{table}

\textbf{Column Mask.}
In the subsequent tables, we detail the imputation outcomes when operating under column mask settings. These results are gauged using three pivotal metrics: mean squared error (MSE), Pearson correlation, and efficacy on downstream tasks. For a comprehensive understanding, refer to \Cref{tab:ap-cm-ca-har,tab:pr-cm-ca-har,tab:ds-cm-ca-har}.

\begin{table}[!ht]
\caption{Imputation performance comparison with \texttt{Harmonization} in column mask (MAR) setting on CA, evaluated by MSE. The best results are in \textbf{bold}.}
\label{tab:ap-cm-ca-har}
\begin{center}\resizebox{\linewidth}{!}{
\begin{tabular}{l|cccc}
    \toprule
       \textbf{Imputation Methods} & {1} & {2} & {3} & {4}\\
\hline
    \textbf{\texttt{Harmonization} w/ MLP} & 0.02660 & 0.0296 & 0.0264 & 0.0189 \\
    \textbf{\texttt{Harmonization} w/ ResNet} & 0.0184 & 0.0203 & 0.0173 & 0.0095 \\
    \textbf{\texttt{Harmonization} w/ Transformer} & \textbf{0.0173} & \textbf{0.0202} & \textbf{0.0164} & \textbf{0.0084} \\
    \textbf{\texttt{Harmonization} w/ U-Net} & 2.1512 &	0.1604 & 2.5408	& 4.2775 \\
    \bottomrule
  \end{tabular}}
\end{center}
\end{table}

\begin{table}[!ht]
\caption{Pearson correlation performance comparison with \texttt{Harmonization} in column mask (MAR) setting on CA. The best results are in \textbf{bold}.}
\label{tab:pr-cm-ca-har}
\begin{center}\resizebox{\linewidth}{!}{
\begin{tabular}{l|cccc}
    \toprule
       \textbf{Imputation Methods} & {1} & {2} & {3} & {4}\\
\hline
    \textbf{\texttt{Harmonization} w/ MLP} & 0.0929 & 0.5159 & 0.6368 & 0.8193 \\
    \textbf{\texttt{Harmonization} w/ ResNet} & 0.2462 & 0.6083 & 0.7690 & 0.9112 \\
    \textbf{\texttt{Harmonization} w/ Transformer} & \textbf{0.4130} & \textbf{0.6877} & \textbf{0.8064} & \textbf{0.9286} \\
    \textbf{\texttt{Harmonization} w/ U-Net} & 0.1795 &	0.3662 & 0.1771	& 0.0948 \\
    \bottomrule
  \end{tabular}}
\end{center}
\end{table}

\begin{table}[!ht]
\caption{Downstream task performance comparison with \texttt{Harmonization} in column mask (MAR) setting on CA, evaluated by RMSE. The best results are in \textbf{bold}.}
\label{tab:ds-cm-ca-har}
\begin{center}\resizebox{\linewidth}{!}{
\begin{tabular}{l|cccc}
    \toprule
       \textbf{Imputation Methods} & {1} & {2} & {3} & {4}\\
\hline
    \textbf{\texttt{Harmonization} w/ MLP} & 0.8175 & 0.9961 & 1.2466 & 1.2839 \\
    \textbf{\texttt{Harmonization} w/ ResNet} & 0.7557 & 0.8718 & 1.0723 & 1.0908 \\
    \textbf{\texttt{Harmonization} w/ Transformer} & \textbf{0.7111} & \textbf{0.7647} & \textbf{0.9425} & \textbf{0.9991} \\
    \textbf{\texttt{Harmonization} w/ U-Net} & 0.9452 &	1.6025 & 1.4419	& 1.9054 \\
    \bottomrule
  \end{tabular}}
\end{center}
\end{table}

\subsection{Ablation Results of \texttt{Impute-DDIM}.}
The tables below display the experimental results of imputation performance using the \texttt{Impute-DDIM} technique on the CA dataset, with the retraced step set to $j=5$ and $\tau\in\{10,25,50,100,250\}$.

\textbf{Random Mask.}
The tables below shows the imputation performance with \texttt{Impute-DDIM} as evaluated by mean squared error (MSE) setting $\tau\in\{10,25,50,100,250\}$, under the random mask settings (\Cref{tab:ap-rm-ca-har-ddim-10,tab:ap-rm-ca-har-ddim-25,tab:ap-rm-ca-har-ddim-50,tab:ap-rm-ca-har-ddim-100,tab:ap-rm-ca-har-ddim-250}).

\begin{table}[!ht]
\caption{Imputation performance comparison with \texttt{Impute-DDIM} setting $\tau=10$ under the random mask (MCAR) setting on CA, evaluated by MSE. The best results are in \textbf{bold}.}
\label{tab:ap-rm-ca-har-ddim-10}
\begin{center}
\resizebox{\linewidth}{!}{
\begin{tabular}{l|ccccccccc}
    \toprule
       \textbf{Imputation Methods} & {10\%} & {20\%} & {30\%} & {40\%} & {50\%} & {60\%} & {70\%}& {80\%}& {90\%}\\
\hline
    \textbf{\texttt{Impute-DDIM} w/ MLP} & 0.2725	&0.2775	&0.2807	&0.2814	&0.2825	&0.2829	&0.2835	&0.2842	&0.2849\\
    \textbf{\texttt{Impute-DDIM} w/ ResNet} & 0.2483	&0.2539	&0.2580	&\textbf{0.2594} &\textbf{0.2608}	&\textbf{0.2615}	&\textbf{0.2623}	&\textbf{0.2633}	&\textbf{0.2640}\\
    \textbf{\texttt{Impute-DDIM} w/ Transformer} & \textbf{0.2438}	&\textbf{0.2511}	&\textbf{0.2571}	&0.2602	&0.2634	&0.2657	&0.2677	&0.2699	&0.2718\\
    \textbf{\texttt{Impute-DDIM} w/ U-Net} & 0.2678	&0.2719	&0.2748	&0.2752	&0.2759	&0.2760	&0.2762	&0.2766	&0.2771\\
    \bottomrule
  \end{tabular}
  }
\end{center}
\end{table}

\begin{table}[!ht]
\caption{Imputation performance comparison with \texttt{Impute-DDIM} setting $\tau=25$ under the random mask (MCAR) setting on CA, evaluated by MSE. The best results are in \textbf{bold}.}
\label{tab:ap-rm-ca-har-ddim-25}
\begin{center}
\resizebox{\linewidth}{!}{
\begin{tabular}{l|ccccccccc}
    \toprule
       \textbf{Imputation Methods} & {10\%} & {20\%} & {30\%} & {40\%} & {50\%} & {60\%} & {70\%}& {80\%}& {90\%}\\
\hline
    \textbf{\texttt{Impute-DDIM} w/ MLP} & 0.2301	&0.2354	&0.2398	&0.2417	&0.2437	&0.2452	&0.2467	&0.2483	&0.2499\\
    \textbf{\texttt{Impute-DDIM} w/ ResNet} & 0.1763	&0.1822	&0.1876	&0.1904	&0.1927	&0.1946	&\textbf{0.1962}	&\textbf{0.1980}	&\textbf{0.1997}\\
    \textbf{\texttt{Impute-DDIM} w/ Transformer} & \textbf{0.1601}	&\textbf{0.1692}	&\textbf{0.1774}	&\textbf{0.1834}	&\textbf{0.1890}	&\textbf{0.1937}	&0.1980	&0.2024	&0.2064\\
    \textbf{\texttt{Impute-DDIM} w/ U-Net} & 0.2191	&0.2236	&0.2268	&0.2279	&0.2289	&0.2296	&0.2302	&0.2311	&0.2321\\
    \bottomrule
  \end{tabular}
  }
\end{center}
\end{table}

\begin{table}[!ht]
\caption{Imputation performance comparison with \texttt{Impute-DDIM} setting $\tau=50$ under the random mask (MCAR) setting on CA, evaluated by MSE. The best results are in \textbf{bold}.}
\label{tab:ap-rm-ca-har-ddim-50}
\begin{center}
\resizebox{\linewidth}{!}{
\begin{tabular}{l|ccccccccc}
    \toprule
       \textbf{Imputation Methods} & {10\%} & {20\%} & {30\%} & {40\%} & {50\%} & {60\%} & {70\%}& {80\%}& {90\%}\\
\hline
    \textbf{\texttt{Impute-DDIM} w/ MLP} & 0.1778	&0.1832	&0.1881	&0.1911	&0.1940	&0.1964	&0.1990	&0.2014	&0.2039\\
    \textbf{\texttt{Impute-DDIM} w/ ResNet} & 0.1027	&0.1077	&0.1129	&0.1163	&0.1192	&0.1217	&0.1240	&0.1264	&0.1285\\
    \textbf{\texttt{Impute-DDIM} w/ Transformer} & \textbf{0.0801}	&\textbf{0.0867}	&\textbf{0.0934}	&\textbf{0.0992}	&\textbf{0.1049}	&\textbf{0.1103}	&\textbf{0.1152}	&\textbf{0.1204}	&\textbf{0.1253}\\
    \textbf{\texttt{Impute-DDIM} w/ U-Net} & 0.1638	&0.1673	&0.1701	&0.1720	&0.1734	&0.1750	&0.1760	&0.1774	&0.1788\\
    \bottomrule
  \end{tabular}
  }
\end{center}
\end{table}

\begin{table}[!ht]
\caption{Imputation performance comparison with \texttt{Impute-DDIM} setting $\tau=100$ under the random mask (MCAR) setting on CA, evaluated by MSE. The best results are in \textbf{bold}.}
\label{tab:ap-rm-ca-har-ddim-100}
\begin{center}
\resizebox{\linewidth}{!}{
\begin{tabular}{l|ccccccccc}
    \toprule
       \textbf{Imputation Methods} & {10\%} & {20\%} & {30\%} & {40\%} & {50\%} & {60\%} & {70\%}& {80\%}& {90\%}\\
\hline
    \textbf{\texttt{Impute-DDIM} w/ MLP} & 0.1135	&0.1175	&0.1224	&0.1259	&0.1291	&0.1324	&0.1358	&0.1390	&0.1420\\
    \textbf{\texttt{Impute-DDIM} w/ ResNet} & 0.0443	&0.0466	&0.0495	&0.0518	&0.0541	&0.0560	&0.0579	&0.0599	&0.0617\\
    \textbf{\texttt{Impute-DDIM} w/ Transformer} & \textbf{0.0281}	&\textbf{0.0303}	&\textbf{0.0329}	&\textbf{0.0351}	&\textbf{0.0375}	&\textbf{0.0399}	&\textbf{0.0423}	&\textbf{0.0451}	&\textbf{0.0477}\\
    \textbf{\texttt{Impute-DDIM} w/ U-Net} & 0.1064	&0.1091	&0.1114	&0.1131	&0.1147	&0.1165	&0.1180	&0.1199	&0.1218\\
    \bottomrule
  \end{tabular}
  }
\end{center}
\end{table}

\begin{table}[!ht]
\caption{Imputation performance comparison with \texttt{Impute-DDIM} setting $\tau=250$ under the in random mask (MCAR) setting on CA, evaluated by MSE. The best results are in \textbf{bold}.}
\label{tab:ap-rm-ca-har-ddim-250}
\begin{center}
\resizebox{\linewidth}{!}{
\begin{tabular}{l|ccccccccc}
    \toprule
       \textbf{Imputation Methods} & {10\%} & {20\%} & {30\%} & {40\%} & {50\%} & {60\%} & {70\%}& {80\%}& {90\%}\\
\hline
    \textbf{\texttt{Impute-DDIM} w/ MLP} & 0.0492	&0.0512	&0.0537	&0.0555	&0.0576	&0.0596	&0.0617	&0.0641	&0.0661\\
    \textbf{\texttt{Impute-DDIM} w/ ResNet} & 0.0210	&0.0219	&0.0230	&0.0238	&0.0248	&0.0257	&\textbf{0.0266}	&0.0276	&0.0285\\
    \textbf{\texttt{Impute-DDIM} w/ Transformer} & \textbf{0.0152}	&\textbf{0.0165}	&\textbf{0.0180}	&\textbf{0.0191}	&\textbf{0.0205}	&\textbf{0.0215}	&0.0277	&\textbf{0.0240}	&\textbf{0.0251}\\
    \textbf{\texttt{Impute-DDIM} w/ U-Net} & 0.0748	&0.0758	&0.0777	&0.0794	&0.0808	&0.0824	&0.0845	&0.0870	&0.0900\\
    \bottomrule
  \end{tabular}
  }
\end{center}
\end{table}

\textbf{Column Mask.}
The tables below shows the imputation performance with \texttt{Impute-DDIM} setting $\tau\in\{10,25,50,100,250\}$, as evaluated by mean squared error (MSE) under column mask settings (\Cref{tab:ap-cm-ca-har-ddim-10,tab:ap-cm-ca-har-ddim-25,tab:ap-cm-ca-har-ddim-50,tab:ap-cm-ca-har-ddim-100,tab:ap-cm-ca-har-ddim-250}).

\begin{table}[!ht]
\caption{Imputation performance comparison with \texttt{Impute-DDIM} setting $\tau=10$ under the column mask (MAR) setting on CA, evaluated by MSE. The best results are in \textbf{bold}.}
\label{tab:ap-cm-ca-har-ddim-10}
\begin{center}\resizebox{\linewidth}{!}{
\begin{tabular}{l|cccc}
    \toprule
       \textbf{Imputation Methods} & {1} & {2} & {3} & {4}\\
\hline
    \textbf{\texttt{Impute-DDIM} w/ MLP} & 0.2770 	&0.2922	&0.2715	&0.2581\\
    \textbf{\texttt{Impute-DDIM} w/ ResNet} & 0.2557 	&0.2707	&0.2505	&\textbf{0.2381}\\
    \textbf{\texttt{Impute-DDIM} w/ Transformer} &\textbf{0.2438} 	&\textbf{0.2635}	&\textbf{0.2477}	&0.2391\\
    \textbf{\texttt{Impute-DDIM} w/ U-Net} & 0.2732 	&0.2472	&0.3016	&0.2704\\
    \bottomrule
  \end{tabular}}
\end{center}
\end{table}

\begin{table}[!ht]
\caption{Imputation performance comparison with \texttt{Impute-DDIM} setting $\tau=25$ under the column mask (MAR) setting on CA, evaluated by MSE. The best results are in \textbf{bold}.}
\label{tab:ap-cm-ca-har-ddim-25}
\begin{center}\resizebox{\linewidth}{!}{
\begin{tabular}{l|cccc}
    \toprule
       \textbf{Imputation Methods} & {1} & {2} & {3} & {4}\\
\hline
    \textbf{\texttt{Impute-DDIM} w/ MLP} & 0.2333 	&0.2471	&0.2347	&0.2195\\
    \textbf{\texttt{Impute-DDIM} w/ ResNet} & 0.1854 	&0.1978	&0.1858	&0.1731\\
    \textbf{\texttt{Impute-DDIM} w/ Transformer} & \textbf{0.1572} 	&\textbf{0.1758}	&\textbf{0.1714}	&\textbf{0.1667}\\
    \textbf{\texttt{Impute-DDIM} w/ U-Net} & 0.2244 	&0.2067	&0.2461	&0.2302\\
    \bottomrule
  \end{tabular}}
\end{center}
\end{table}

\begin{table}[!ht]
\caption{Imputation performance comparison with \texttt{Impute-DDIM} setting $\tau=50$ under the column mask (MAR) setting on CA, evaluated by MSE. The best results are in \textbf{bold}.}
\label{tab:ap-cm-ca-har-ddim-50}
\begin{center}\resizebox{\linewidth}{!}{
\begin{tabular}{l|cccc}
    \toprule
       \textbf{Imputation Methods} & {1} & {2} & {3} & {4}\\
\hline
    \textbf{\texttt{Impute-DDIM} w/ MLP} & 0.1791 	&0.1908	&0.1874	&0.1708\\
    \textbf{\texttt{Impute-DDIM} w/ ResNet} & 0.1128 	&0.1216	&0.1157	&0.1036\\
    \textbf{\texttt{Impute-DDIM} w/ Transformer} & \textbf{0.0791} 	&\textbf{0.0889}	&\textbf{0.0916}	&\textbf{0.0861}\\
    \textbf{\texttt{Impute-DDIM} w/ U-Net} & 0.1679 	&0.1602	&0.1815	&0.1817\\
    \bottomrule
  \end{tabular}}
\end{center}
\end{table}

\begin{table}[!ht]
\caption{Imputation performance comparison with \texttt{Impute-DDIM} setting $\tau=100$ under the column mask (MAR) setting on CA, evaluated by MSE. The best results are in \textbf{bold}.}
\label{tab:ap-cm-ca-har-ddim-100}
\begin{center}\resizebox{\linewidth}{!}{
\begin{tabular}{l|cccc}
    \toprule
       \textbf{Imputation Methods} & {1} & {2} & {3} & {4}\\
\hline
    \textbf{\texttt{Impute-DDIM} w/ MLP} & 0.1133 	&0.1216	&0.1255	&0.1096\\
    \textbf{\texttt{Impute-DDIM} w/ ResNet} & 0.0506 	&0.0547	&0.0529	&0.0425\\
    \textbf{\texttt{Impute-DDIM} w/ Transformer} & \textbf{0.0312} 	&\textbf{0.0334}	&\textbf{0.0329}	&\textbf{0.0225}\\
    \textbf{\texttt{Impute-DDIM} w/ U-Net} & 0.1064 	&0.1098	&0.1163	&0.1250\\
    \bottomrule
  \end{tabular}}
\end{center}
\end{table}

\begin{table}[!ht]
\caption{Imputation performance comparison with \texttt{Impute-DDIM} setting $\tau=250$ under the column mask (MAR) setting on CA, evaluated by MSE. The best results are in \textbf{bold}.}
\label{tab:ap-cm-ca-har-ddim-250}
\begin{center}\resizebox{\linewidth}{!}{
\begin{tabular}{l|cccc}
    \toprule
       \textbf{Imputation Methods} & {1} & {2} & {3} & {4}\\
\hline
    \textbf{\texttt{Impute-DDIM} w/ MLP} & 0.0492 	&0.0536	&0.0555	&0.0452\\
    \textbf{\texttt{Impute-DDIM} w/ ResNet} & 0.0236 	&0.0262	&0.0238	&0.0154\\
    \textbf{\texttt{Impute-DDIM} w/ Transformer} & \textbf{0.0177} 	&\textbf{0.0205}	&\textbf{0.0168}	&\textbf{0.0085}\\
    \textbf{\texttt{Impute-DDIM} w/ U-Net} & 0.0622 	&0.0772	&0.0851	&0.0770\\
    \bottomrule
  \end{tabular}}
\end{center}
\end{table}

\subsection{Inference Time Ablation Study.}
In the subsequent tables, we present the inference durations associated with four distinct denoising networks. Specifically, we focus on the impact of integrating the \texttt{Harmonization} and \texttt{Impute-DDIM} techniques on the CA dataset.

\textbf{Impact of \texttt{Harmonization} on Inference Time.}
The table that follows delineates the inference durations for four denoising networks when the \texttt{Harmonization} technique is employed with a retraced step of \(j=5\). For a comprehensive understanding, we also provide a comparative analysis against scenarios where the \texttt{Harmonization} technique is not utilized (\Cref{ablation_result_inference_time}).

\begin{table}
\caption{Ablation of inference time comparison for \texttt{Harmonization}. The inference time is about five times longer when employing the \texttt{Harmonization} technique, which aligns with our algorithm~\ref{python_harmonization}. Time is measured in seconds.}
\label{ablation_result_inference_time}
\begin{center}\resizebox{\linewidth}{!}{
\begin{tabular}{c|cccc}
\toprule
\multicolumn{1}{c}{Technique} &\multicolumn{1}{c}{MLP} &\multicolumn{1}{c}{ResNet} &\multicolumn{1}{c}{ Transformer}&\multicolumn{1}{c}{U-Net}\\
\hline
    w/o \texttt{Harmonization} &3 &4 &27 &11 \\
    \texttt{Harmonization}  &15 &19 &53 &29 \\
\bottomrule
\end{tabular}}
\end{center}
\end{table}

\textbf{\texttt{Impute-DDIM} Inference Time.}
The table below illustrates the inference time of four denoising networks when utilizing the \texttt{Impute-DDIM} technique, with $\tau$ sequentially taking values from ${10, 25, 50, 100, 250, 500}$. The retraced step $j$ remains fixed at 5 in this context. Time is measured in seconds (\Cref{tab:ap-rm-ca-har-ddim}).

\begin{table}[!ht]
\caption{Imputation performance comparison with \texttt{Impute-DDIM} in random mask setting on CA, measured in seconds. Note that when $\tau=500$, no \texttt{Impute-DDIM} is applied.}
\label{tab:ap-rm-ca-har-ddim}
\begin{center}
\resizebox{\linewidth}{!}{
\begin{tabular}{l|cccccc}
    \toprule
       \textbf{Imputation Methods} & \multicolumn{1}{c}{$\tau=10$} & \multicolumn{1}{c}{$\tau=25$} & \multicolumn{1}{c}{$\tau=50$} & \multicolumn{1}{c}{$\tau=100$} & \multicolumn{1}{c}{$\tau=250$} & \multicolumn{1}{c}{$\tau=500$} \\
\hline
    \textbf{\texttt{Impute-DDIM} w/ MLP} & 2 & 1 & 2 & 3 & 8
 & 15   \\
    \textbf{\texttt{Impute-DDIM} w/ ResNet} & 1 & 1 & 2 & 4 & 10 & 19   \\
    \textbf{\texttt{Impute-DDIM} w/ Transformer} & 1 & 2 & 5 & 11
 & 26 & 53   \\
    \textbf{\texttt{Impute-DDIM} w/ U-Net} & 1 & 7 & 15 & 30 & 74
 & 149  \\
    \bottomrule
  \end{tabular}
  }
\end{center}
\end{table}

\subsection{Comparison Results with MIWAE (VAE-based Method).}

\textbf{Random Mask.}
In the subsequent tables, we present the imputation results when employing the MIWAE method \cite{mattei_2019_miwae}, a VAE-based approach, gauged using MSE under random mask conditions. 
This evaluation spans five datasets, specifically \Cref{tab:ap-rm-MIWAE,tab:pc-rm-MIWAE,tab:ds-rm-MIWAE}. 
It's worth noting that our experiments with MIWAE were confined to the CA, HE, JA, HI, and AL datasets. 
This limitation arises from the memory-intensive nature of the MIWAE method. 
Despite utilizing high-end GPUs like the NVIDIA A100, MIWAE often results in memory errors, underscoring its significant memory demands.

\begin{table}[!ht]
\caption{Imputation performance in terms of random mask setting (MCAR), using the MIWAE method, evaluated with MSE and downstream task metrics across five datasets. According to the experimental results from \Cref{tab:ap-rm-ca,tab:ap-rm-he,tab:ap-rm-ja,tab:ap-rm-hi,tab:ap-rm-al,tab:ap-rm-ye,tab:ap-rm-co}, MIWAE method is inferior to \texttt{DiffImpute} in most of the mask settings.}
\label{tab:ap-rm-MIWAE}
\begin{center}
\resizebox{\linewidth}{!}{
\begin{tabular}{l|ccccccccc}
    \toprule
       \textbf{Dataset} & {10\%} & {20\%} & {30\%} & {40\%} & {50\%} & {60\%} & {70\%}& {80\%}& {90\%}\\
\hline
    CA & 0.0228	&0.0233	&0.0233	&0.0231	&0.0234	&0.0236	&0.0235	&0.0234	&0.0235 \\
    HE & 0.0414	&0.0413	&0.0405	&0.0395	&0.0385	&0.0372	&0.0373	&0.0346	&0.0352   \\
    JA & 0.0388	&0.0395	&0.0430	&0.0402	&0.0412	&0.0390	&0.0380	&0.0369	&0.0350  \\
    HI & 0.0631	&0.0629	&0.0628	&0.0628	&0.0629	&0.0629	&0.0628	&0.0628	&0.0627  \\
    AL & 0.0199	&0.0199	&0.0199	&0.0199	&0.0200	&0.0200	&0.0200	&0.0200	&0.0200   \\
    \bottomrule
  \end{tabular}
  }
\end{center}
\end{table}

\begin{table}[!ht]
\caption{Imputation performance in terms of random mask setting (MCAR), using the MIWAE method, evaluated with Pearson correlation and downstream task metrics across five datasets.According to the experimental results from \Cref{tab:pc-rm-ca,tab:pc-rm-he,tab:pc-rm-ja,tab:pc-rm-hi,tab:pc-rm-al,tab:pc-rm-ye,tab:pc-rm-co}, MIWAE method is inferior to \texttt{DiffImpute} in most of the mask settings.}
\label{tab:pc-rm-MIWAE}
\begin{center}
\resizebox{\linewidth}{!}{
\begin{tabular}{l|ccccccccc}
    \toprule
       \textbf{Dataset} & {10\%} & {20\%} & {30\%} & {40\%} & {50\%} & {60\%} & {70\%}& {80\%}& {90\%}\\
\hline
    CA & 0.7995	&0.7962	&0.7950	&0.7957	&0.7938	&0.7926	&0.7924	&0.7942	&0.7940 \\
    HE & 0.6857	&0.6861	&0.6895	&0.6954	&0.7008	&0.7087	&0.7079	&0.7247	&0.7191   \\
    JA & 0.6501	&0.6450	&0.6253	&0.6402	&0.6349	&0.6461	&0.6510	&0.6569	&0.6714  \\
    HI & 0.5762	&0.5762	&0.5810	&0.5807	&0.5805	&0.5811	&0.5817	&0.5821	&0.5825  \\
    AL & 0.6399	&0.6402	&0.6400	&0.6408 &0.6400 &0.6399	&0.6393	&0.6393	&0.6391   \\
    \bottomrule
  \end{tabular}
  }
\end{center}
\end{table}

\begin{table}[!ht]
\caption{Imputation performance in terms of random mask setting (MCAR), using the MIWAE method, evaluated with downstream task metrics and downstream task metrics across five datasets. According to the experimental results from \Cref{tab:ds-rm-ca,tab:ds-rm-he,tab:ds-rm-ja,tab:ds-rm-hi,tab:ds-rm-al,tab:ds-rm-ye,tab:ds-rm-co}, MIWAE method is inferior to \texttt{DiffImpute} in most of the mask settings.}
\label{tab:ds-rm-MIWAE}
\begin{center}
\resizebox{\linewidth}{!}{
\begin{tabular}{l|ccccccccc}
    \toprule
       \textbf{Dataset} & {10\%} & {20\%} & {30\%} & {40\%} & {50\%} & {60\%} & {70\%}& {80\%}& {90\%}\\
\hline
    CA & 0.8768	&1.0215	&1.1207	&1.2059	&1.2682	&1.3132	&1.3511	&1.3535	&1.3535 \\
    HE & 0.3017	&0.2489	&0.2036	&0.1625	&0.1306	&0.1048	&0.0791	&0.0574	&0.0370   \\
    JA & 0.6792	&0.6423	&0.6088	&0.5766	&0.5428	&0.5054	&0.4717	&0.4302	&0.3858  \\
    HI & 0.6934	&0.6683	&0.6451	&0.6224	&0.6006	&0.5815	&0.5597	&0.5437	&0.5241  \\
    AL & 0.8210 &0.6897	&0.5375	&0.3893	&0.2558	&0.1477	&0.0735	&0.0284	&0.0081   \\
    \bottomrule
  \end{tabular}
  }
\end{center}
\end{table}

\textbf{Column Mask.}
In the following tables, we detail the imputation results using the MIWAE method, a VAE-based approach, assessed by the mean squared error (MSE) under column mask conditions. This assessment encompasses five datasets, as referenced in \Cref{tab:ap-cm-MIWAE,tab:pc-cm-MIWAE,tab:ds-cm-MIWAE}.

\begin{table}[!ht]
\caption{Imputation performance in terms of column mask setting (MAR), using the MIWAE method, evaluated with MSE across five datasets. According to the experimental results from \Cref{tab:ap-cm-ca,tab:ap-cm-he,tab:ap-cm-ja,tab:ap-cm-hi,tab:ap-cm-al,tab:ap-cm-ye,tab:ap-cm-co}, MIWAE method is inferior to \texttt{DiffImpute} in most of the mask settings.}
\label{tab:ap-cm-MIWAE}
\begin{center}
\begin{tabular}{l|cccc}
    \toprule
       \textbf{Dataset} & {1} & {2} & {3} & {4}\\
\hline
    CA & 0.0658	&0.0007	&0.0067	&0.0112\\
    HE & 0.0008	&0.0148	&0.0324	&0.0627\\
    JA & 0.0308	&0.0386	&0.0571	&0.0286\\
    HI & 0.0022	&0.0036	&0.0339	&0.0968\\
    AL & 0.0242	&0.0487	&0.0192	&0.0192\\
    \bottomrule
  \end{tabular}
\end{center}
\end{table}

\begin{table}[!ht]
\caption{Imputation performance in terms of column mask setting (MAR), using the MIWAE method, evaluated with Pearson correlation across five datasets. According to the experimental results from \Cref{tab:pr-cm-ca,tab:pr-cm-he,tab:pr-cm-ja,tab:pr-cm-hi,tab:pr-cm-al,tab:pr-cm-ye,tab:pr-cm-co}, MIWAE method is inferior to \texttt{DiffImpute} in most of the mask settings.}
\label{tab:pc-cm-MIWAE}
\begin{center}
\begin{tabular}{l|cccc}
    \toprule
       \textbf{Dataset} & {1} & {2} & {3} & {4}\\
\hline
    CA & 0.2132	&0.0073	&0.7670	&0.8795\\
    HE & -0.0152&0.8110	&0.4187	&0.3196\\
    JA & -0.0016&0.1886	&0.4394	&0.6895\\
    HI & 0.0129	&0.0356	&0.7108	&0.4627\\
    AL & -0.0039&0.3794	&0.1198	&0.5804\\
    \bottomrule
  \end{tabular}
\end{center}
\end{table}

\begin{table}[!ht]
\caption{Imputation performance in terms of column mask setting (MAR), using the MIWAE method, evaluated with downstream task metrics across five datasets. According to the experimental results from \Cref{tab:ds-cm-ca,tab:ds-cm-he,tab:ds-cm-ja,tab:ds-cm-hi,tab:ds-cm-al,tab:ds-cm-ye,tab:ds-cm-co}, MIWAE method is inferior to \texttt{DiffImpute} in most of the mask settings.}
\label{tab:ds-cm-MIWAE}
\begin{center}
\begin{tabular}{l|cccc}
    \toprule
       \textbf{Dataset} & {1} & {2} & {3} & {4}\\
\hline
    CA & 0.7122	&0.6853	&1.0053	&1.2930\\
    HE & 0.3571	&0.3570	&0.3065	&0.2556\\
    JA & 0.7130	&0.6845	&0.6699	&0.6951\\
    HI & 0.6566	&0.6882	&0.6834	&0.6794\\
    AL & 0.9126	&0.8780	&0.8977	&0.8887\\
    \bottomrule
  \end{tabular}
\end{center}
\end{table}
\clearpage

\end{document}